\documentclass{article}
\usepackage[utf8]{inputenc} % allow utf-8 input
\usepackage[main=english]{babel}	
\usepackage{arxiv}

\usepackage[T1]{fontenc}    % use 8-bit T1 fonts
\usepackage{hyperref}       % hyperlinks
\usepackage{url}            % simple URL typesetting
\usepackage{booktabs}       % professional-quality tables
\usepackage{amsfonts}       % blackboard math symbols
\usepackage{nicefrac}       % compact symbols for 1/2, etc.
\usepackage{microtype}      % microtypography
\usepackage{lipsum}         % Can be removed after putting your text content
\usepackage{graphicx}
\usepackage{natbib}
\usepackage{doi}
\usepackage{amsmath}
\usepackage{algorithm}
\usepackage{algpseudocode}
\usepackage{amsthm}
\usepackage{amssymb}
\usepackage{xcolor}
\usepackage{soul}
\usepackage{thmtools, thm-restate}
\newtheorem{theorem}{Theorem}
\newtheorem*{theorem-non}{Theorem}

\newtheorem*{lemma-non}{Lemma}

\title{A Mathematical Model of the Hidden Feedback Loop Effect in Machine Learning Systems}

\author{
    Andrey Veprikov\\
	Department of Intelligent Systems\\
	MIPT\\
	Dolgoprudny, Russia \\
	\texttt{veprikov.as@phystech.edu} 
    \And
    Alexander Afanasyev \\
    IITP\\
    Moscow, Russia \\
    \texttt{apa@isa.ru}
    \And
    Anton Khritankov\\
    HSE University, MIPT\\
    Moscow, Russia\\
    \texttt{akhritankov@hse.ru}\\
    }

\date{}

\renewcommand{\shorttitle}{OA Model of the Hidden Feedback Loop in ML Systems}

\begin{document}
\maketitle

\begin{abstract}

    Widespread deployment of societal-scale machine learning systems necessitates a thorough understanding of the resulting long-term effects these systems have on their environment, including loss of trustworthiness, bias amplification, and violation of AI safety requirements.
    We introduce a repeated learning process to jointly describe several phenomena attributed to unintended hidden feedback loops, such as error amplification, induced concept drift, echo chambers and others. The process comprises the entire cycle of obtaining the data, training the predictive model, and delivering predictions to end-users within a single mathematical model.
    A distinctive feature of such repeated learning setting is that the state of the environment becomes causally dependent on the learner itself over time, thus violating the usual assumptions about the data distribution.
    We present a novel dynamical systems model of the repeated learning process and prove the limiting set of probability distributions for positive and negative feedback loop modes of the system operation.
    We conduct a series of computational experiments using an exemplary supervised learning problem on 
    two synthetic data sets. The results of the experiments correspond to the theoretical predictions derived from the dynamical model. 
    Our results demonstrate the feasibility of the proposed approach for studying the repeated learning processes in machine learning systems and open a range of opportunities for further research in the area.

\end{abstract}

% keywords can be removed

%%%%%%%%% Introduction %%%%%%%%%
\section{Introduction} \label{Introduction}

    Societal-scale machine learning and decision making systems are, by definition, intended to have a major impact on society as a whole. Recent analysis \citep{cais2023statementairisk} presents a wide range of potential problems and areas of concern associated with such systems \citep{suresh2020misplaced}.     
    Addressing these challenges and various aspects of engineering trustworthy systems \citep{li2023trustworthy} requires different methods for designing machine learning and artificial intelligence systems, combining formal mathematical modelling,  data-driven engineering methods, long-term risk analysis  \citep{sifakis2023trustworthy, pei2022requirements, he2021challenges}.
    One of the key quality attributes of trustworthy ML systems \citep{serban2021practices,toreini2020relationship,siebert2020towards} and socially responsible AI (SRA) algorithms \citep{cheng2021socially} systems is their ability to behave in a way that users expect without any unintended side-effects.

    A \emph{repeated machine learning} process describes a situation in a machine learning system
    where the input data to a learning algorithm may depend in part on the previous predictions made by the system. 
 
    Machine learning methods usually take specific assumptions about the data, such as data has to be i.i.d., or stationary with white noise, or the environment the agent operates in remains the same, or there is a data drift independent from the learning agent. A distinctive feature of the repeated learning setting---the one that justifies introducing the name---is that the state of the environment in the process becomes causally dependent on the learning algorithm and the predictions given. 
    
    When there is a high automation bias, that is, when the use of predictions is high and adherence to them is tight, a so-called \emph{positive feedback loop} occurs \citep{khritankov2021hidden}. As a result of the loop, the learning algorithm is repeatedly applied to the data containing previous predictions. This repeated learning produces a noticeable unintended shift in the distributions of the input data and the predictions of the system \citep{khritankov2023positive}. Therefore, such a machine learning system would violate the trustworthiness requirements for socially responsible AI algorithms. 
    
    In many cases, the repeated ML process can represent the behaviour of a system interacting with its users. For example, in systems that recommend products to consumers or forecast market prices \citep{khritankov2021existence, sinha2016deconvolving} and learn from user responses, healthcare decision support systems \citep{adam2020hidden}, and predictive policing and public safety systems \citep{ensign2018runaway} that introduce bias in the training data as a result of an unintended feedback loop.

    The main contribution of this paper is a dynamical systems \citep{galor2007discrete} model of the repeated machine learning process. Formally, we consider a set $\textbf{F}$ of probability density functions (PDFs), each of which describes the data available to a machine learning system at a given time step $t$. We then introduce a mapping $\text{D}_t$ that acts on a given density function $f_t(x) \in \textbf{F}$ to produce a new data distribution $f_{t+1}(x)$. A general model of the repeated learning process we are studying can be written as
    \begin{equation*}
        f_{t+1}(x) = \text{D}_t(f_t)(x),\, f_0(x)\, \text{is known}, t = \{0, 1, 2,...\}.
    \end{equation*}
    In this model, the mapping $\text{D}_t$ may include the following actions performed by the system: sampling training data from $f_t(x)$, learning or updating parameters of a predictive model, taking inputs and providing one or more predictions to users. The application of the mapping $\text{D}_t$ results in a new probability distribution with a density function $f_{t+1}$.

    The structure of the paper is as follows. In Section~\ref{Related_work} we provide background on the problem of the hidden feedback loops and related work. In Section~\ref{Problem_statement}, we formally describe the mathematical model of the repeated learning process and reframe the aforementioned trustworthiness problems as research questions in terms of dynamical systems models. We then present our main theoretical results in Section~\ref{Main_results}. We find sufficient conditions for mapping $\text{D}_t$ to be a transformation on $\textbf{F}$. We find the limiting set and autonomy criterion for the dynamical system we introduce, prove convergence conditions and sufficient conditions for the mapping $\text{D}_t$ to be a non-contraction. In Section~\ref{Experiments} we test our theoretical results in a series of experiments on several data sets.

%%%%%%%%% Related Work %%%%%%%%%

\section{Related Work} \label{Related_work}    

    Let us give some background to the problem we are investigating. In many practical applications, the context in which a machine learning system is used may itself change the training data over time. Such \emph{data drift} may be due to some external factors or produced by the system itself. An illustration \citep{khritankov2021hidden} of this could be a housing prices prediction model that relies on actual purchases recommended by the model. In this way, the model learns in part from its own predictions. Case studies of similar effects, such as echo chambers and filter bubbles, are widely described in the literature \citep{davies2018redefining, spohr2017fake, michiels2022filter, khritankov2021existence}. Ensign et al. \citep{ensign2018runaway} have documented a positive feedback loop effect where a predictive policing system changes incidents data and introduces prediction bias. The workshop on fairness in machine learning \citep{chouldechova2020snapshot} revealed that unregulated hidden feedback loops can lead to decision biases, making them undesirable effects.
    
    An illustrative example is given by \cite{taori2023data}. If a training data set is dominated by elements of a particular class, then any optimal Bayes classifier would only predict new examples as being of that majority class. This would introduce bias into the data set. \cite{adam2022error} examine how feedback loops in machine learning systems affect classification algorithms. The authors describe \textit{error amplification} as an effect of an unintended feedback loop in a healthcare AI system which results in the loss of the prediction quality over time due to earlier prediction errors.

    \cite{khritankov2023positive} demonstrates a machine learning system that allows for unintended feedback loops, and the authors prove sufficient conditions for a positive feedback loop to occur, and provide a measurement procedure to estimate these conditions in practice. In this paper, we take a different approach and use dynamical system modeling.

    Our formulation of the problem employs an apparatus of discrete dynamical systems \citep{galor2007discrete, sandefur1990discrete}. The subject is less studied than continuous-time dynamical systems \citep{katok1995introduction, nemytskii2015qualitative, pauline2022observer, ouannas2017simple}. Primary research questions for discrete dynamical systems include: are there any fixed points for a given dynamical system \citep{milnor2018analytic}, what is the limit behaviour \citep{sharma2015uniform}, is a particular trajectory regular or chaotic \citep{zhang2006discrete}. 

%%%%%%%%% Problem statement %%%%%%%%%

\section{Mathematical Model and Problem Statement} \label{Problem_statement}

    Let us devise a mathematical model for a repeated machine learning process. In such a process, each step would roughly correspond to taking a sample of input data from a probability distribution, learning a model on a training set, evaluating model predictions on a held-out test sample, and mixing the predictions with the original data to get a new probability distribution at the next step thus introducing a causal loop \citep{khritankov2021hidden}.

    % added this to explain, why we can study y - y'
    Let us correspond a random vector $\text{X}_t$ to the internal state of the repeated learning process at time step $t$. We aim at describing the limit set $\lim_{t \to \infty} \text{X}_t$. Let $\mathbb{D}_t$ be a mapping such that the following recurrence relation holds
    \[
    \text{X}_{t+1} = \mathbb{D}_t (\text{X}_t).
    \]
    From the earlier work we may conclude that under certain conditions $\mathbb{D}_t$ could be a contractive transformation and might have a random fixed point \citep{itoh1977random}. So that, informally, $\text{X}_{\infty} = \lim_{t \to \infty} \text{X}_{t} = \mathbb{D}_t (\text{X}_{\infty})$.  

    % check this interpretation against Itoh's article
    The latter could be interpreted as follows. Given area measure $\rho$ in sample space $\Omega \subset \mathbb{R}^n$ and probability measure, for any subset $e \subset \Omega$ with measure $\rho(e) \leq \epsilon$, probability $\text{Pr}(\text{X}_t \in e)$ would tend to zero if the neighborhood does not contain the fixed point $X_{\infty}$, and $1$ otherwise, so that total $\text{Pr}(\text{X}_{\infty} \in e) = 1$. Hence, area measure $\rho$ for the limit set in $\mathbb{R}^n$ would be zero, that is $\text{X}_{\infty}$ could be a manifold of dimension at most $n-1$ and there could exist a non-zero functional $L_{\infty}: \mathbb{R}^n \to \mathbb{R}$ such that $L_{\infty}(X) = 0$ for any sample $X$ of $\text{X}_\infty$.
        
    Let $\text{Y}_t$ be an unbiased estimator of $\text{X}_t$, that is $\mathbb{E}\left[\text{X}_t - \text{Y}_t\right] = 0$, such that in the limit $\lim_{t \to \infty} \left(\text{X}_t - \text{Y}_t \right) = 0$.  For supervised learning problems let $X_t$ at step $t$ be a sample of $\text{X}_t$ of the form $X_t = (A_t, b_t)$. Then estimator $h_t$ is a solution to $L_t = L(b_t, h_t(A_{t-1})) \to \min$ with some loss function $L$. In this case, $Y_t = (A_t, h_t(A_t))$ and $X_t - Y_t = (0, b_t - h_t(A_{t}))$ for any sample $X_t$. Therefore, if there is a limit in the sequence of residuals of samples $b_t - h_t(A_t)$, then there is a point-wise limit of differences $\text{X}_t - \text{Y}_t$. Hence, for supervised problems, we can consider just the limit set of the random vectors of the form $\text{b}_t - h_t(\text{A}_t)$. % notation for h
    
    As we know, the theory of mappings of random vectors is well-developed only for smooth bijective transformations to solve random differential equations and this is a too strict assumption for the purpose of this research. Therefore, instead of approaching the problem of finding the limit set for random vectors $\text{X}_t - \text{Y}_t$ as it is, we reformulate it using probability density functions. 
      
    Consider a set $\textbf{F}$ of probability density functions and a series $\text{D}_t \in \mathbf{D}$ of transformations in the space $L_1(\mathbb{R}^n)$ containing these functions. Making a step in the repeated learning process corresponds to applying  mapping $\text{D}_t$ to a given density $f_t(x) \in \textbf{F}$ to get a new density function $f_{t+1}$.

    Therefore, the dynamics of the process can be described by the following recurrence relation with time step number $t$:
    \begin{equation}
        \label{system}
        f_{t+1}(x) = \text{D}_t(f_t)(x) ~ \forall x \in \mathbb{R}^n, t \in \mathbb{N}~\text{ and }~ \text{D}_t \in \mathbf{D},
    \end{equation}
    where $\text{D}_t$ is commonly called an evolution mapping, and the initial function $f_0(x)$ is given. 
    
    Compared to recurrence relations with linear operators or continuous dynamical systems with non-linear evolution operators, the evolution mapping in our case is an arbitrary transformation. We do not assume it to be deterministic, smooth or continuous. For example, a repeated learning process may use a stochastic learning algorithm, a random sampling procedure, and other data transformation methods, making the problem \eqref{system} a complex one to study.

    If the mappings $\{\text{D}_t\}_{t=0}^{\infty} \subset \mathbf{D}$ are independent of the time step $t$, then the relation \eqref{system} becomes
    \begin{equation}
        \label{system_aut}
        f_{t+1}(x) = \text{D}(f_t)(x) ~ \forall x \in \mathbb{R}^n \text{ and } t \in \mathbb{N}.
    \end{equation}

    The difference between these two systems \eqref{system} and \eqref{system_aut} is that the evolution operator in the latter \eqref{system_aut} remains the same for all $t$, making the system  autonomous, while it can change with step number $t$ in the former \eqref{system}. The learning algorithm or its parameters can therefore change over time in the \eqref{system} system, but remain the same for the other system.
    
    We use the proposed problem statement \eqref{system} to answer the following research questions for the repeated machine learning process.

    First, what are the necessary and sufficient conditions for  mapping $D_t$ to be a transformation on a set of probability density functions $\textbf{F}$. In other words, that the real-world processes and machine learning systems of interest can be represented as a repeated learning process.

    Second, we study the limit behaviour of the system \eqref{system} as the time step $t$ tends to infinity. It is important to understand how the repetitive nature of the learning process affects the operation of the system and the distribution of data in the long run.

    Third, we investigate if it is possible to empirically distinguish whether a particular machine learning system is better described by a non-autonomous \eqref{system} or an autonomous \eqref{system_aut} system models.

%%%%%%%%% Main Results %%%%%%%%%

\section{Main Results} \label{Main_results}

    In this Section, we derive the main theoretical results. Basic notation is provided in Section~\ref{Basic_Notation}. In Section~\ref{th_1_and_2} we derive the necessary modeling requirements on $\text{D}_t$ for our approach be applicable to real-world problems. In Section~\ref{th_3} we investigate properties of the arbitrary system \eqref{system}, while Section~\ref{th_6} is devoted to the autonomous system \eqref{system_aut}. Section~\ref{lemma_4_and_5} is devoted to investigation of Conjecture 1 from \citet{khritankov2021hidden}.

    \subsection{Basic Notation} \label{Basic_Notation}

    Set $\textbf{F}$ of probability density functions, is defined as
    \begin{equation} \label{R}
         \textbf{F} := \left\{f : \mathbb{R}^n \rightarrow \mathbb{R}_+ ~\text{and}~ \int_{\mathbb{R}^n}f(x)dx = 1\right\}.
     \end{equation}
    Let $f(x), x \in \mathbb{R}^n$, be a probability density function (PDF) from $\textbf{F}$, and $g(x)$ be a Lebesgue-measurable function from $L_1(\mathbb{R^n})$, and $\{\psi_t\}_{t = 0}^{\infty} > 0$ be a non-negative sequence over $\mathbb{R}$. 

    It is convenient to denote an application of a sequence of $t$ different mappings $\{\text{D}_t\}$ with $\text{D}_{\overline{1, t}}(\cdot) := \text{D}_t(\text{D}_{t-1} ( ... \text{D}_1( \cdot ) ... ))$.

    Let $\phi(x)$ be any continuous function with a compact support. We use Dirac's delta function $\delta(x)$, defined as a distribution
    $
           \int_{\mathbb{R}^n} \delta(x) \phi(x) dx =  \phi(0) ~~ \forall \phi(x)
    $, and introduce \emph{zero distribution} $\zeta(x)$, which we define as 
    $
        \int_{\mathbb{R}^n} \zeta(x) \phi(x) dx =  0 ~~ \forall \phi(x).
    $

    We say that $f_t(x) \underset{t \to +\infty}{\longrightarrow} f(x)$ \emph{weakly} when 
    $
        \int_{\mathbb{R}^n} f_t(x) \phi(x) \, dx \underset{t \to +\infty}{\longrightarrow} \int_{\mathbb{R}^n} f(x) \phi(x) dx ~~ \forall \phi(x).
    $
    We also use a classical definition of the $k$-th moment $\nu_{k}$ of a random variable $\xi$, that is
    $
        \nu_{k} := \mathbb{E}[\xi^k] = \int_{\mathbb{R}} x^{k} f_{\xi}(x) dx,
    $
    where $f_{\xi}(x)$ is the PDF of the random variable $\xi$.

    \subsection{Preliminary and Modeling Considerations} \label{th_1_and_2}

    Let us now derive useful properties of the problem. We begin with a more general case when the system is not necessarily autonomous, that is, has the form \eqref{system}.

    \begin{theorem}[\citealp{feller1991introduction}] \label{based}
        If the function $f: \mathbb{R}^n \to \mathbb{R}$ such that $f(x) \geq 0$ for almost every $x \in \mathbb{R}^n$ and $\|f\|_1 = \int\limits_{\mathbb{R}^n} f(x) dx = 1$, then there exists a random vector $\mathbf{\xi}$, for which $f$ will be a probability density  function.
    \end{theorem}

    Exactly on the basis of Theorem~\ref{based} we define $\textbf{F}$ \eqref{R} in this way. While in practice we can only observe distinct samples from our data at each step $t$, Theorem~\ref{based} links these samples to a random vector and its PDF. For the dynamical system model \eqref{system} to be interpretable and useful, we need each function $f \in \textbf{F}$ to be a probability density function of a random vector.

    Let us now derive conditions under which an arbitrary mapping $\text{D}_t$ is a transformation on $\textbf{F}$, that is, $\text{D}_t$ translates $\textbf{F}$ into $\textbf{F}$.

    \begin{restatable}[Conditions for $\text{D}_t$ to be a transformation on $\textbf{F}$]{theorem}{RtoR} \label{R_to_R}
    
        If the $\| \cdot \|_1$ norm of the mapping $\text{D}_t$ equals one, $\| \text{D}_t \|_1 = 1$, and for all PDFs $f(x) \in \textbf{F}$ holds that $\text{D}_t(f)(x) \geq 0$ for almost every $x \in \mathbb{R}^n$, and exists an inverse mapping $\text{D}_t^{-1}$ such that $\|\text{D}_t^{-1}\|_1 \leq 1$, then $\text{D}_t$ is a transformation on $\textbf{F}$.
        
    \end{restatable}

    The proof of Theorem~\ref{R_to_R} is provided in Section~\ref{pr_R_to_R}.

    \paragraph{Discussion of Theorem~\ref{R_to_R}.} In the experiments it is often difficult to compute the inverse mapping $\text{D}_t^{-1}$ and especially its norm, therefore we provide different conditions. 
    The distribution of our data can be approximated by an empirical distribution function \citep{dvoretzky1956asymptotic} as follows, when $n=1$:
    \begin{equation*}\label{F_approx}
        \hat{F}_N(x) := \frac{\text{number of elements in sample} \leq x}{N} = \dfrac{1}{N}\sum\limits_{i=1}^N \textbf{1}_{X_i \leq x},
    \end{equation*}
    where $X_i$ are elements of the sample. We assume that $(X_1, X_2, X_3, ... , X_N)$ are i.i.d. real random variables with the same cumulative distribution function (CDF) $F(x)$. If this condition is true, then the Dvoretzky–Kiefer–Wolfowitz–Massart (DKW) inequality holds
    \begin{equation*}\label{DKW}
        \mathbb{P}\left\{\underset{x \in \mathbb{R}}{\sup}\left|\hat{F}_N(x) - F(x)\right| > \varepsilon \right\} \leq C e^{-2N\varepsilon^2} \quad 
        \forall \varepsilon > 0.
    \end{equation*}
    Then we construct an interval that contains the true CDF of our data $F(x)$ with probability $1 - \alpha$ as
    \begin{equation*}\label{inter}
        \hat{F}_N(x) - \varepsilon \leq F(x) \leq \hat{F}_N(x) + \varepsilon, ~~ \text{ where } ~~ \varepsilon = \sqrt{\frac{\ln(2/\alpha)}{2N}}
    \end{equation*}
    In this case, the mapping $\text{D}_t$ transforms our data, that is, translates one empirical PDF into another empirical PDF. Thus, $\text{D}_t$ is by construction a transformation on $\textbf{F}$ in any practical application, including our experiments.

    \subsection{Results for a General System (\ref{system})} \label{th_3}
    
    The following theorem is the main result of the paper. It gives sufficient conditions for a feedback loop to occur in the system given by equation \eqref{system}.
    
    \begin{restatable}[Limit set]{theorem}{deltathm} \label{delta}
        For any probability density function $f_0(x), x \in \mathbb{R}^n$ and discrete dynamical system \eqref{system}, if there exists a measurable function $g(x)$ from $L_1\left(\mathbb{R}^n\right)$ and a non-negative sequence $\psi_t \geq 0$ such that $f_t\left(x\right) \leq \psi_t^n \cdot |g(\psi_t \cdot x)|$ for all $t \in \mathbb{N}$ and $x \in \mathbb{R}^n$.

        Then, if $\psi_t$ diverges to infinity, the density $f_t(x)$ tends to Dirac's delta function, $f_t(x) \underset{t \to +\infty}{\longrightarrow} \delta(x)$ weakly.  

        If $\psi_t$ converges to zero, then the density $f_t(x)$ tends to a zero distribution, $f_t(x) \underset{t \to +\infty}{\longrightarrow} \zeta(x)$ weakly.
        
    \end{restatable}

    The proof of Theorem~\ref{delta} is provided in Section~\ref{pr_delta}.

    \paragraph{Discussion of Theorem~\ref{delta}.}
    As we have already noted above in the Section~\ref{Problem_statement}, in practice, for a range of supervised learning problems with input data of the form $(X, y), y_i \in \mathbb{R}^n$, where $y_i$ is the target variable of the $i$-th item and model $h$ to solve $y' = h(X)$. 
    That is, instead of studying system \eqref{system} on $(X, y)$, we can consider a system constrained to residuals of model $y_i - y_i'$. This transition is necessary to conform assumptions of Theorem~\ref{delta}. In this case Theorem~\ref{delta} states that when the PDFs of residuals tend to delta function the positive feedback loop occurs, since the model errors tend to zero. And if these PDFs tend to zero distribution, then the error amplification occurs. In that case, with the help of Theorem~\ref{delta}, we can understand which of these two situations occurs in our system~\eqref{system}.

    We should note that the zero distribution $\zeta(x)$ in this paper is not the same as the zero. We introduce zero distribution as a limit of white noise PDFs. That is, $\zeta(x)$ can be thought of, for example,  as the PDF of a uniform distribution on the interval $(-\infty; +\infty)$ for $n=1$; or as the PDF of a normal distribution with infinite variance $\sigma^2 \to +\infty$. Dirac's delta function $\delta(x)$ in this case would be the PDF of a non-random variable equal to zero, or the PDF of a normal distribution with zero variance $\sigma^2 \to 0$.

    Since the Theorem~\ref{delta} conditions are satisfied for the sequence of functions $g_t(x) : = \psi_t^n \cdot |g(x \cdot \psi_t)| / \|g\|_1$, implying that the sequence $g_t$ tends towards either the delta function $\delta(x)$ or the zero distribution $\zeta(x)$, and $g_t(x)$ serves as an envelope for the sequence $f_t(x) = \text{D}_{\overline{1, t}}(f_0)(x)$, we can presume that our mappings $\text{D}_{\overline{1, t}}$ are in the form
    \begin{equation} \label{cool_D}
        \text{D}_{\overline{1, t}}(f_0)(x) = \psi_t^n \cdot f_0(\psi_t \cdot x) \quad \forall x \in \mathbb{R}^n \text{ and } \forall t \in \mathbb{N}.
    \end{equation}
    The mappings $\text{D}_{\overline{1, t}}$ transform the initial PDF $f_0(x)$ using a sequence $\psi_t$. The repeated application of $\text{D}_{\overline{1, t}}$ leads to the transformed PDF, determined by the behavior of the sequence at infinity.
    
    When $\psi_t$ converges to a constant $c \in (0, +\infty)$, then according to equation \eqref{cool_D} the distribution of our data remains the same, that is the mapping $\text{D}_{\overline{1, t}}$ is an identity mapping after some time step in the process.
    If we substitute $x = 0$ into the equation \eqref{cool_D}, then we can get an expression for $\psi_t$: $\psi_t = \sqrt[n]{f_t(0) / g(0)}$. And since we are interested in the behavior of $\psi_t$ at the infinity, we can assume that $\psi_t = f_t(0)$.  Let us take $\kappa > 0$ and consider an integral of the form 
    \begin{equation} \label{J_t}
        \int\limits_{B^n(\kappa)}f_t(x)dx = \int\limits_{B^n(\kappa)}\psi_t^n \cdot f_0(\psi_t \cdot x)dx = \int\limits_{B^n(\kappa \cdot \psi_t)} f_0(y)dy,
    \end{equation}
    Where $B^d(R)$ is a Euclidean ball with radius $R$. Therefore, if $\psi_t$ diverges to infinity, then integral \eqref{J_t} converges to $\|f_0\|_1 = 1$, and if $\psi_t$ converges to zero, then integral \eqref{J_t} will also converge to zero. In the experiments we measure $\psi_t = f_t(0)$ and $\hat{F}_t(\kappa) - \hat{F}_t(-\kappa)$, where $\kappa > 0$ is sufficiently small and $\hat{F}_t(x)$ is empirical CFD on step $t$ (in our experiments we consider case $n=1$). 

\begin{figure}
        \centering
        \includegraphics[width = 0.4\linewidth]{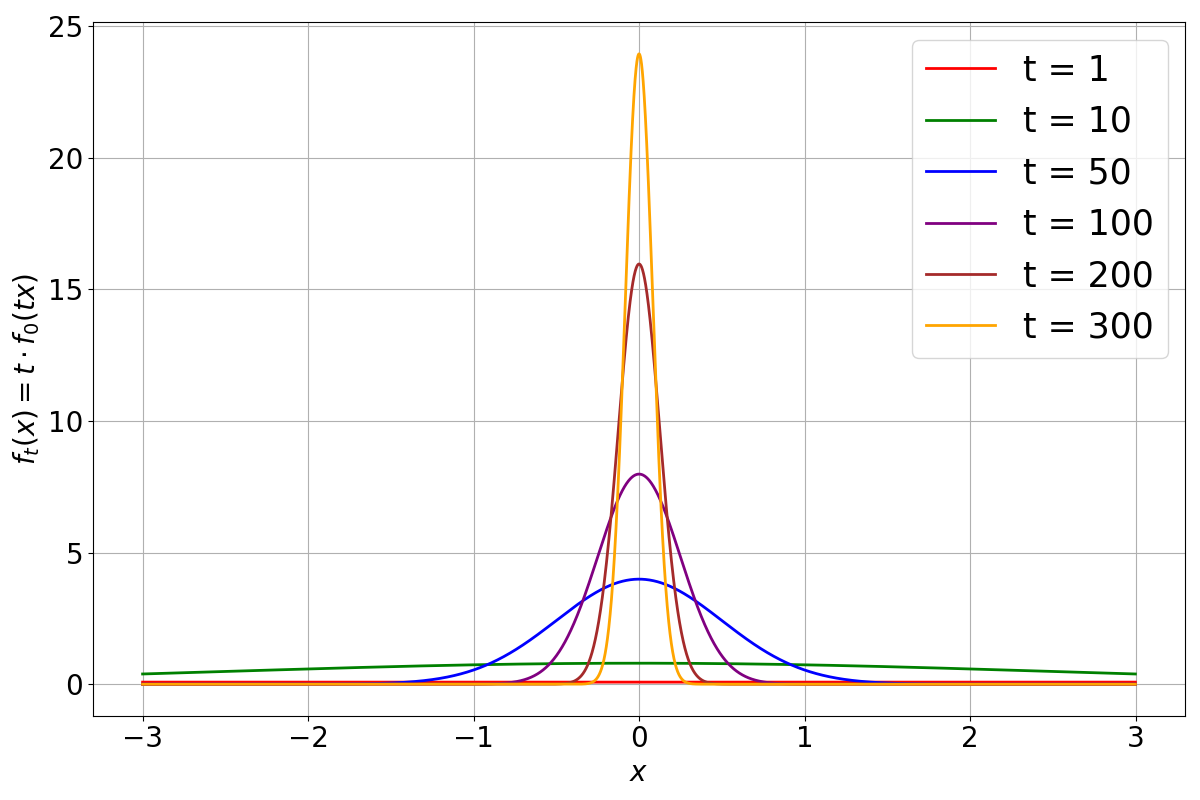}
        \includegraphics[width = 0.4\linewidth]{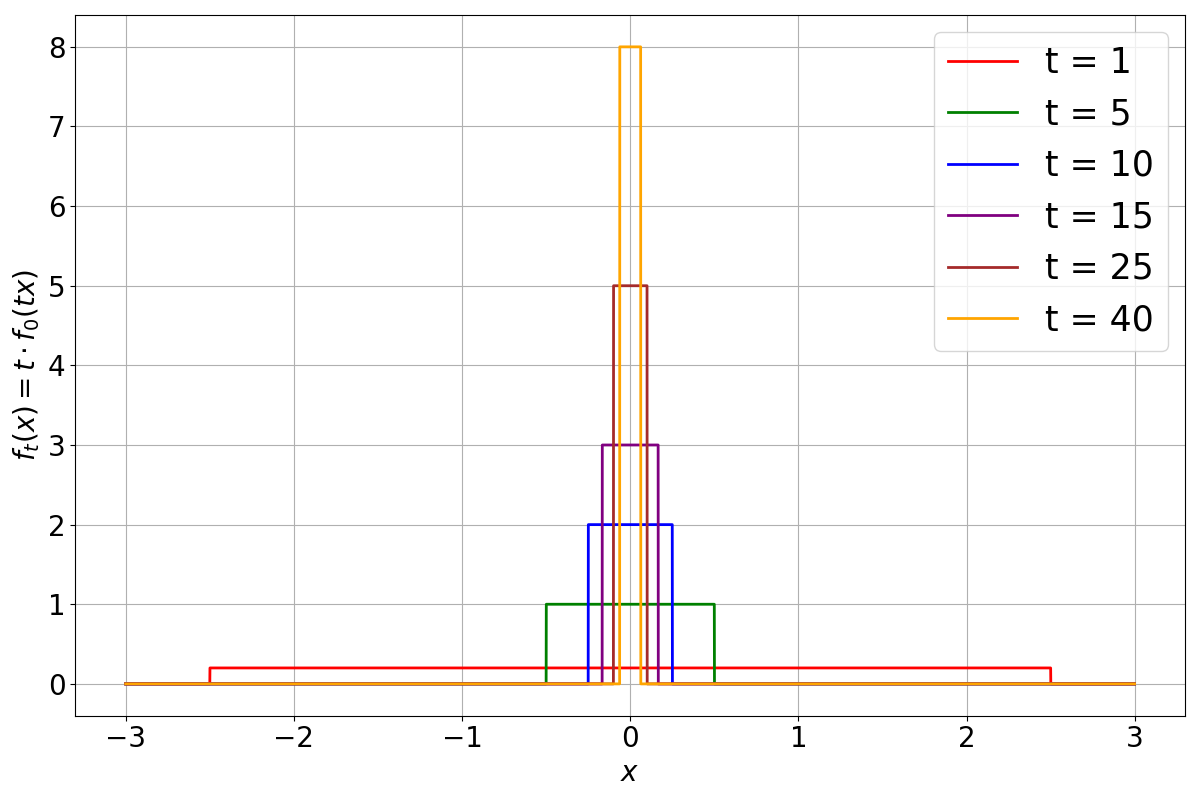}
        
        \caption{Illustration of weak limit to delta function. $\mathcal{N}(0, 5^2)$ (left), $\mathcal{U}[-2.5, 2.5]$ (right).}
        \label{example1_fig}
    \end{figure}

    \paragraph{Example of mappings $\text{D}_t$.} An important example of mappings $\text{D}_t$ that transform any function $f_0$ from $\textbf{F}$ into a delta function, is the following
    \begin{equation*} \label{example1}
        \text{D}_{\overline{1, t}}(f_0)(x) = t \cdot f_0(t \cdot x), ~~x \in \mathbb{R}.
    \end{equation*}
    Here we take $\psi_t = t$. Fig.~\ref{example1_fig} shows how this mapping translates the PDF of the normal distribution $\mathcal{N}(0, 5^2)$ and the continuous uniform distribution $\mathcal{U}[-2.5, 2.5]$ as $t \to +\infty$.

    \subsection[Analysis of Conjecture 1]{Analysis of Conjecture 1 (\citealp{khritankov2021hidden})} \label{lemma_4_and_5}

    In this Section, we investigate Conjecture 1 given by \citet{khritankov2021hidden}, which says that a positive feedback loop exists in system \eqref{system} if the operator $\text{D}_t$ is compressive in the metric space of the model predictions. However, this Conjecture was not proved in the original paper, we will give results that prove its correctness.

    We now give a corollary of the Theorem~\ref{delta}, which will help us  estimate the tendency of the moments of the random variables %$\mathbf{y} - \mathbf{y'}$ 
    $y_i - y_i'$ towards zero. 

    \begin{restatable}[Decreasing moments]{lemma}{momentslemma} \label{moments}
        If a system \eqref{system} with $n=1$ satisfies the conditions of the Theorem~\ref{delta} and $\psi_t$ diverges to infinity, then all even moments of the random variable 
        $y_i - y_i'$ (if they exist) decrease with a speed of at least $\psi_t^{-2k}$, that is $\nu_{2k}^t \leq \psi_t^{-2k} \nu_{2k}^0$, where $\nu_{2k}^t$ is a $2k$-th moment on a step $t$.

        If the evolution operator of the system \eqref{system} satisfies \eqref{cool_D}, then all (not only even) moments of the random variable 
        $y_i - y_i'$ (if they exist) decrease with the speed $\psi_t^{-k}$.

        If there exists $q \in [1; +\infty]$ such that $\{\nu_k^0\}_{k=1}^{+\infty} \in l_q$ and the system evolution operator \eqref{system} satisfies \eqref{cool_D}, then $\{\nu_k^t\}_{k=1}^{+\infty} \in l_1$ and $\{\nu_k^t\}_{k=1}^{+\infty} \underset{t \to \infty}{\overset{l_1}{\longrightarrow}} 0$.
    \end{restatable}

    The proof of Lemma~\ref{moments} is provided in Section~\ref{pr_moments}.

    \paragraph{Discussion of Lemma~\ref{moments}.} Lemma~\ref{moments} is interesting from a practical point of view, since calculations of moments of random variables are much easier to verify in a computational experiment than the conditions of Theorem~\ref{delta}.

    This Lemma also helps in analysis of Conjecture 1  \citep{khritankov2021hidden}, since even moments, such as variance, can be considered as an indicator of predictions quality. Consequently, if there is a tendency towards a delta function, then the mapping $\text{D}_t$ is contractive in a vector space of regression quality metrics. But if sequence $f_t$ tends towards the delta function, then there is a positive feedback loop in our system \eqref{system}, that is Conjecture~1 is true in this case.
    
    The next lemma also attempts to analyse the Conjecture~1 in our formulation of the problem \eqref{system}.
    
    \begin{restatable}[Inequality on $\|\text{D}_t\|_q$]{lemma}{ineqqlemma} \label{ineq_q}
        Consider a function
        \begin{equation*}
            f_A(x) = \dfrac{1}{\lambda(A)} \cdot \textbf{1}_{A}(x),
        \end{equation*}
        where $A \subset \mathbb{R}^n$ is an arbitrary Lebesgue-measurable set of a non-path measure, $\lambda(A)$---the measure of a set $A$.

        Then for all $A \subset \mathbb{R}^n$ such that $\lambda(A) \in (0; +\infty)$ and for all $q \in [1; +\infty]$ such that $\text{D}_t(f_A) \in L_q(\mathbb{R}^n)$ the following inequality holds  

        \begin{equation*}
            \|\text{D}_t\|_q \geq \int\limits_{A} \text{D}_t(f_A)(x)dx.
        \end{equation*}
    \end{restatable}

    The proof of Lemma~\ref{ineq_q} is provided in Section~\ref{pr_ineq_q}.
        
    \paragraph{Discussion of Lemma~\ref{ineq_q}.} First of all, note that the result of Lemma~\ref{ineq_q} does not depend in any way on whether $\text{D}_t$ is a transformation on $\textbf{F}$ or not, it is a consequence of H\"older's inequality. 
    
    If we consider the problem statement from this paper, hence $\text{D}_t$ translates $\textbf{F}$ into $\textbf{F}$, then the function $f_A$ is a PDF of vectors uniformly distributed on a set $A$. Consequently, $\int_{A} \text{D}_t(f_A)(x)dx \leq 1$, that is from Lemma~\ref{ineq_q} we can only conclude that the $\| \cdot \|_q$ norm of the operator $\text{D}_t$ is greater than or equal to one. But if $\|\text{D}_t\|_q \geq 1$, then $\text{D}_t$ would not be a contraction mapping in $\|\cdot\|_q$, because there would always be a function $f \in \textbf{F}$ such that $\|\text{D}_t(f)\|_q \geq \|f\|_q$. This is important because if there is a tendency to zero distribution $\zeta(x)$ in our system, then the mappings $\text{D}_t$ would be contractions in any norm $\| \cdot \|_q$, that is $\left\|\text{D}_t\right\|_q \leq 1$. 

    Lemma~\ref{ineq_q} helps us to investigate Conjecture~1 from prior work \citep{khritankov2021hidden}. If a mapping $\text{D}_t$ is expansive in $\| \cdot \|_q$-norm, that is, $\left\|\text{D}_t\right\|_q \geq 1$, then it is contractive in a vector space of regression quality metrics, since the $\| \cdot \|_q$-norm of the function increases as it tends towards the delta function.

    \subsection{Results for an Autonomous System (\ref{system_aut})} \label{th_6}

    Autonomy is an important property of any dynamical system. Such systems do not depend on the initial time $t_0$ from which the observation of this system started. Also, the Theorem~\ref{delta} of this paper considers the tendency of time step $t$ to infinity, but in practice we can only consider finite $t$. However, if the system is autonomous, we can study our system on a finite interval and understand its behavior at infinite $t$.

    From Theorem~\ref{delta} we can see that there is a special kind of mappings \eqref{cool_D} that bound the PDFs of our data from above, that is the mappings of this type that we will consider in this section. For such mappings we can derive an autonomy criterion for the system \eqref{system_aut}.

    \begin{restatable}[Autonomy criterion]{theorem}{semigroupthm} \label{semigroup}
    
        If the evolution operators $\text{D}_t$ of a dynamic system \eqref{system} have the form \eqref{cool_D}, then the system is autonomous if and only if

        \begin{equation} \label{cond_semigroup}
            \psi_{\tau + \kappa} = \psi_{\tau} \cdot \psi_{\kappa} ~\forall \tau, \kappa \in \mathbb{N}.
        \end{equation}
        
    \end{restatable}

    The proof of Theorem~\ref{semigroup} provided in Section~\ref{pr_semigroup}.

    \paragraph{Discussion of Theorem~\ref{semigroup}.} This criterion is easy to check in practice, since the condition \eqref{cond_semigroup} means that the sequence $\psi_t$ is a power sequence, that is $\psi_t = a^t$ for some $a > 0$. 
    
    An example of a mapping of the form \eqref{cool_D} is given below with the name Sampling update \ref{ex_set} in Section~\ref{Experiments}.

    The autonomy of the system helps us to experimentally determine the shape of the graph of $\psi_t$, it should be similar to a power function. We demonstrate how Theorem~\ref{semigroup} can be applied in Experiment~\ref{exp_4}.

%%%%%%%%% Experiments %%%%%%%%%

\section{Experiments} \label{Experiments}

    The goal of our experiments is to compare the theoretical predictions given in Section~\ref{Main_results} with actual measurements. In Section~\ref{design}, we present two experiment designs: sliding window and sampling update. In Section~\ref{exp_1}, we describe an empirical study of the limit set of the system \eqref{system}. Section~\ref{exp_2} focuses on analysing the normality of the distribution of the training sample as time step $t$ tends to infinity. In Section~\ref{exp_3}, we check the predictions of Theorem~\ref{delta} against measurements in one-dimensional case. In Section~\ref{exp_4}, we analyse the behavior of the system in the experiment design introduced earlier for autonomy using Theorem~\ref{semigroup}. Section~5 will be devoted to verifying Lemma~\ref{moments} in practice.

    \subsection{Experiment Design} \label{design}

        We conduct several experiments to test theoretical predictions we devised in the previous section. The goal of these experiments is to compare the predictions with the actual observations in a controlled environment.
    
        Here is a formal statement of a exemplary problem to demonstrate the repeated learning process. Following the problem statement \eqref{system}, let $\textbf{F}$ \eqref{R} be a space of probability density functions. At step $t = 0$ we take the initial $f_0 \in \textbf{F}$ and sample an original set $(\textbf{X}, \mathbf{y})$ of size $n$ from $f_0$ and take $\epsilon$ as normally distributed noise. We consider a regression problem with a loss function $L$, that is find $\theta^*$ such that
        
        \begin{equation*} \label{regression}
            \theta^* = \text{argmin}_{\theta} L (\mathbf{y}, \textbf{X} \cdot \theta + \epsilon).
        \end{equation*}

        In order to explore how regularization, cross-validation and learning algorithms affect the theoretical results, we use a linear regression model without regularization learned with an SGD algorithm with the maximum number of iterations equal to 50, a Ridge regression model without any regularization and solve it in closed-form using Cholesky decomposition, and RidgeCV regression model with the regularization parameter equal to $0.1$ solved using SVD. Models and learning algorithms are implemented in the Scikit-learn library \citep{pedregosa2011scikit}.

        We take synthetic initial data sets in order to limit unknown confounding factors and isolate the effect of the repeated learning. In the first data set, input data $\textbf{X}$ is normally distributed and $\mathbf{y}$ is a linear function of $\textbf{X}$ with additional normal noise. As the second data set we take Friedman problem \citep{friedman1991multivariate}, which is not linear.
        Both of these data sets are obtained from the Scikit-learn library \citep{pedregosa2011scikit} using make\_regression() and make\_friedman1() routines.
        The number of objects in the data set $\textbf{X}$ equals $2000$ and the number of features is $10$.
        At step $t=0$ the input data in both cases is i.i.d., relation between $\textbf{X}$ and $\mathbf{y}$ is linear in the first case. We run each experiment ten times to reduce the randomness.

        We employ MLDev reproducible experiments toolkit \citep{khritankov2021mldev} to implement the simulation environment. The source code and initial data to reproduce the experiments can be found in the Gitlab repository \footnotemark.
        
        \footnotetext{Source code for the experiments: \url{https://gitlab.com/repeated_ml/dynamic-systems-model}}

        In the \emph{sliding window update} experiment setting \citep{khritankov2023positive}, at round $r = 0$ we first sample $30\%$ of the original data $\textbf{X}$ on which our model $h_0$ is trained on a $80\%$ subset of input data.   
        Then, at each step $t$, we randomly sample without replacement an item $(\mathbf{x}^i, y_i)$---the features and target variable of the item $i$ from the remaining data. Next, we obtain the prediction $y'_i$ from the model $y_i' = h_t(\mathbf{x^i})$ and sample $z_i \sim \mathcal{N}(y_i', s \cdot \sigma^2)$, where $s$ is \emph{adherence} \citep{khritankov2023positive}, a parameter of the experiment, and $\sigma$ is the mean squared error of the predictions on the held-out $30\%$ subset of the current set of step $t - 1$. After that, we evict the earliest item from the current set and append the new item $(\mathbf{x^i}, z_i)$ with \emph{usage} \citep{khritankov2023positive} probability $p$ or item $(\mathbf{x^i}, y_i)$ with probability $(1-p)$. We repeat the procedure until we run out of items in the original set, making a total of $0.7 \cdot n$ steps. After every $T$ steps, $r$ is increased: $r = r + 1$ and the machine learning model $h_t$ is retrained with a training size of $80\%$ on the active set.

        In the \emph{sampling update} setting, the model $h_0$ is trained on the entire sample $\textbf{X}$ at round $r = 0$. Then the procedure is very similar to the sliding window update, but when we take the element $(\mathbf{x^i}, y_i)$ from the original set, we replace it using the same rules as in the sliding window setting. The consequence of such change is, first, that in the sampling update setting, the system would be autonomous as the procedure, and therefore the evolution operator $\text{D}$, do not depend on step $t$. Second, is that we can run the sampling update experiment for unlimited time steps, whereas the sliding window experiment allows for only a maximum of $0.7 \cdot n$ iterations. 

        In both cases the size of the sliding window remains constant--$0.3 \cdot n$ for both the sliding window update and $n$ for the sampling update. Indeed, at each step we remove one item and add one item to the current set. The mapping $\text{D}_t$ transforms $f_t$ at each step $t$, thus there exists an $f_t$ for the current set. As we compute $h(x)$ only on the data from the data set, only the distribution of the target variable changes. 
        
\begin{figure}
            \centering
            \includegraphics[width=0.47\linewidth]{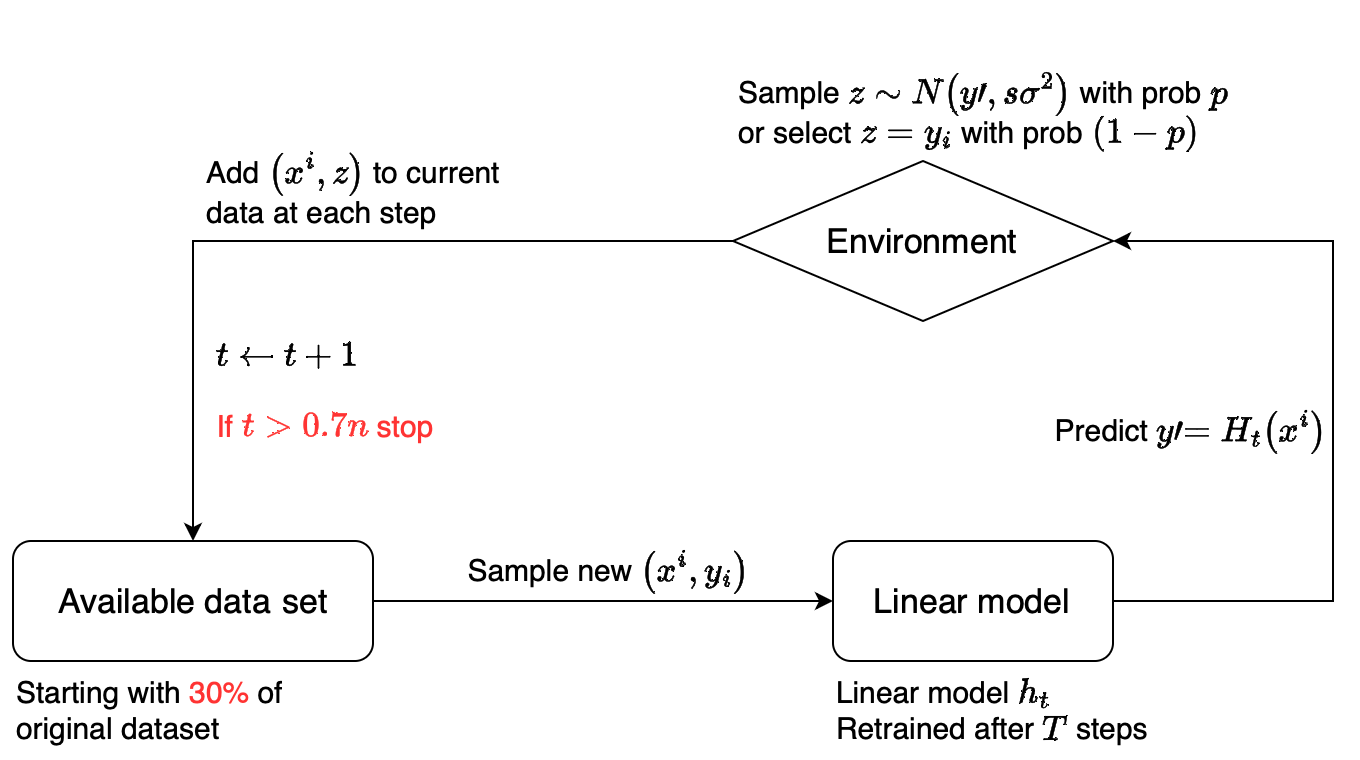}
            \includegraphics[width=0.47\linewidth]{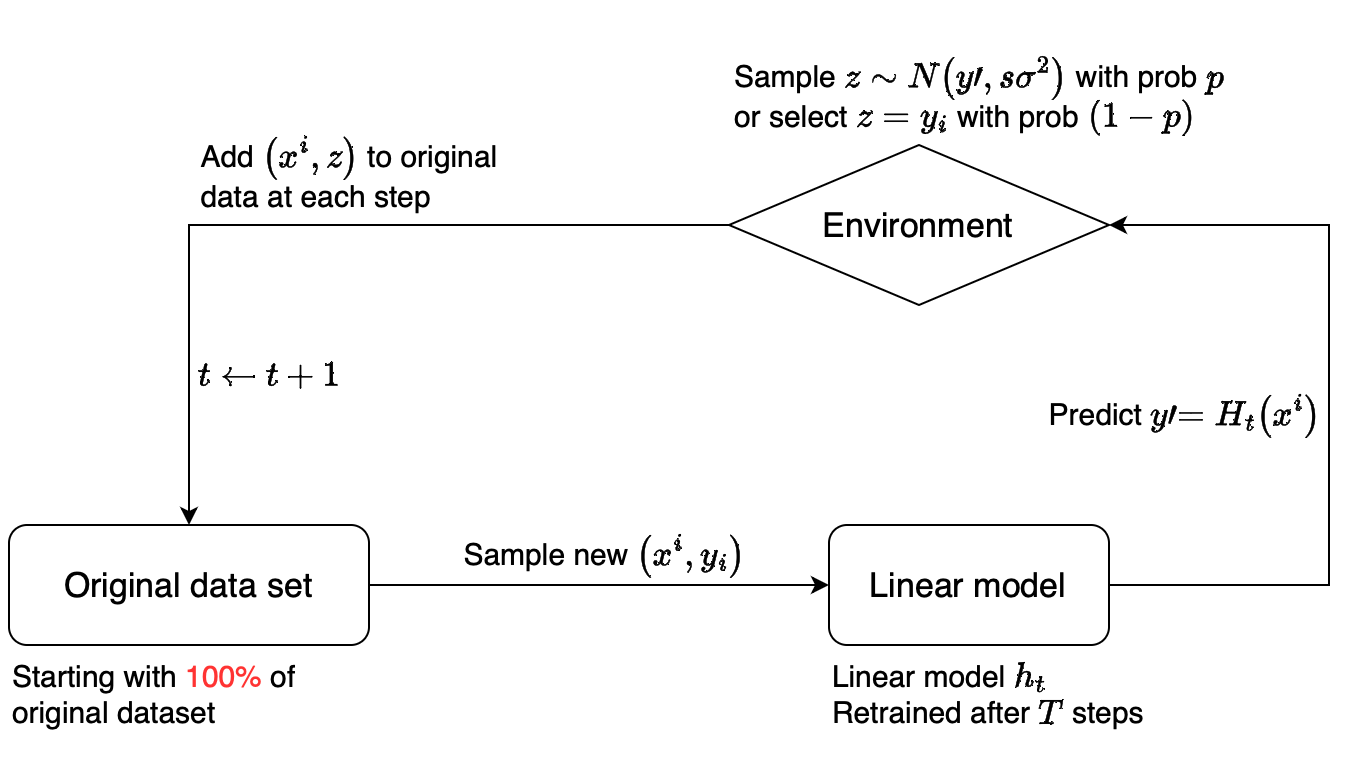}
            
            \caption{Two different experiments schemes. Sliding window update setup (left) and sampling update setup (right).}
            \label{ex_set}
        \end{figure}

        The schemes of the experiments are shown at Fig.~\ref{ex_set}. 

        Data collection takes about two hours for the experiment in Section~\ref{exp_1} and about an hour for experiments in Section~\ref{exp_2} and the rest. All experiments were conducted on a laptop with an four-core ARM-64 processor and reproduced on another four-core desktop PC without any hardware acceleration.

    \subsection{Analysis of Prediction Error} \label{exp_1}
    
        According to Theorem~\ref{delta} the two possible limiting distributions of $f_t$ are delta function $\delta(x)$ or zero distribution $\zeta(x)$. In this experiment we explore how the standard deviation of the prediction error changes over time. Based on Lemma~\ref{moments} and Theorem~\ref{delta}, we should observe two cases in this experiment: the tendency of the variance to zero (that is, the tendency of the PDFs to the delta function) or the tendency of the variance to infinity (that is, the tendency of the PDFs to the zero distribution). For this reason, at every $N$ steps we calculate the standard deviation of the difference $y_i - y_i'$. 
        
\begin{figure}
            \centering
            \includegraphics[width=0.49\linewidth]{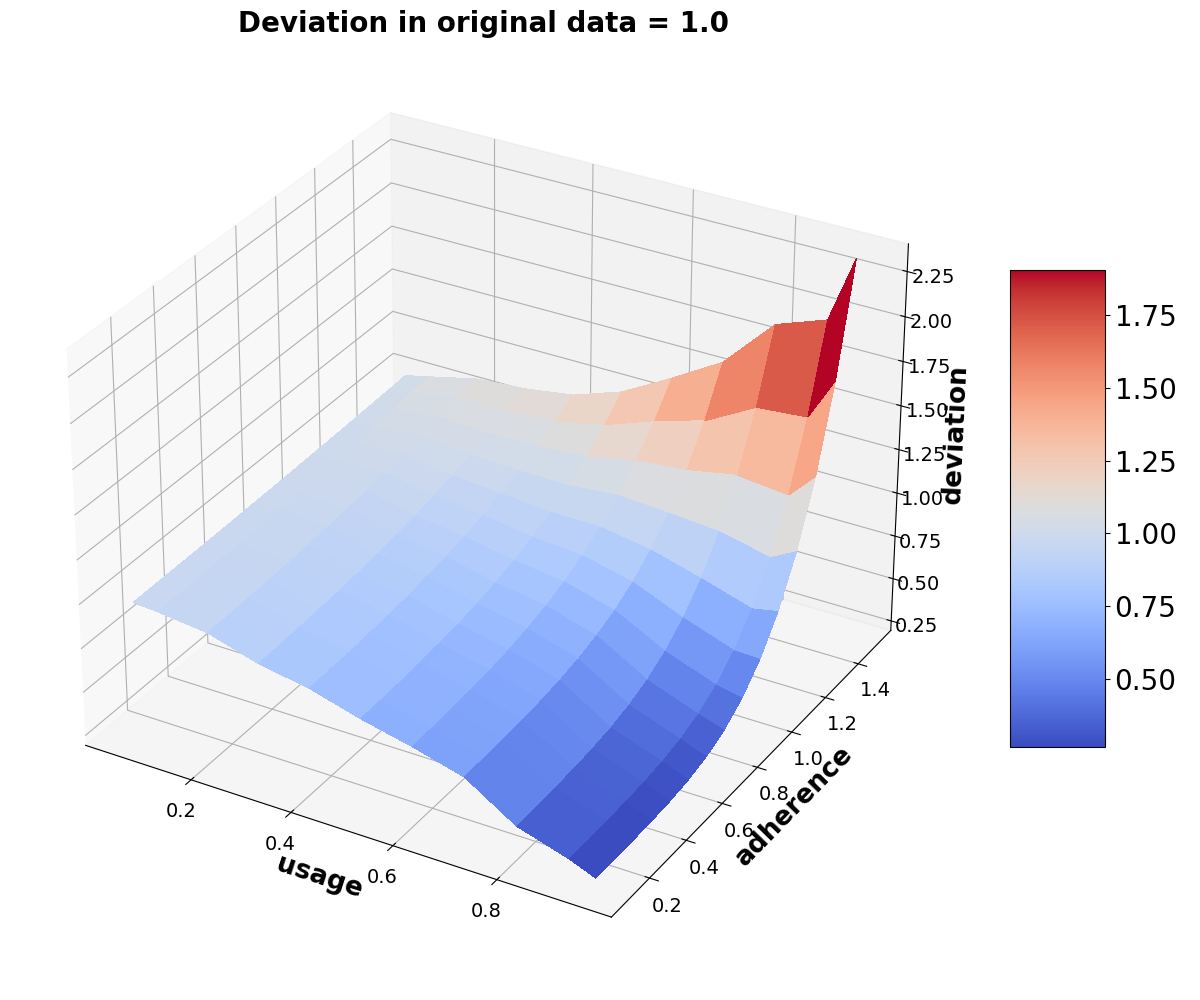}
            \includegraphics[width=0.49\linewidth]{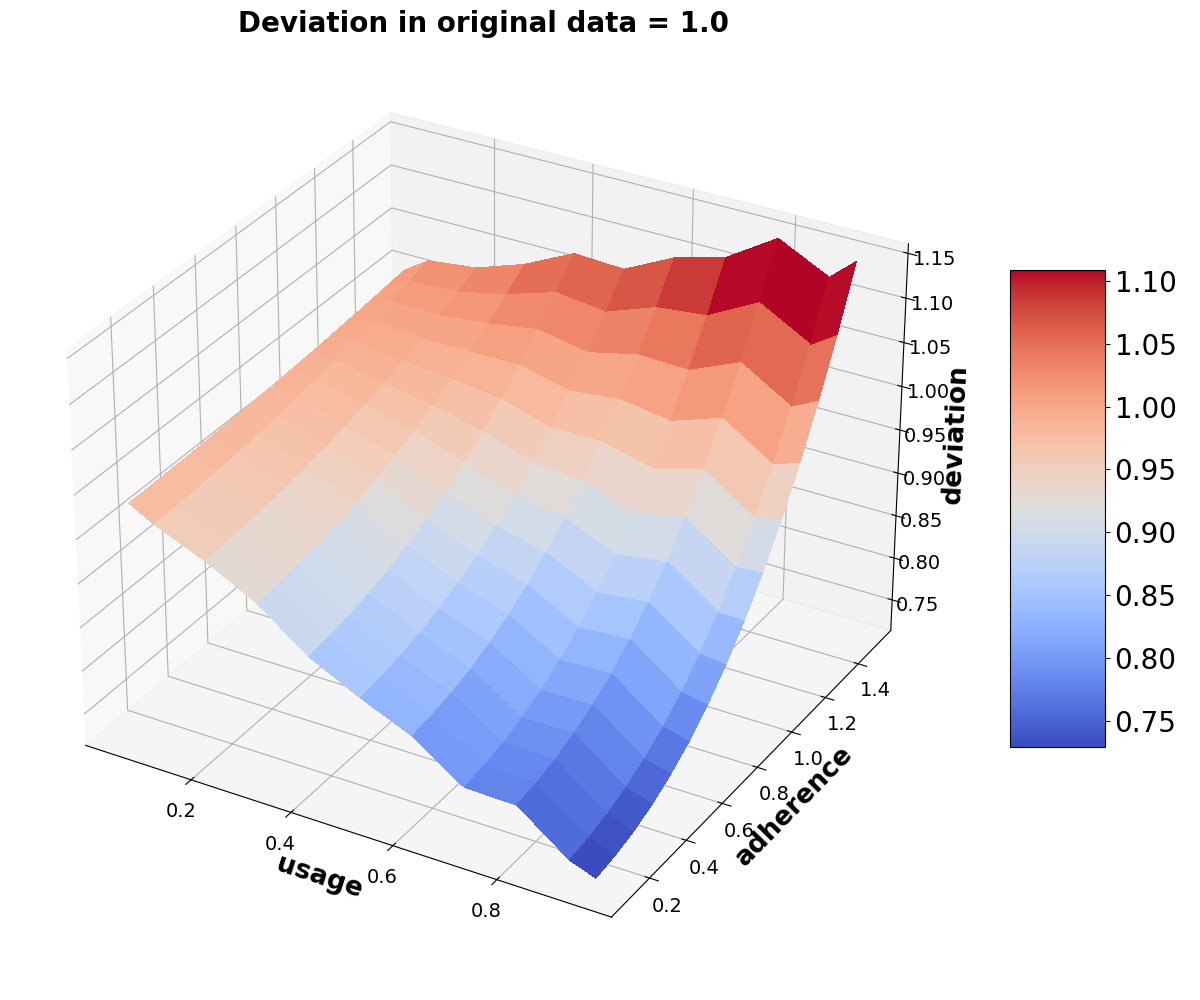}
            
            \caption{Change in the standard deviation of the model error for different usage and adherence. Sliding window setup (left), sampling update setup (right). The graph is almost everywhere either red or blue, hence Theorem~\ref{delta} is applicable in practice.}
            \label{3D}
        \end{figure}

        We run the experiment for different usage \citep{khritankov2023positive} values $p$---what portion of the predictions are seen by the users, the probability with which we include $(\mathbf{x}^i, z_i)$ in the current set, and adherence $s$---how closely the predictions of the model are followed by the users, the parameter to multiply $\sigma^2$ when sampling $z_i$.  We also vary the variance in the normal distribution of the noise $\epsilon$: $0.1, 0.3, 1, 3$ and $10$. Fig.~\ref{3D} shows only the result with noise variance $1$, the picture is similar for other values. More figures can be found in the experiment repository. In this experiment we consider SGD regression model and synthetic linear data set.

        As we can see from Fig.~\ref{3D}, as usage $p$ increases and adherence \citep{khritankov2023positive} $s$ decreases, the deviation decreases, because we start to add less noisy data to the current set.

        In almost all cases explored the response surface shown at Fig.~\ref{3D} is either of red or blue color, that is for most of the usage and adherence combinations there is a tendency towards either the delta or the zero function.

    \subsection{Normality Test for the Prediction Error} \label{exp_2}
        
        In this experiment we consider a linear regression model with squared loss and synthetic linear data set. Because the original training data is normally distributed and $\mathbf{y}$ is relates with $\textbf{X}$ linearly, it is feasible to check prediction error $y_i - y_i'$ for normality. In this experiment we explore how the probability density $f_t$ of the data changes as a result of the feedback loop.
        
        We use D’Agostino and Pearson’s test for normality and calculate the $p$-value. Here we take usage and adherence values from on the previous Experiment~\ref{exp_1}: $1.0$ and $0.0$, $0.1$ and $0.9$, $1$ and $3$ correspondingly. 

        As we can see from the measurements, the original data is normally distributed with $p$-value larger than the threshold. However, subsequently the $p$-value decreases and the normality of the sampled data at step $t$ breaks down. The $p$-values become very close to zero and mostly lower than the chosen $0.05$ threshold as time step progresses. We may conclude from the histogram that a possible reason for this is that distribution of $y_i - y_i'$ is a mixture of the two or more normal probability distributions. 

        Figures with $p$-value change over time Fig.~\ref{p_value} and example histograms Fig.~\ref{hist} are provided in Appendix~\ref{Supplementary_materials}.

    \subsection{Limit to Delta Function or Zero Distribution} \label{exp_3}
    
\begin{figure}
            \centering
            \includegraphics[width=0.32\linewidth]{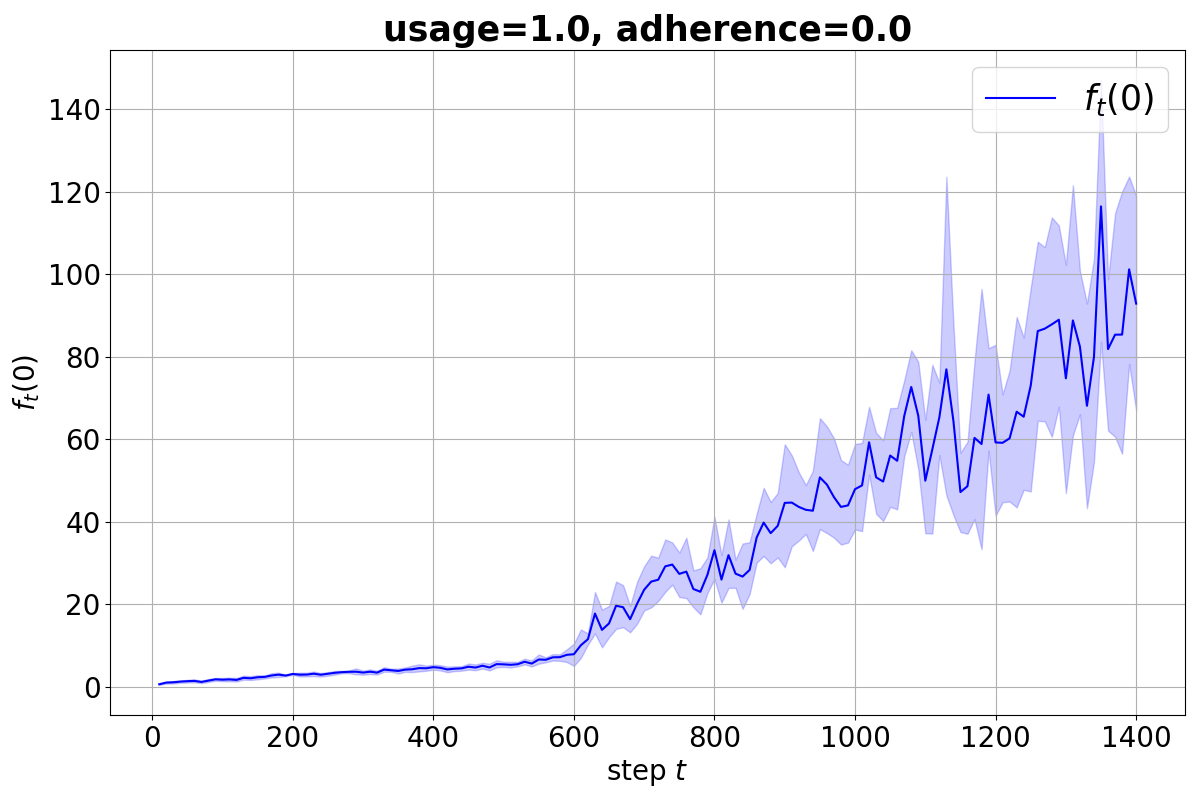}~
            \includegraphics[width=0.32\linewidth]{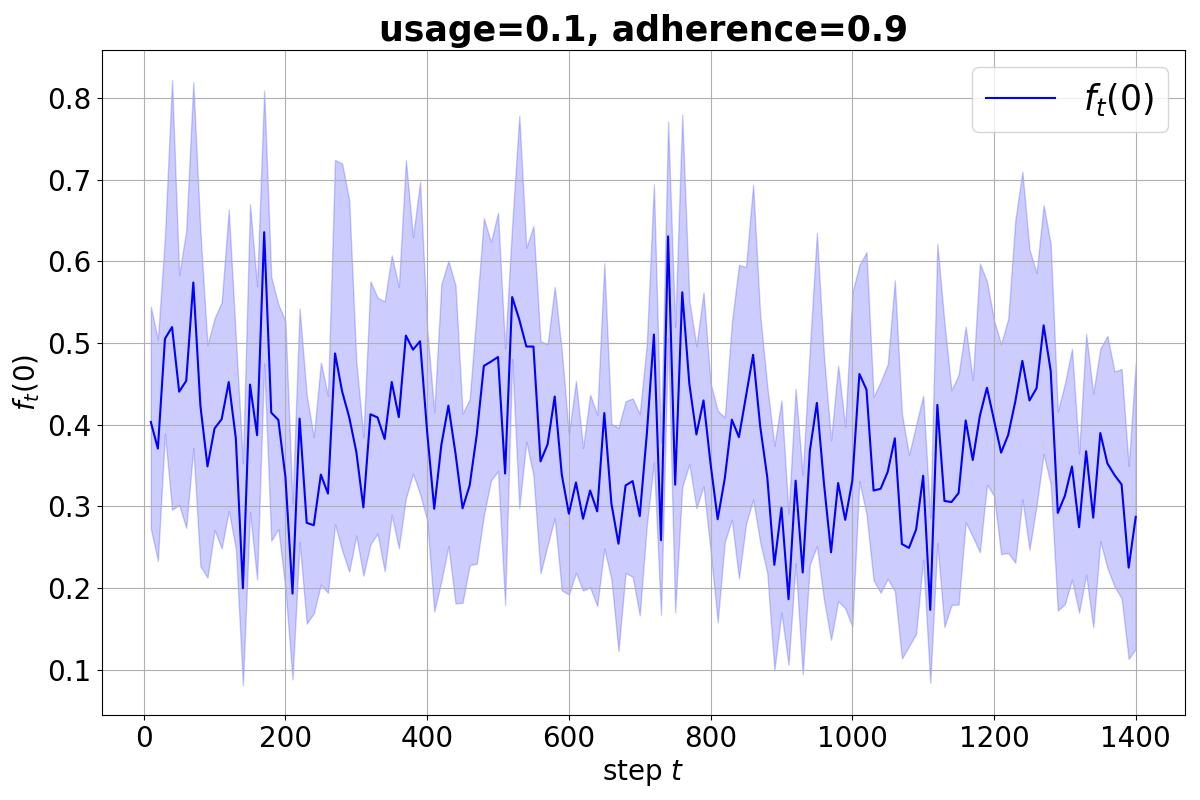}~
            \includegraphics[width=0.32\linewidth]{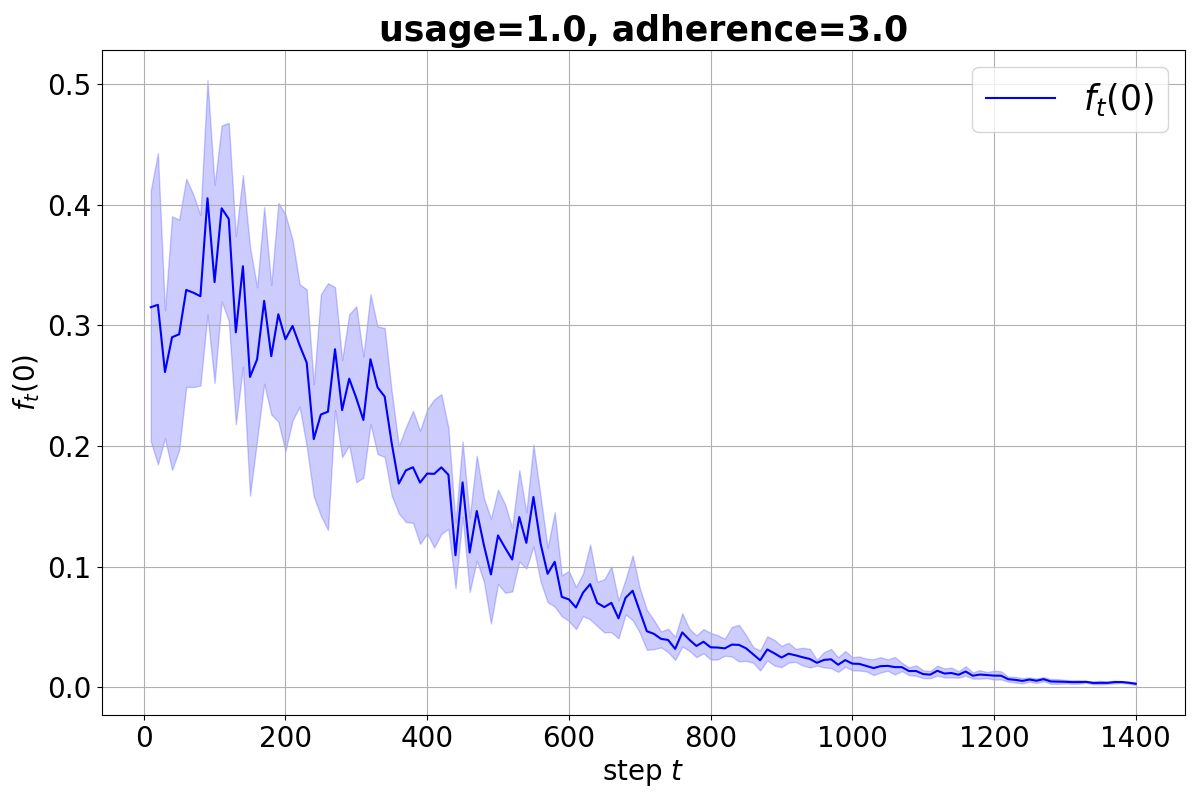}

            \includegraphics[width=0.32\linewidth]{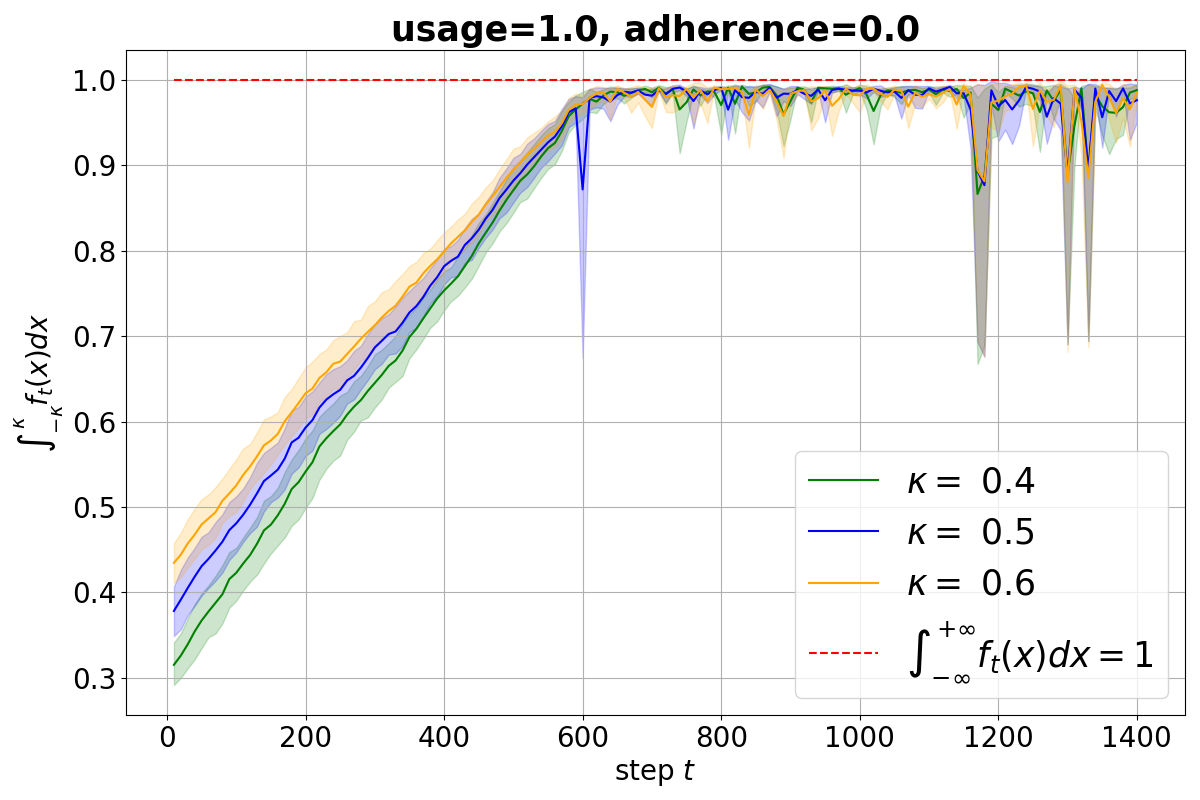}~
            \includegraphics[width=0.32\linewidth]{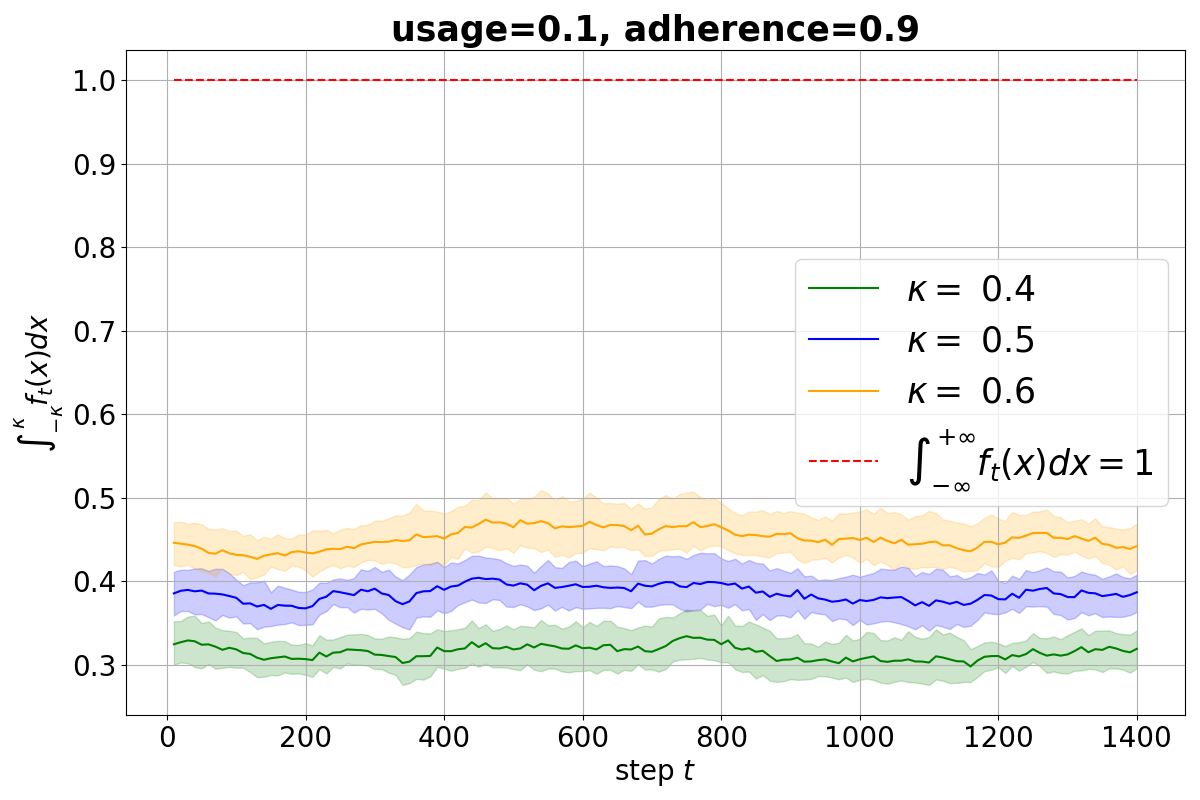}~
            \includegraphics[width=0.32\linewidth]{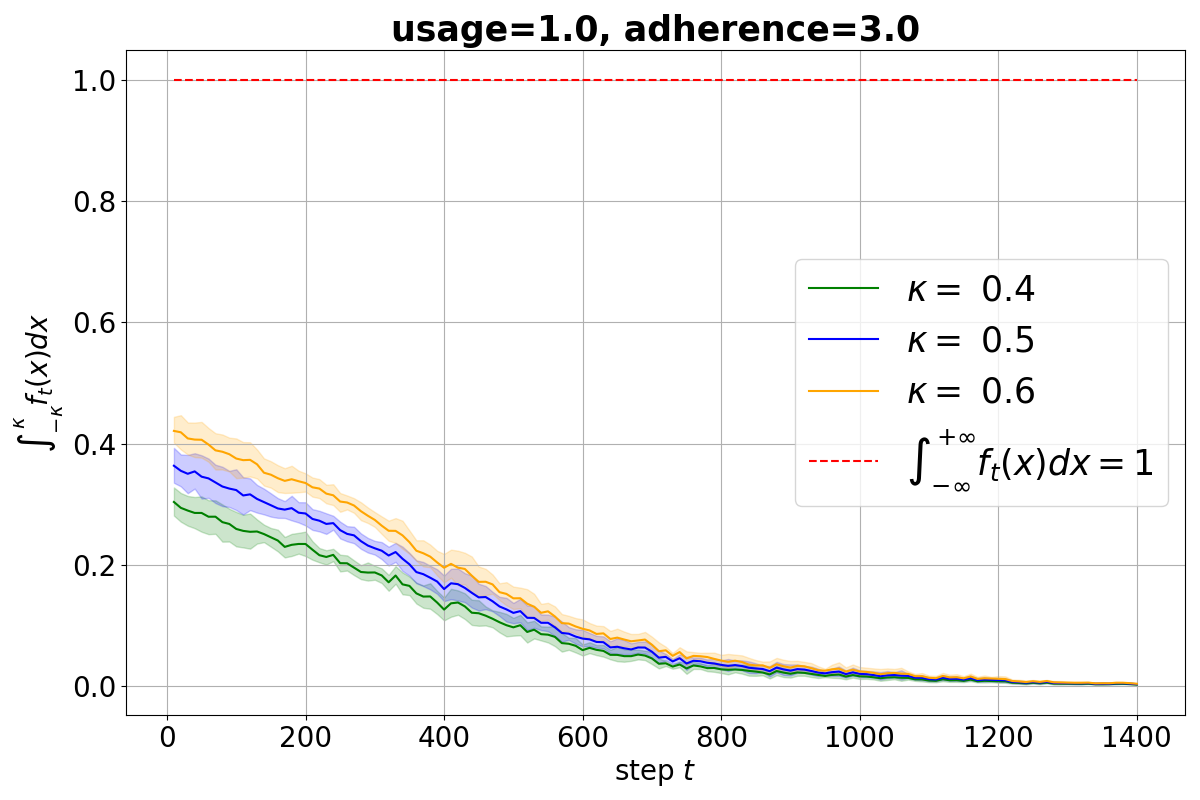}
            
            \caption{Counting $f_t(0)$ and $\int_{-\kappa}^{\kappa}f_t(x)dx$ for sliding window setup for SGD regression model on synthetic linear data set. We consider such parameters: usage, adherence = $1$, $0$ (left); $0.1$, $0.9$ (middle); $1$, $3$ (right). In this picture, we can see the entire limit set of the system \eqref{system} from Theorem~\ref{delta}.}
            \label{delta_loop}
        \end{figure}

\begin{figure}
            \centering
            \includegraphics[width=0.32\linewidth]{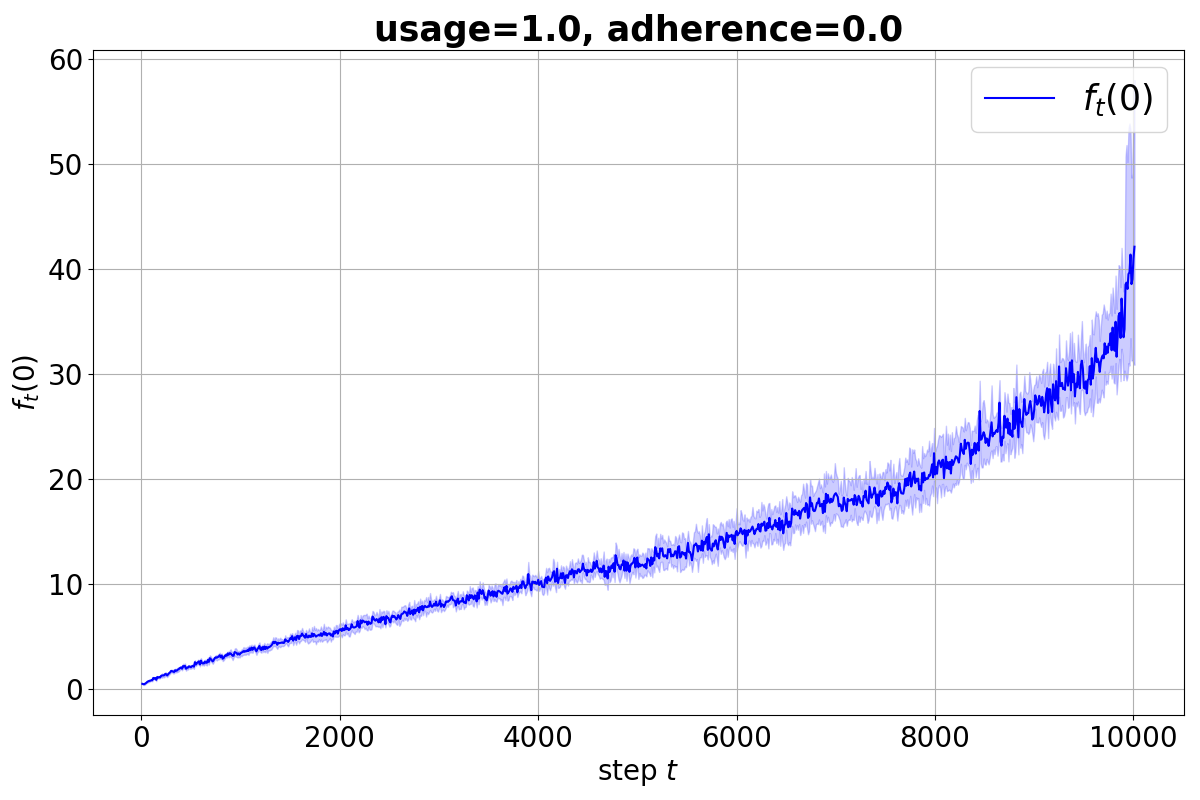}~
            \includegraphics[width=0.32\linewidth]{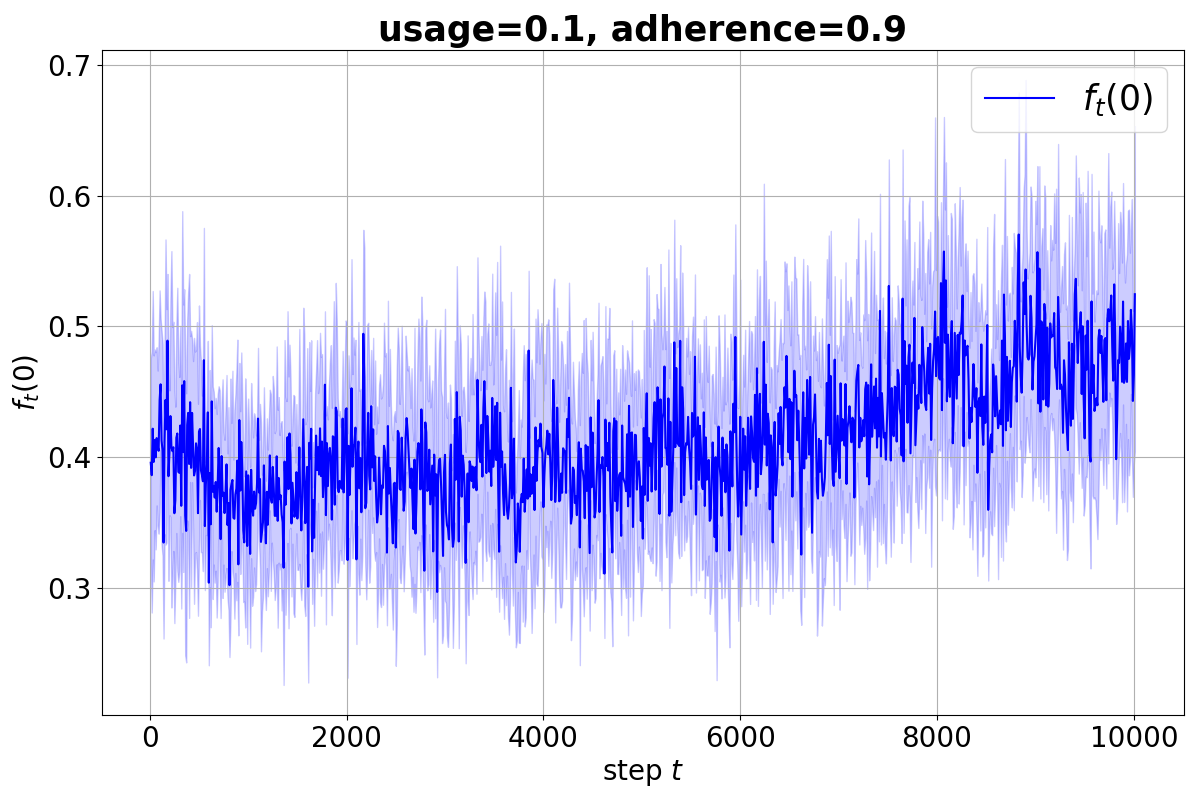}~
            \includegraphics[width=0.32\linewidth]{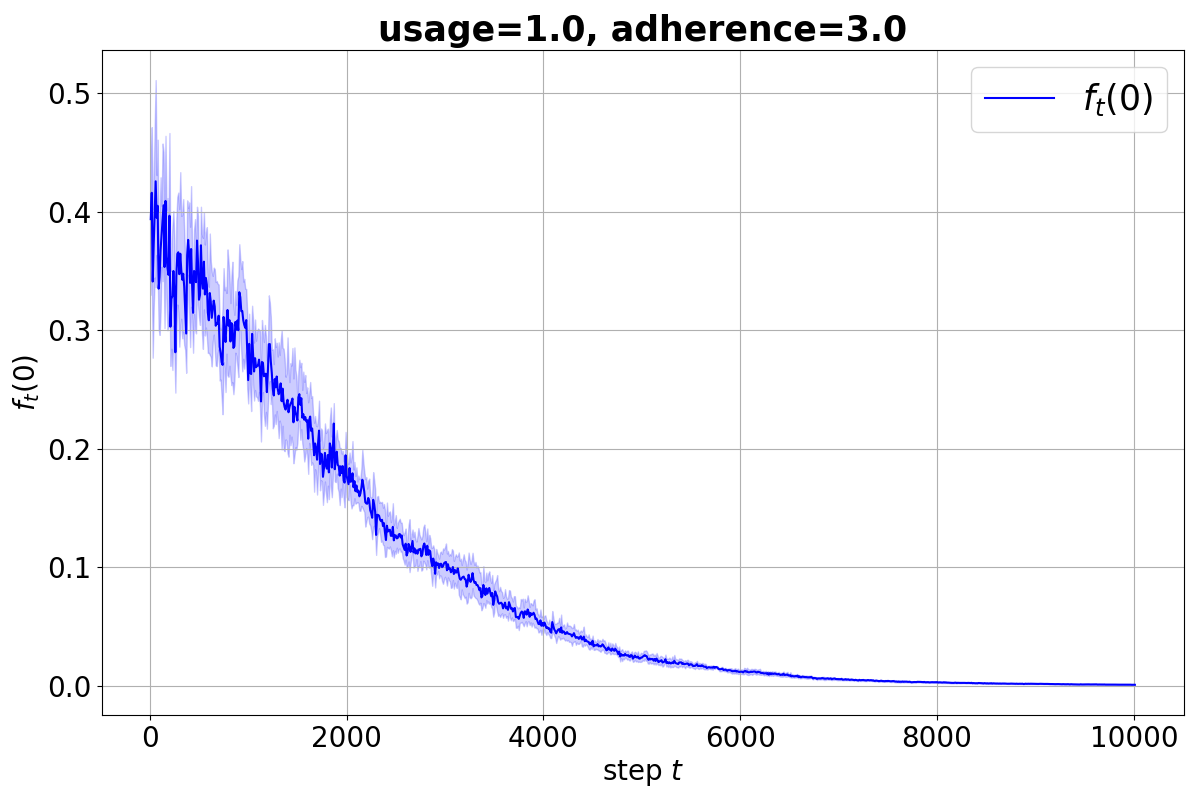}

            \includegraphics[width=0.32\linewidth]{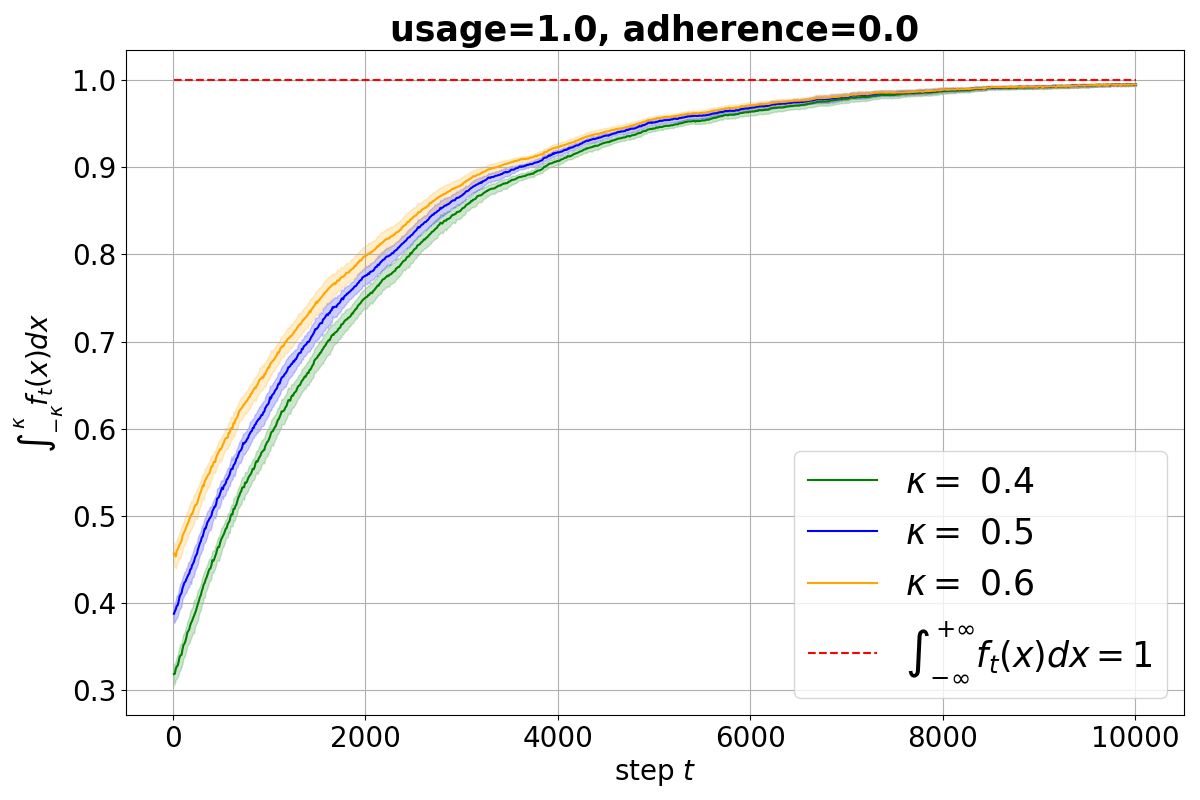}~
            \includegraphics[width=0.32\linewidth]{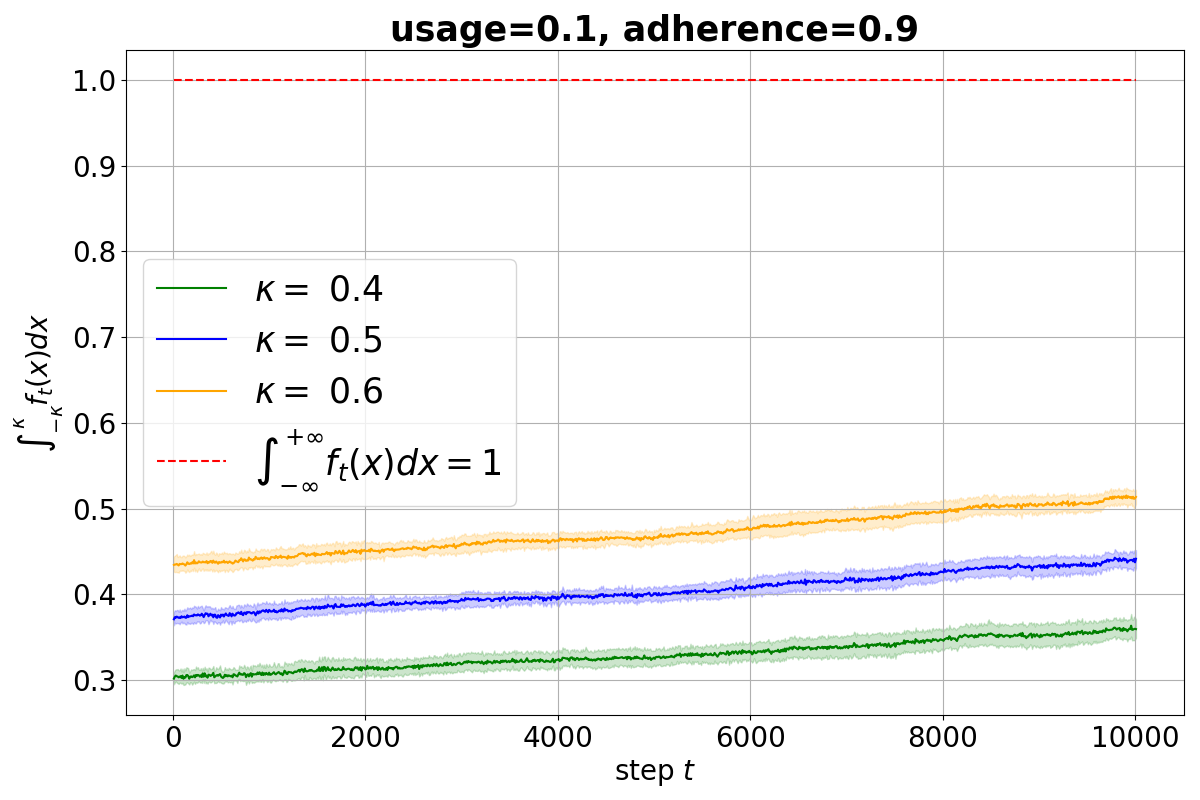}~
            \includegraphics[width=0.32\linewidth]{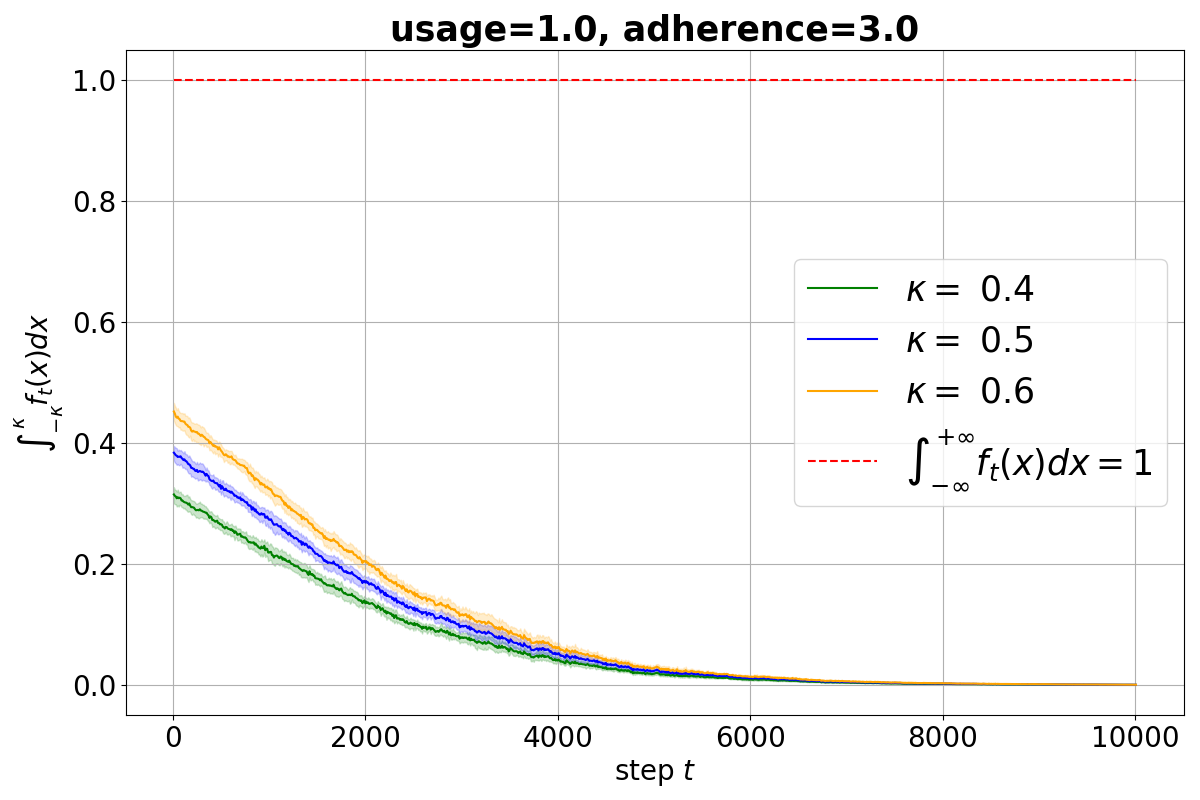}
            
            \caption{Counting $f_t(0)$ and $\int_{-\kappa}^{\kappa}f_t(x)dx$ for sampling update setup for SGD regression model on synthetic linear data set. We consider such parameters: usage, adherence = $1$, $0$ (left); $0.1$, $0.9$ (middle); $1$, $3$ (right). In this picture, we can see the entire limit set of the system \eqref{system} from Theorem~\ref{delta}.}
            \label{delta_sample}
        \end{figure}

        In this experiment we directly test the predictions of Theorem~\ref{delta}, that is we measure $f_t(0)$ and $\int_{-\kappa}^{\kappa}f_t(x)dx$, where $\kappa > 0$ is sufficiently small. We estimate $f_t(0)$ using a more numerically stable linear interpolation heuristic method from the SciPy library \citep{virtanen2020SciPy} that gives results close to theoretically grounded density estimation  \citep{silverman1986density} method.
        The data collected in this experiment are shown at Fig.~\ref{delta_loop} and Fig.~\ref{delta_sample}. In this experiment we consider SGD regression model on synthetic linear data set and Ridge model with no regularization on Friedman data set. 

        As we can see, if usage $p = 1$ and adherence $s = 3$, the limiting probability density of $\text{D}_{\overline{1, t}}(f_0)$, that is the probability density of $y_i - y_i'$, is zero distribution $\zeta(x)$. This corresponds to the fact that $\psi_t$  tends to zero and $\int_{-\kappa}^{\kappa}f_t(x)\,dx$ tends to zero.
        When usage $p = 1$ and adherence $s = 0$ we observe a tendency to the delta function $\delta(x)$, that is $\psi_t$ tends to positive infinity and $\int_{-\kappa}^{\kappa}f_t(x)\,dx$ tends to one. 
        If usage $p = 0.1$ and adherence $s = 0.9$, the probability density of $y_i - y_i'$ remains almost the same, that is $\psi_t$ tends to some constant $c \in (0; +\infty)$. 
        
        Therefore, we can conclude that the observed behavior does not contradict the claims of Theorem~\ref{delta}.

        Additional figures for Ridge model without regularization on Friedman data set are provided in Appendix~\ref{Supplementary_materials}. Since the form of these plots is the same as in Fig.~\ref{delta_loop}~and~\ref{delta_sample}, we can conclude that the results of Theorem~\ref{delta} may generalize to different data sets and machine learning models.

    \subsection{Autonomy check} \label{exp_4}

        We test the claim \eqref{cond_semigroup} of Theorem~ \ref{semigroup}. According to the discussion of Theorem~ \ref{semigroup}, for the sequence $\psi_t$ to satisfy the condition \eqref{cond_semigroup}, it must be a power sequence. 

\begin{figure}
            \centering
            \includegraphics[width=0.32\linewidth]{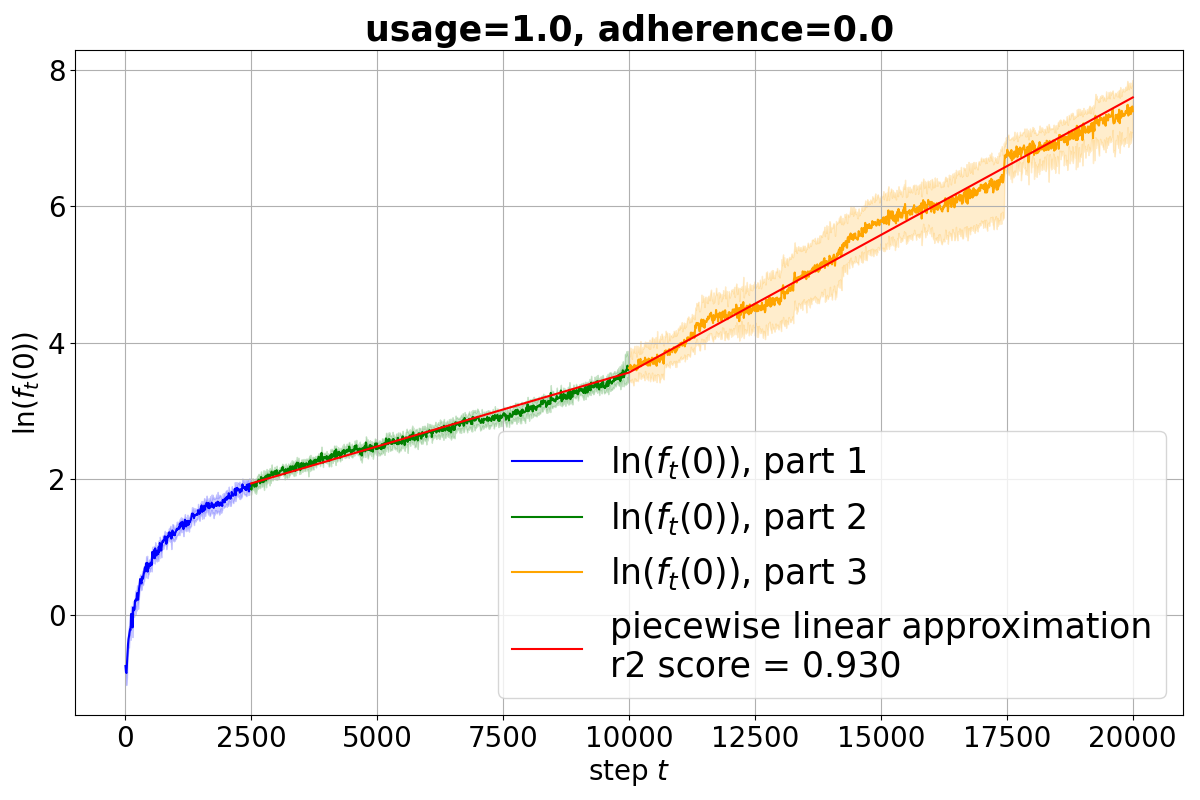}
            \includegraphics[width=0.32\linewidth]{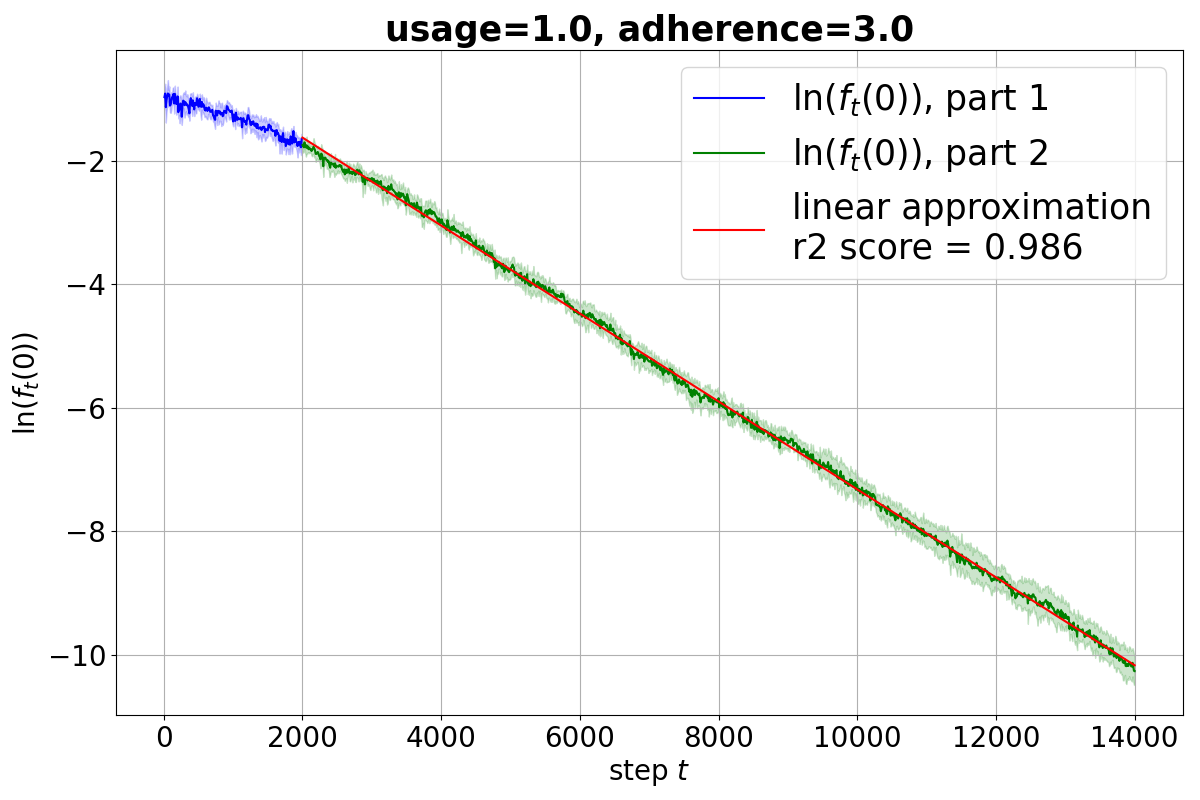}
            \includegraphics[width=0.32\linewidth]{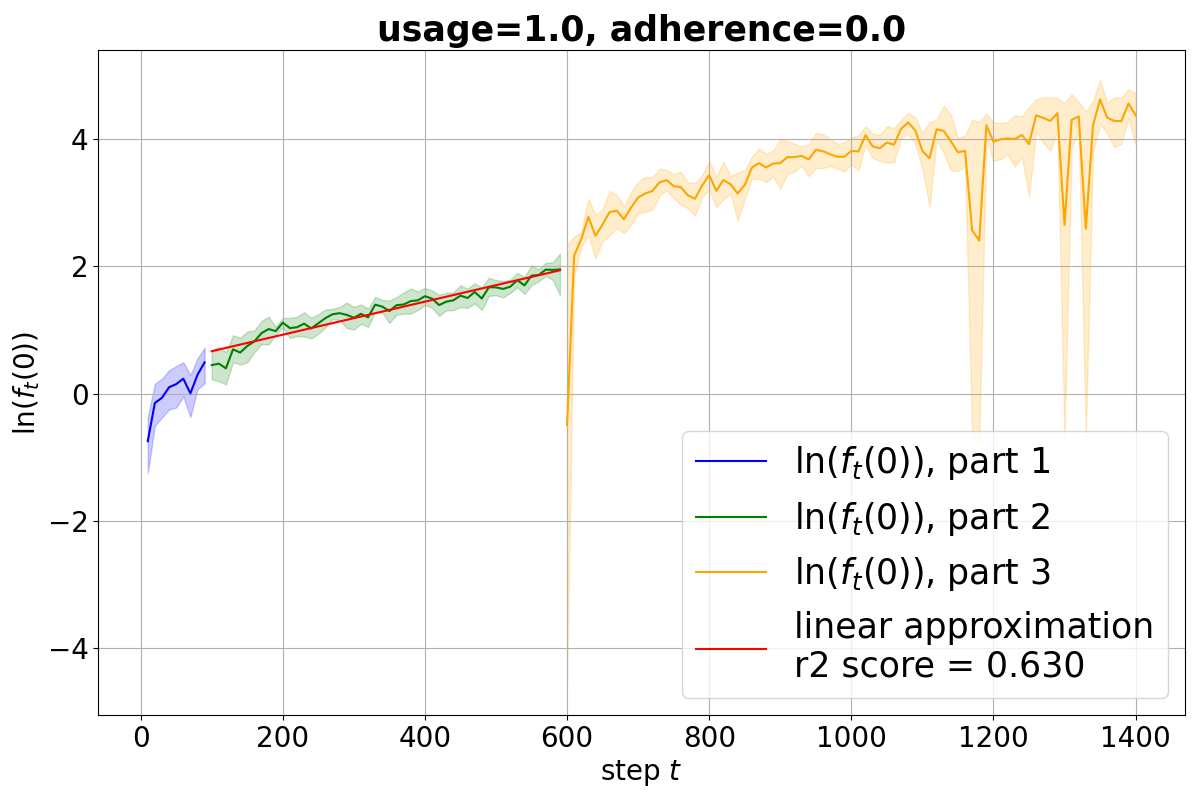}
            
            \caption{Testing two designs for autonomy. Sampling update setup (left and middle) and sliding window setup (right). Results for SGD regression model on the synthetic linear data set. 
            %The sampling update setup is autonomous since plot is linear, and sliding window setup is not autonomous.
            }
            \label{fig_exp_4_1}
        \end{figure}

\begin{figure}
            \centering
            \includegraphics[width=0.32\linewidth]{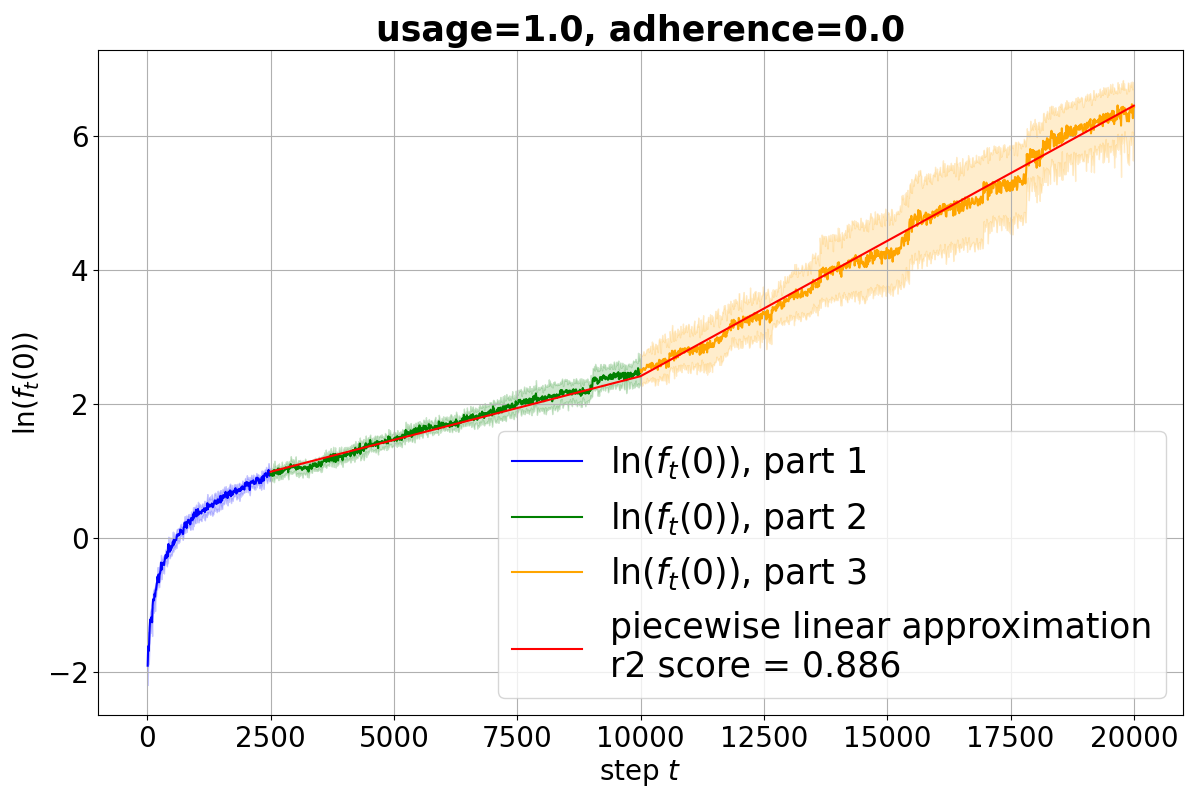}
            \includegraphics[width=0.32\linewidth]{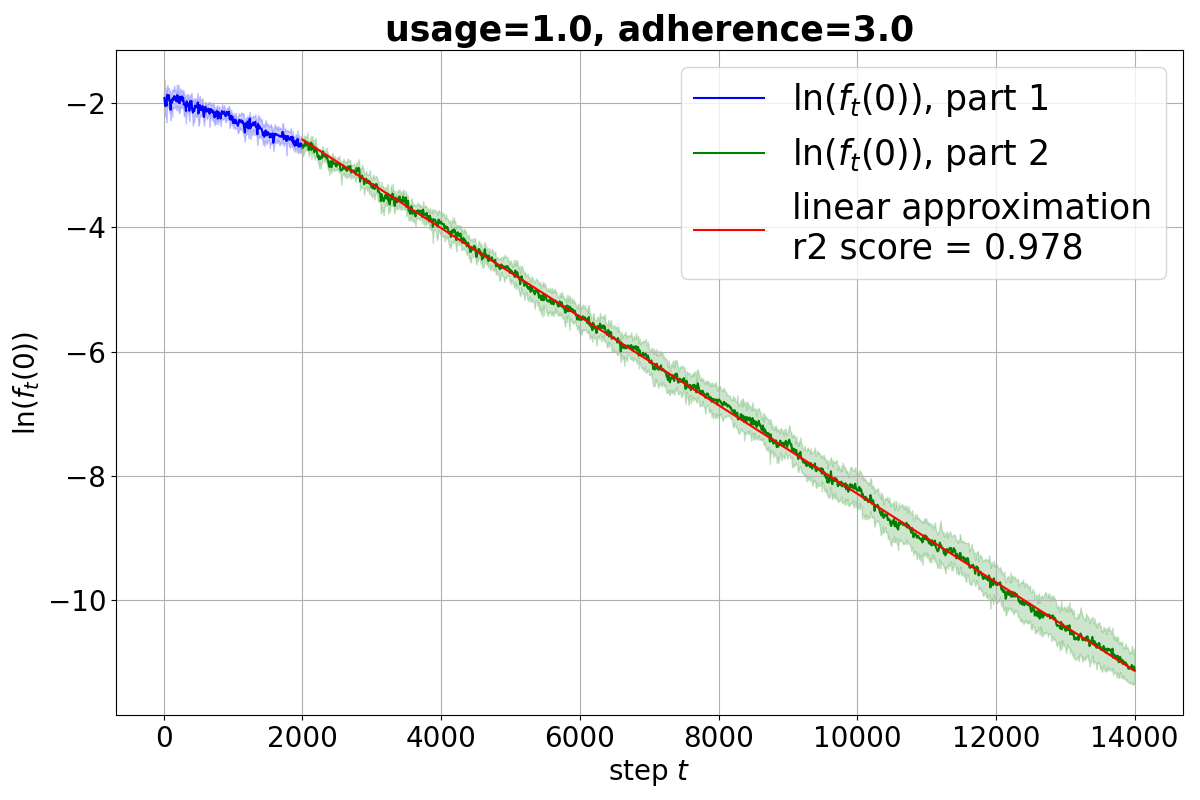}
            \includegraphics[width=0.32\linewidth]{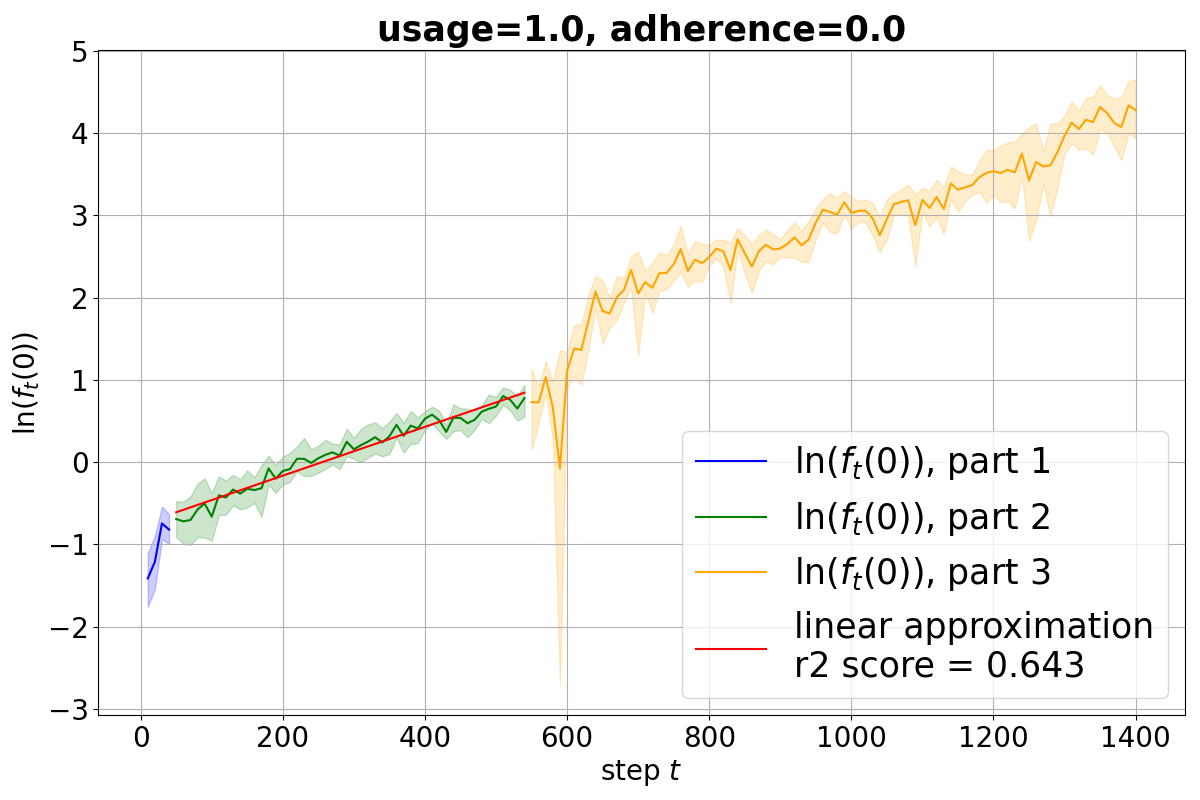}
            
            \caption{Testing two designs for autonomy. Sampling update setup (left and middle) and sliding window setup (right). Results for SGD regression model on the synthetic linear data set. 
            %The sampling update setup is autonomous since plot is linear, and sliding window setup is not autonomous.
            }
            \label{fig_exp_4_2}
        \end{figure}

\begin{figure}
            \centering
            \includegraphics[width=0.32\linewidth]{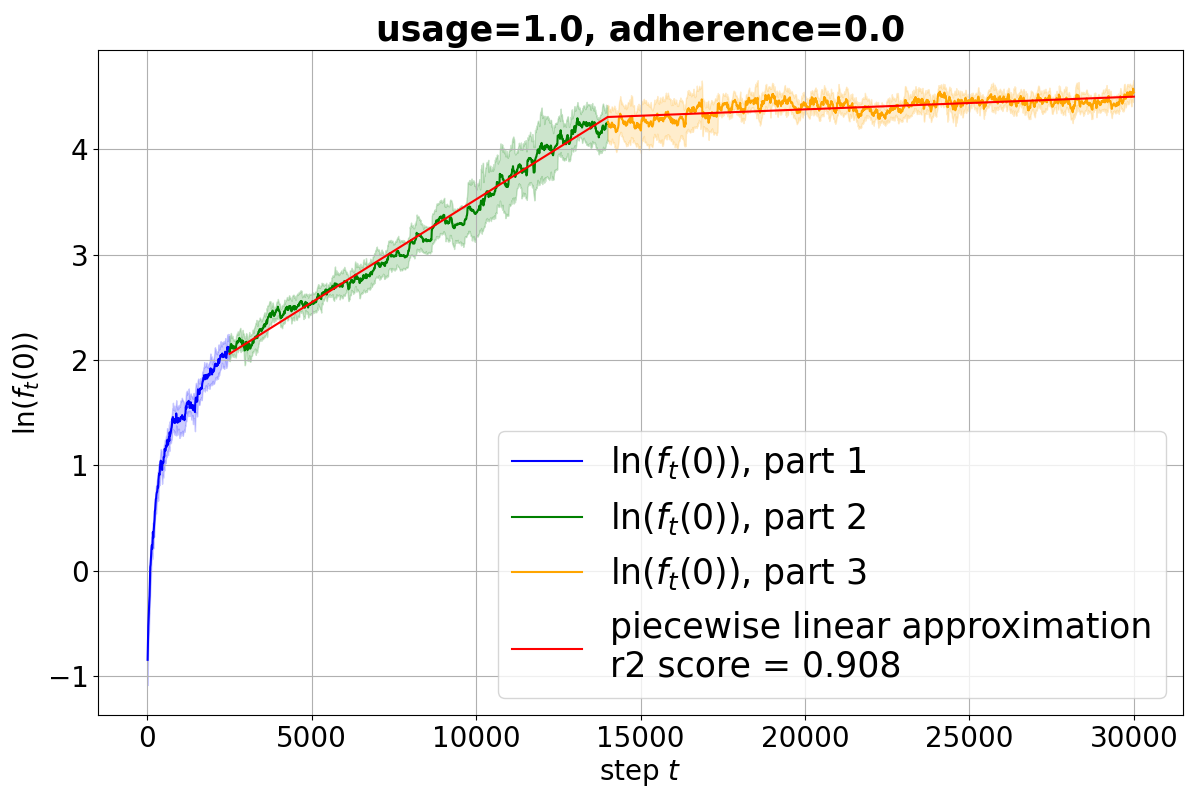}
            \includegraphics[width=0.32\linewidth]{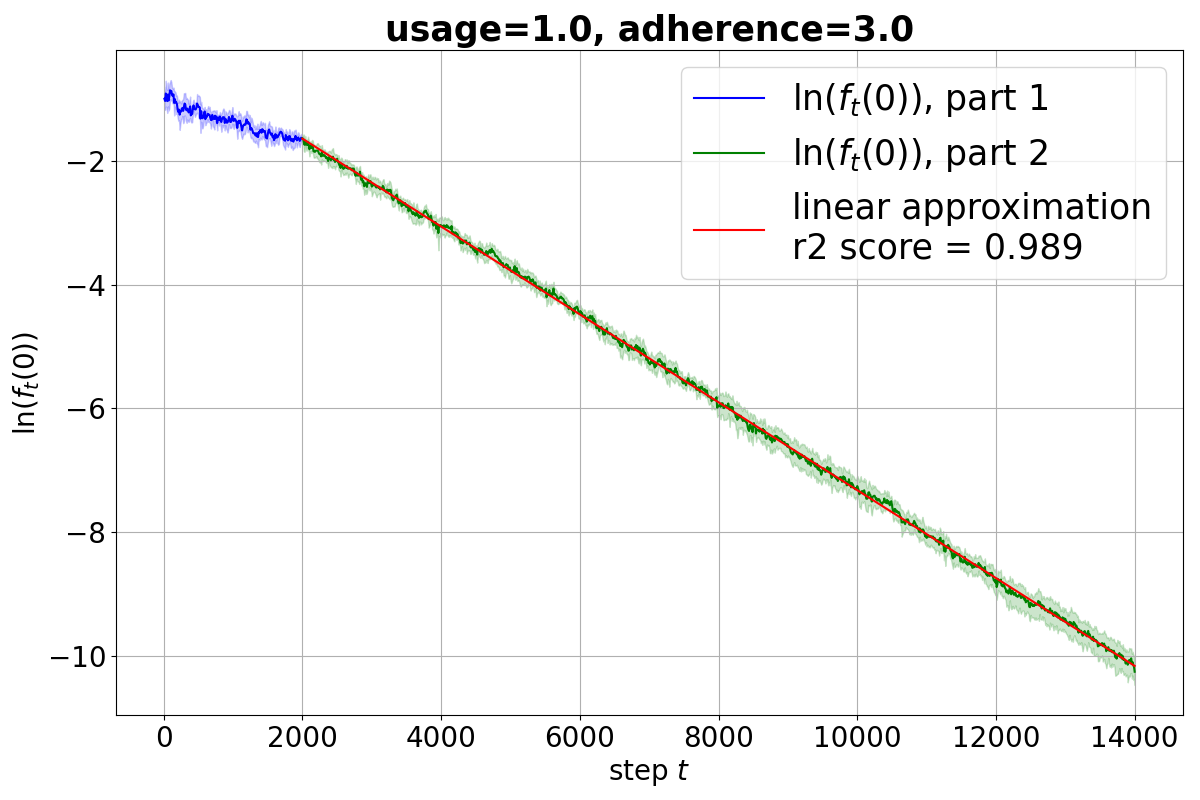}
            \includegraphics[width=0.32\linewidth]{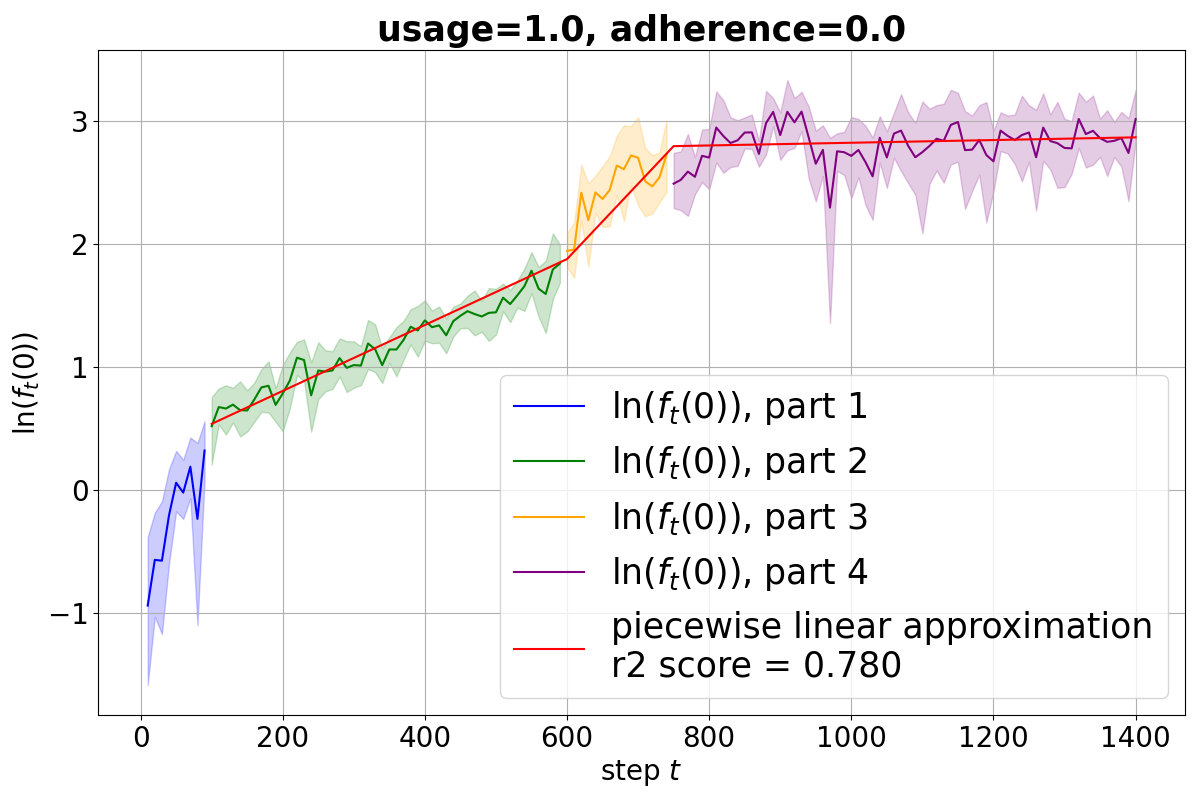}
            
            \caption{Testing two designs for autonomy. Sampling update setup (left and middle) and sliding window setup (right). Results for RidgeCV regression model on the synthetic linear data set. 
            %The sampling update setup is autonomous since plot is linear, and sliding window setup is not autonomous.
            }
            \label{fig_exp_4_3}
        \end{figure}

\begin{figure}
            \centering
            \includegraphics[width=0.32\linewidth]{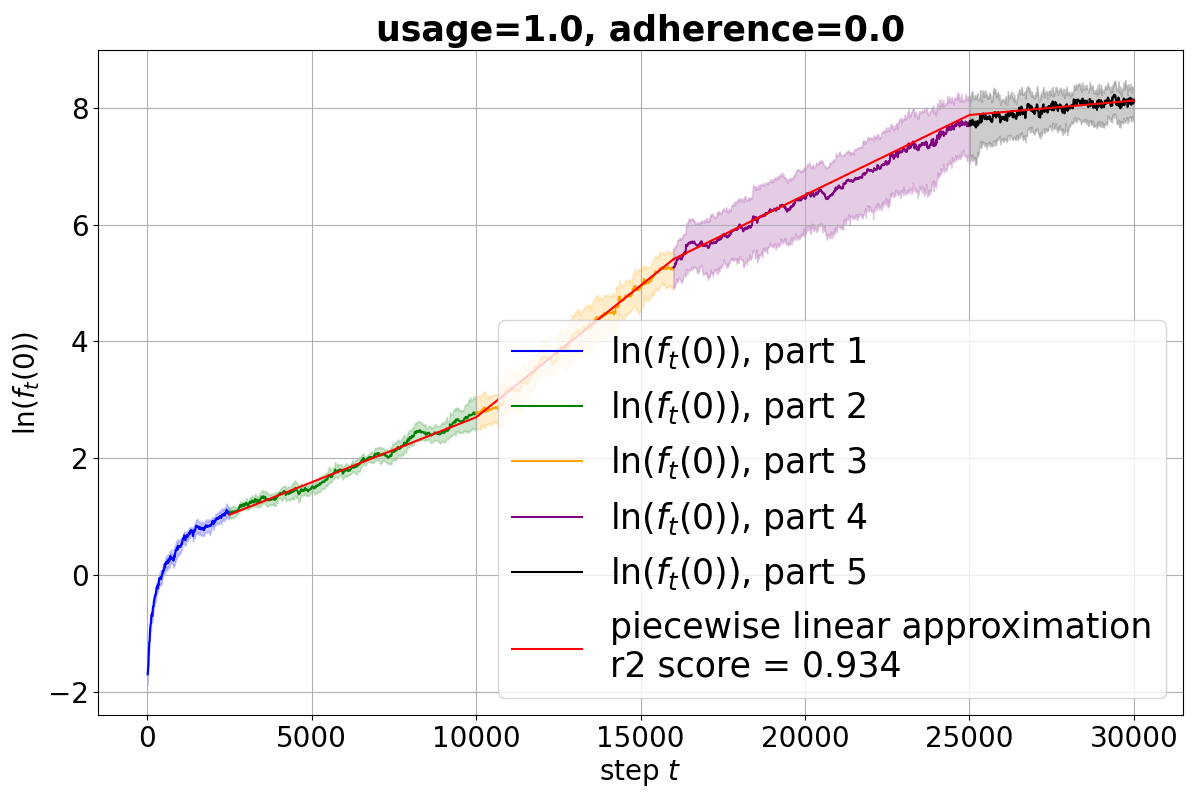}
            \includegraphics[width=0.32\linewidth]{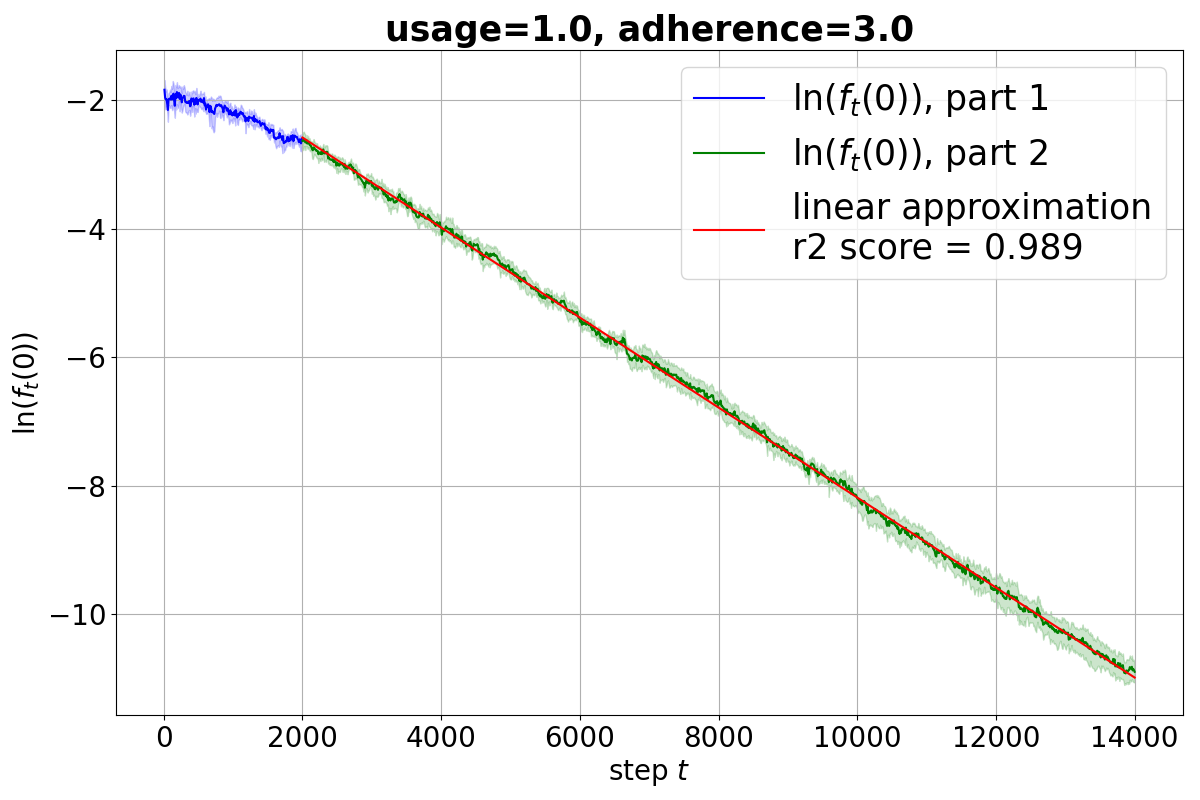}
            \includegraphics[width=0.32\linewidth]{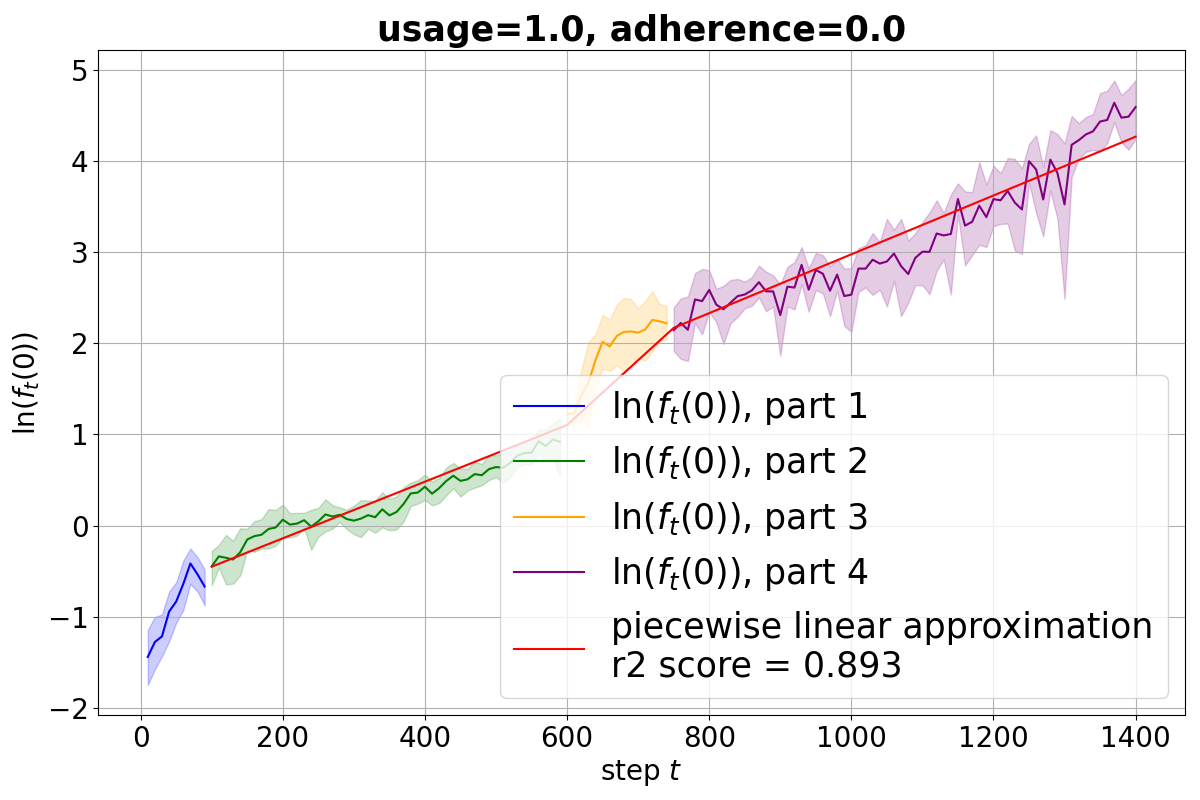}
            
            \caption{Testing two designs for autonomy. Sampling update setup (left and middle) and sliding window setup (right). Results for RidgeCV regression model on the Friedman data set. 
            %The sampling update setup is autonomous since plot is linear, and sliding window setup is not autonomous.
            }
            \label{fig_exp_4_4}
        \end{figure}

        In this experiment, we assess the autonomy of the exemplary machine learning systems in the sliding window update and sampling update settings. The former is not autonomous and the latter is an autonomous system by construction, because the evolution operator $\text{D}_t$, that is the learning algorithm and feedback procedure, does not depend on time step, $\text{D}_t = \text{D}$.
        We plot $\ln(f_t(0))$ and check if there is a straight line when the system is autonomous, or not a straight line for a non-autonomous system. For clarity, we fit a linear model to the measurements and show the $r^2$ score---a statistical measure of how well the line matches the data. The observations and the lines are shown at the Fig.~\ref{fig_exp_4_1}-\ref{fig_exp_4_4}. We provide the results for SGD and RidgeCV regression models on synthetic linear and Friedman data sets.

        As you can see, in case of the sliding window update we obtain a poor fit on all models and data sets, so the system is not autonomous. On the figure there is a notable change in around $t = 600$, this corresponds to the point where the proportion of reused predictions in the data stabilizes at usage rate $p$ does not change any further. The sampling update setup in case of usage $p = 1$ and adherence $s = 3$ is autonomous on all models and data sets, since there is a good fit. In case of usage $p = 1$ and adherence $s = 0$ we observe two linear segments, except in the case of RidgeCV on the Friedman data set. While our hypothesis states there should be just one linear segment, we may see two regions of autonomy. We contemplate that this may be due to the fact that at $t \approx 10000$ all the items in the original i.i.d. data set are replaced by the model predictions.
        
        In order to solve the aforementioned regression problem to obtain the fit, we use the robust Huber regressor. We also perform a Breusch-Pagan Lagrange multiplier test for homoscedasticity \citep{d1971omnibus}. In the sampling update case we get a $p$-value of $\approx 10^{-18}$ and in the sliding window case the $p$-value equals to $0.001$. Therefore, we assume that the residuals in all cases are homoscedastic and the fit is justified.

        With this we may conclude that the claims of Theorem~\ref{semigroup} does not contradict to the observations in this experiment. The hypotheses about the autonomy of the two systems are partially confirmed.

    \subsection{Decreasing Moments} \label{exp_5}
        
        In this experiment we compare the predictions of Lemma~\ref{moments} with the observations. In the sliding window update setting we check the third term of this Lemma: $\|\{\nu_k^t\}_{k=1}^{\infty}\|_1 = \sum_{k=1}^{+\infty} |\nu_k^t| \underset{t \to +\infty}{\longrightarrow} 0$. For computational efficiency reasons, we report only $\sum_{k=1}^{N} |\nu_k^t|$ for $N=300$. 
        
\begin{figure}
            \centering
            \includegraphics[width=0.49\linewidth]{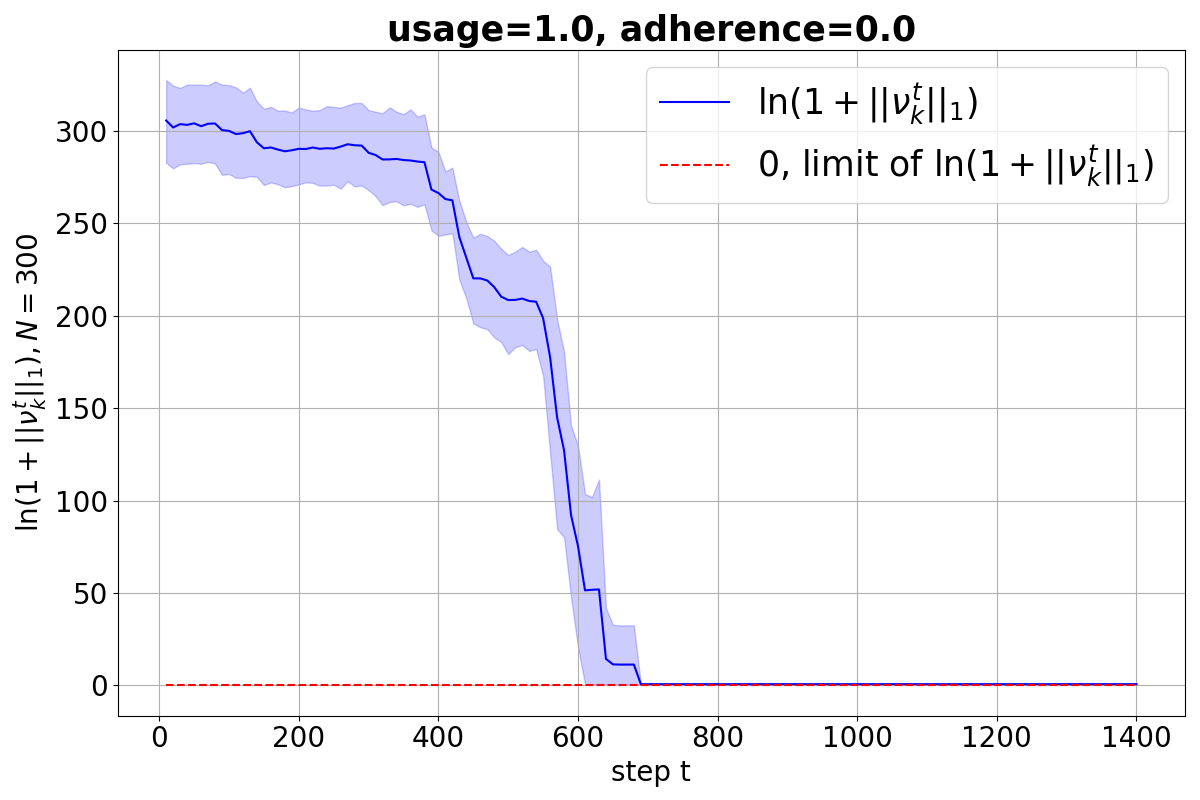}
            \includegraphics[width=0.49\linewidth]{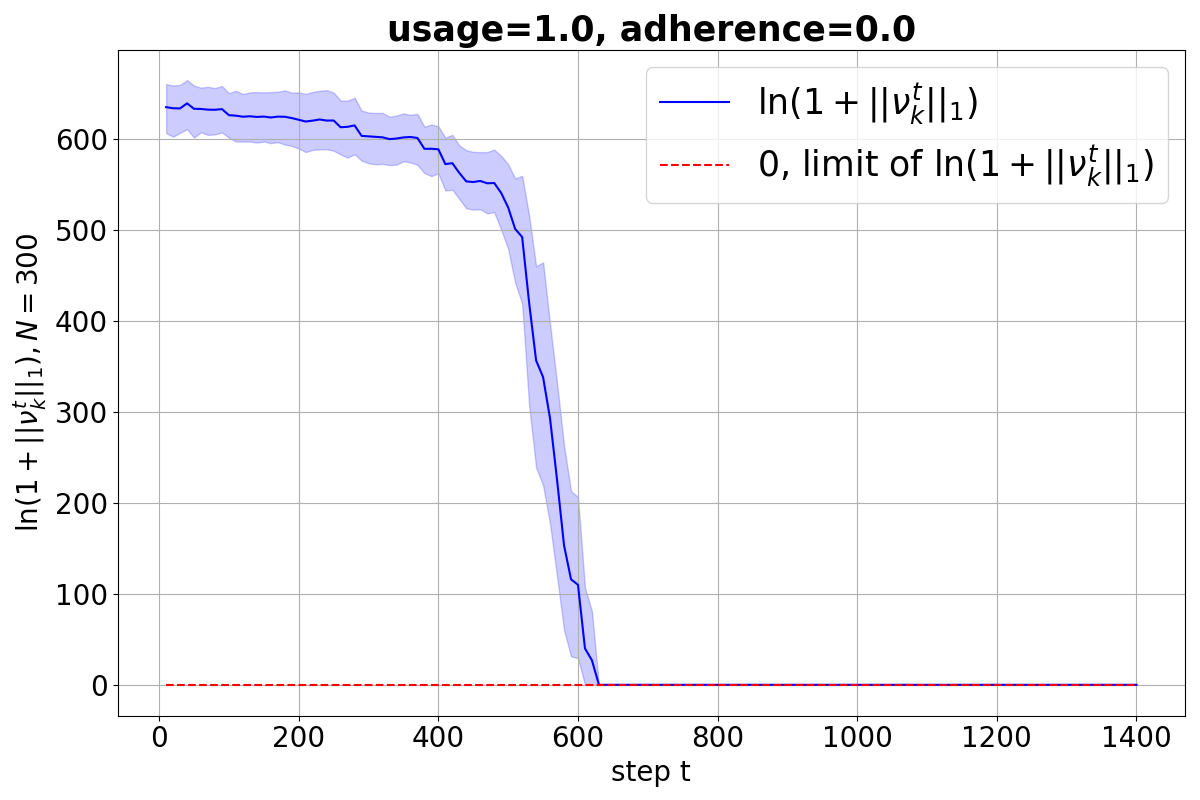}
            
            \caption{Measurement $\|\{\nu_k^t\}_{k=1}^{\infty}\|_1$ for sliding window setup. Synthetic linear data set (left) and Friedman data set (right). As we can see, the second statement of Lemma~\ref{moments} is satisfied.}
            \label{fig_exp_5_1}
        \end{figure}

\begin{figure}
            \centering
            \includegraphics[width=0.49\linewidth]{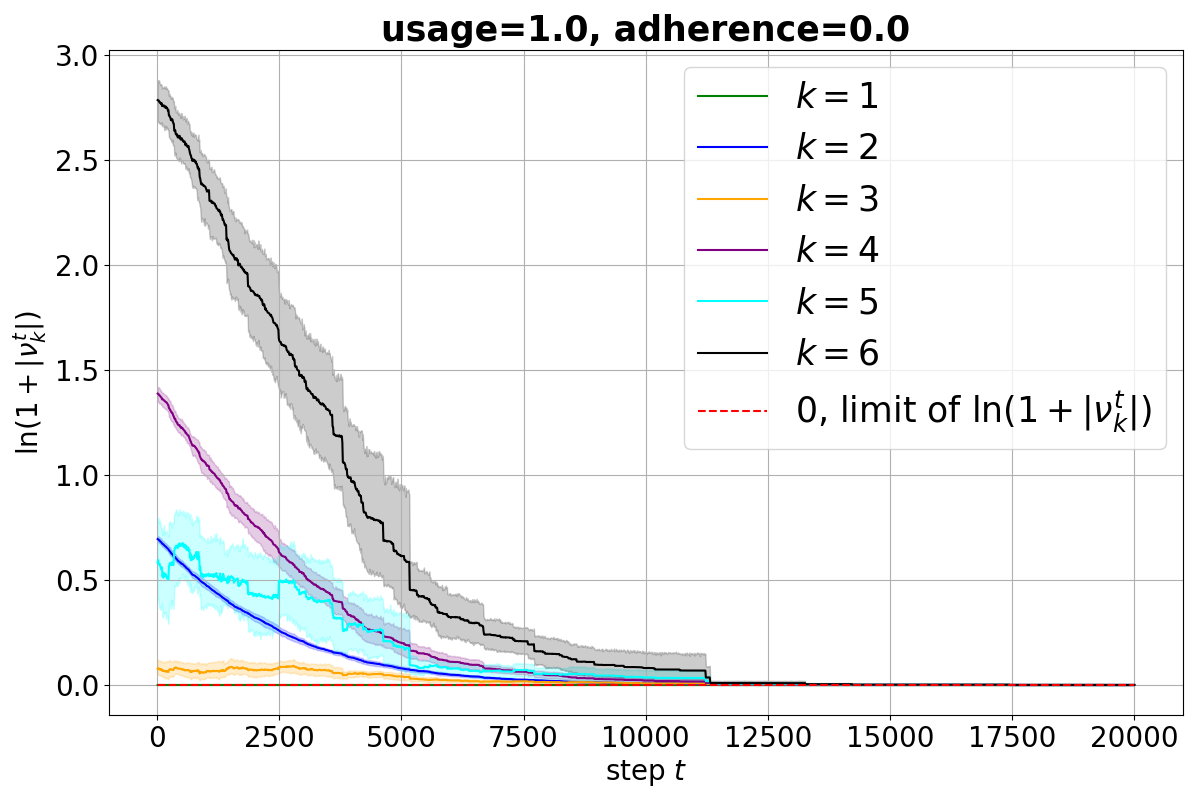}
            \includegraphics[width=0.49\linewidth]{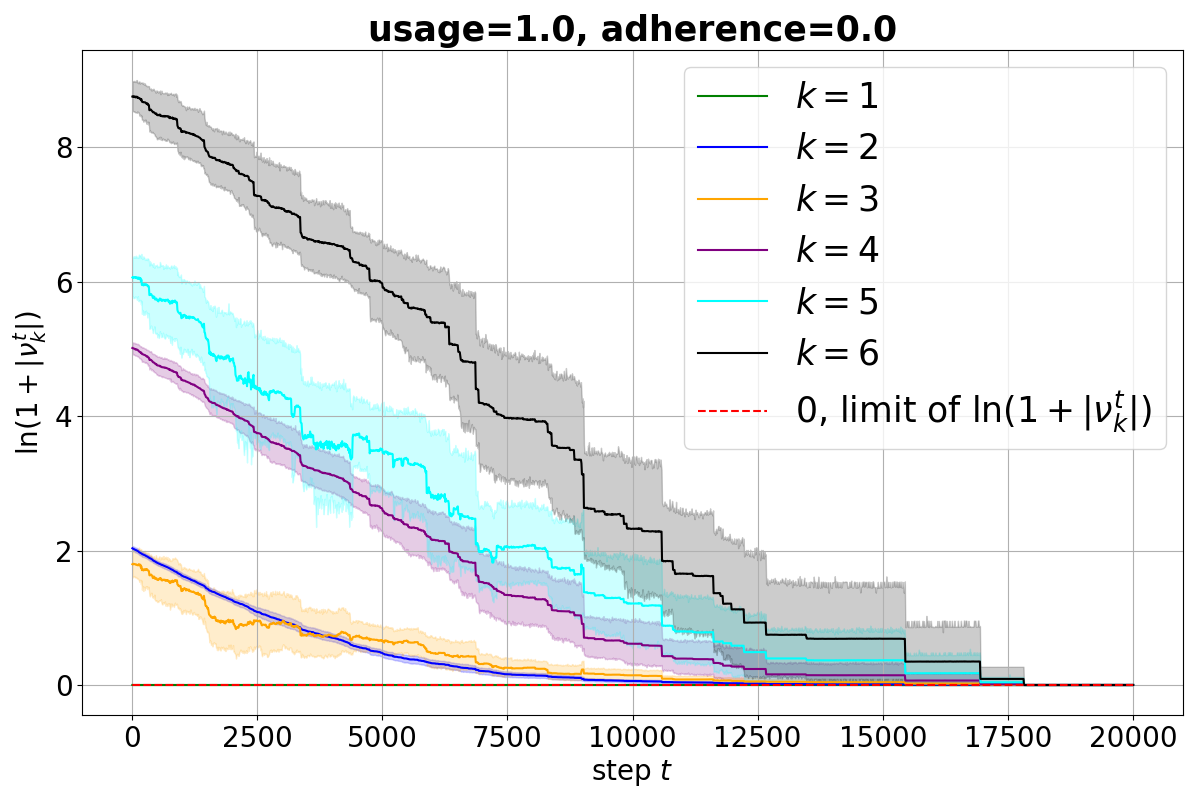}
            
            \caption{Measurement $\nu_k^t$ for $k = 1, 2, 3, 4$ and $5$ for sampling update setup. Synthetic linear data set (left) and Friedman data set (right). As we can see, the third statement of Lemma~\ref{moments} is satisfied.}
            \label{fig_exp_5_2}
        \end{figure}

        In the sampling update experiment setting we check the second claim of the Lemma, that is $\nu_k^t \underset{t \to +\infty}{\longrightarrow} 0$ for $k = 1, 2, ... , 6$. The observations for this experiments are shown at Fig.~\ref{fig_exp_5_1} and Fig.~\ref{fig_exp_5_2}.  All the results in this experiment are for SGD regression model on synthetic linear and Friedman data sets.

        As we can see from the measurements, the second and third claims of Lemma~\ref{moments} are satisfied in all observed cases. When usage $p = 1$ and adherence $s = 0$ the limit of mappings $\text{D}_{\overline{1, t}}(f_0)$, and correspondingly of $y_i - y_i'$, is the delta function $\delta(x)$. However, we do not get exactly zero, as the plot may show, since in all cases we consider only finite $t$.

%%%%%%%%% Limitations %%%%%%%%%
\section{Limitations and Threats to Validity}

    Threats to \emph{internal validity} relate to how the experimental design and other internal factors affect the study we are conducting. The mathematical formulation of the repeated machine learning \eqref{system} describes a broad class of problems. The theory we develop places such constraints on the mappings $\text{D}_t$ and $\widetilde{\text{D}}_t$ that are satisfied in most practical applications. However, in our experiments we only test the theoretical predictions for two synthetic data sets with additive noise, and three linear regression models and learning algorithms. Results of the experiments depend on the model and the data set we use. Another problem could be the imperfection of the Python language and the modeling environment we use.  In the Experiment~\ref{exp_3} we measure $f_t(0)$ and observe distortions in the plot at larger values of step $t$ (see Fig.~\ref{delta_loop}, Fig.~\ref{delta_sample}), which may be caused by rounding errors and approximation errors between the true probability density function and the empirical density function \eqref{F_approx}. 
    
    Threats to \emph{construct validity} relate to the correspondence between theoretical predictions and what is observed in the experiment. In Experiment~\ref{exp_3} we study the convergence of sequences of density functions $f_t$ to the zero distribution (see Fig.~\ref{delta_loop} and Fig.~\ref{delta_sample}). We take the integral $\int_{-\kappa}^{\kappa}f_t(x)\,dx$, but in general the functions $f_t$ can tend to a non-zero value somewhere outside the interval $[-\kappa, \kappa]$. In practice it is very difficult to choose a neighbourhood of infinity, therefore we choose several values for $\kappa$ and check that for all of them the integral tends to zero as the step $t$ increases.

    Threats to \emph{conclusion validity} relate to the correspondence between theory and result. When testing our system for autonomy in Experiment~\ref{exp_4}, we check whether the $r^2$ score for a trend line is small enough to have $\ln(f_t(0)$ follow a straight line with step $t$. We conclude that one of the designs is autonomous and the other is not from this test. However, there is no clear bound on $r^2$ score at which the $\ln(f_t(0))$ graph can be recognised as not a straight line. Moreover, in our theoretical predictions the step $t$ tends to infinity, while in practice we take only finite $t$. In the sliding window experiment (see Fig.~\ref{ex_set}) we take step $t \leq 1400$ because the process has to stop by construction after that. In the sampling update experiment we take $t \leq 30000$, as we consider this number of steps sufficient. In the $l_1$ norm test of the tendency of moments to zero in Experiment~\ref{exp_5} we cannot compute an infinite sum of moments either, therefore we limit ourselves to the first $N=300$ moments assuming that the remaining higher moments do not contribute a lot.

    Threats to \textit{external validity} relate to the differences between the simulation environment and real systems. We propose two experiment designs that are strong simplifications of  repeated machine learning systems. For example, in the real-world systems, a user may or may not adhere to the predictions of the system with a non-variable probability of use. In addition, user adherence may change over time, or the decision to adhere may depend on the model's predictions in more complex ways, such that the linear approximation we make in Section~\ref{Experiments} does not hold. We could give much more examples of how the described schemes of repeated machine learning process do not correspond to reality, but since the purpose of our experiments is to verify the theoretical results, these simulations are sufficient for the goals of the current research.

%%%%%%%%% Conclusion %%%%%%%%%
\section{Conclusion}

    The problem of repeated machine learning we study is very important because a machine learning algorithm is almost always part of a larger software system and interacts with the environment of the system. It is important to understand how this interaction happens, how it changes the quality characteristics of the machine learning system, and are there any violations of the trustworthiness requirements.

    In this paper we apply the apparatus of discrete dynamical systems, mathematical analysis and probability theory to build a theoretical framework to state and solve the problem of repeated supervised learning. We correspond a machine learning system operating in its context of use with a dynamical system defined by the evolutionary mapping $\text{D}_t$ \eqref{system}. Theorem~\ref{delta} gives us the understanding of how this mapping transforms the initial data distribution and what is the limiting distribution: a delta function or a zero distribution. From this result we can judge whether a positive feedback loop exists and produces a concept drift, or a negative loop is present and degrades the prediction performance of the system. 
    
    We also prove Theorem~\ref{semigroup} and Lemma~\ref{moments} using results from Theorem~\ref{delta}. Theorem~\ref{semigroup} is the criterion of the autonomy of our dynamical system, a useful property for the further analysis. Lemma~\ref{moments} is about the tendency to zero moments of the residuals in a positive feedback loop. These results, unlike the statements of the Theorem~\ref{delta}, are easier to verify experimentally, which makes them useful in practical applications.

    We develop a simulation environment in the Python language for the repeated supervised learning process, and demonstrate the effects of repeated machine learning in two settings: sliding window update and sampling update (see Fig.~\ref{ex_set}). Both settings are reusable for any other experiments involving repeated machine learning. As shown in Experiment~\ref{exp_4}, the sampling update setting is autonomous, while the sliding window update setting is not.

    All the results we obtain are strongly proven and tested in computational experiments. Our assumptions turned out to be correct, the practical results agree with the theory.

    Expanding on the ideas in this paper is important for developing machine learning systems and its applications. Future research may include more experiments on real-world data sets and more complex models. 
    In the experiments and in the discussion of Theorem~\ref{delta}, we do not consider deriving the envelope function $g$. Perhaps, future studies should be devoted to its construction. Another direction for future research could be to endow the set $\textbf{F}$ with a metric like Kullback-Leibler metric and analyse mappings $\text{D}_t$ in this metric space.

%%%%%%%%% References %%%%%%%%%
% \bibliographystyle{plainnat}
% \bibliography{refs} 

%%%%%%%%% Appendix %%%%%%%%%
\newpage
\appendix
\begin{center}
    \Large \textbf{Appendix}
\end{center}
\normalsize

\section{Proof of Theorem~\ref{R_to_R}} \label{pr_R_to_R}
    \begin{proof}
        To begin with, let us note that if $\text{D}_t: \textbf{F} \to \textbf{F}$, then $\|\text{D}_t\|_1 = 1$, because by definition of mapping norm:

        \begin{equation*}
            \|\text{D}_t\|_1 = \underset{\|f\|_1 = 1}{\sup}\left\{\|\text{D}_t(f)\|_1\right\}.
        \end{equation*}

        And if $f$ such that $\|f\|_1 = 1$ then $|f| \in \textbf{F}$ and, because $\text{D}_t: \textbf{F} \to \textbf{F}$, $\|\text{D}_t(|f|)\|_1 = 1$. But $\|\text{D}_t\|_1 = 1$ is only a necessary but not a sufficient condition.

        If $\|\text{D}_t\|_1 = 1$, then for all $f \in \textbf{F}$ holds that $\text{D}_t(f) \leq 1$. If exists $f_0 \in \textbf{F}$ such that $\|\text{D}_t(f_0)\|_1 < 1$, then we get a contradiction because

        \begin{multline*}
            \|D^{-1}\|_1 \overset{def}{=} \underset{\|f\|_1 \neq 0}{\sup}\left\{\dfrac{\|\text{D}_t^{-1}(f)\|_1}{\|f\|_1}\right\} 
            \geq \left[f_1 = \text{D}_t(f_0)\right] \geq
            \dfrac{\|\text{D}_t^{-1}(f_1)\|_1}{\|f_1\|_1} = \\ = 
            \dfrac{\|\text{D}_t^{-1}(\text{D}_t(f_0))\|_1}{\|\text{D}_t(f_0)\|_1} = 
            \dfrac{\|f_0\|_1}{\|\text{D}_t(f_0)\|_1} = \dfrac{1}{\|\text{D}_t(f_0)\|_1} > 1.
        \end{multline*}

        We assume that $\|\text{D}_t^{-1}\|_1 \leq 1$. Then for all $f \in \textbf{F}$ holds that $\|\text{D}_t(f)\|_1 = 1$.

        But according to Theorem~\ref{based} to $\text{D}_t : \textbf{F} \to \textbf{F}$ we also need second assumption: $\forall f \in \textbf{F} \hookrightarrow \text{D}_t(f)(x) \geq 0$ for almost every $x \in \mathbb{R}^n$.
        
    \end{proof}

\section{Proof of Theorem~\ref{delta}} \label{pr_delta}
    \begin{proof}
        For simplicity, let us assume the following notation $I_t(A) := \int_{A} f_t(x) (\phi(x) - \phi(0) ) dx$, where $A \subseteq \mathbb{R}^n$. Let us now prove the first claim. 
        
        \begin{equation*}
            I_t\left(\mathbb{R}^n\right) = \int\limits_{\mathbb{R}^n} f_t(x) \phi(x) dx - \phi(0) \cdot \int\limits_{\mathbb{R}^n} f_t(x) dx = \int\limits_{\mathbb{R}^n} f_t(x) \cdot [\phi(x) - \phi(0)] dx.
        \end{equation*}
        The first equality holds since $\|f_t\|_1 = 1, f_t(x) \geq 0$.    
        Replacing the variable $y = \psi_t \cdot x, dy = \psi_t^n \cdot dx$ we get
        \begin{equation} \label{tmp2_A}
            I_t\left(\mathbb{R}^n\right) = \dfrac{1}{\psi_t^n} \cdot \int\limits_{\mathbb{R}^n} f_t\left(\frac{y}{\psi_t}\right) \cdot \left[\phi\left(\frac{y}{\psi_t}\right) - \phi(0)\right] dy.
        \end{equation}
        Next, split the integral \eqref{tmp2_A} into two parts: $         I_t\left(\mathbb{R}^n\right) = I_t\left(B^n(R) \right) + I_t\left(\mathbb{R}^n \setminus B^n(R)\right)
        $, where $B^n(R)$ is a Euclidean ball with radius $R$. We consider each term separately, start with $I_t\left(\mathbb{R}^n \setminus B^n(R) \right)$. Function $\phi$ is continuous with a compact support, therefore $\phi$ is bounded by some constant $M$, that is for all $x \in \mathbb{R}^n$ exist big enought constant $M = M(\phi) > 0$ such that $|\phi(x)| \leq M$, hence 
        \begin{equation*}
            \left|I_t\left(\mathbb{R}^n \setminus B^n(R) \right)\right| \leq \int\limits_{\mathbb{R}^n \setminus B^n(R)} 2 M \cdot \dfrac{1}{\psi_t^n} f_t\left(\frac{y}{\psi_t}\right) dy 
            \leq \int\limits_{\mathbb{R}^n \setminus B^n(R)} 2 M \cdot |g(y)| dy .
        \end{equation*}
        The last equality holds since $f_t(x) \leq \psi_t^n |g(\psi_t x)|$. Since $g \in L_1(\mathbb{R}^n)$, for all $\varepsilon > 0$ exists big enough constant $R = R(\varepsilon) > 0$ such that $\left|I_t\left(\mathbb{R}^n \setminus B^n(R) \right)\right| \leq \varepsilon$.

        Consider the integral $I_t\left(B^n(R) \right)$. Function $\phi$ is continuous, therefore for all $\varepsilon > 0$ exist small enough constant $\delta = \delta(\varepsilon) > 0$ such that if $\|x - 0\| \leq \delta$, then $|\phi(x) - \phi(0)| \leq \varepsilon$. Sequence $\psi_t \to +\infty$, therefore for all $\varepsilon > 0$ and for all $y \in B^n(R)$ exist big enough constant $T = T(\delta, R) = T(\varepsilon) \in \mathbb{N}$ such that for all $t \geq T$:
        \begin{equation*}
            \left\|\frac{y}{\psi_t} - 0\right\| \leq \delta ~\text{ and respectively }~ \left|\phi\left(\frac{y}{\psi_t}\right) - \phi(0)\right| \leq \varepsilon.
        \end{equation*}
        Therefore, the following inequality holds 
        \begin{equation*}
            \left|I_t\left(B^n(R) \right)\right| \leq \varepsilon \cdot \int\limits_{B^n(R)} \dfrac{1}{\psi_t^n} f_t\left(\frac{y}{\psi_t}\right) dy = 
            \varepsilon \cdot \int\limits_{B^n(R / \psi_t)} f_t\left(x\right) dx \leq \varepsilon \cdot \int\limits_{\mathbb{R}^n} f_t\left(x\right) dx = \varepsilon.
        \end{equation*}

        Finally we get

        \begin{equation*}
            \forall \varepsilon > 0 ~~ \exists~ R = R(\varepsilon) >0, ~  \exists~ \delta = \delta(\varepsilon) > 0, ~ \exists~ T = T(\delta, R) = T(\varepsilon) \in \mathbb{N} : \forall t \geq T \hookrightarrow \left|I_t\left( \mathbb{R}^n \right)\right| \leq 
             2 \varepsilon.
        \end{equation*}
        Thus $f_t(x) \underset{t \to \infty}{\longrightarrow} \delta(x)$ in a weak sense.

        The proof of the second term is quite similar to the first one. Again for simplicity let us assume the following notation $J_t(A) := \int_{A} f_t(x) \phi(x) dx$. Replacing the variable $y = \psi_t \cdot x, dy = \psi_t^n \cdot dx$ we get
        \begin{equation*}
            J_t\left( \mathbb{R}^n\right) = \dfrac{1}{\psi_t^n} \cdot \int\limits_{\mathbb{R}^n} f_t\left(\frac{y}{\psi_t}\right) \phi\left(\frac{y}{\psi_t}\right) dy = J_t\left(B^n(R) \right) + J_t\left(\mathbb{R}^n \setminus B^n(R) \right).
        \end{equation*}

        Consider integral $J_t\left(B^n(R) \right)$. Similarly to the results above, we can obtain 
        $
            \left| J_t\left(B^n(R) \right) \right| \leq \int_{B^n(R)} M \cdot |g(y)| dy. 
        $
        Since $g \in L_1\left( \mathbb{R}^n \right)$, for all $\varepsilon > 0$ exists small enough constant $R = R(\varepsilon) > 0$ such that $\left|J_t\left(B^n(R) \right)\right| \leq \varepsilon$. This fact follows from absolute continuity of the Lebesgue integral. 
        
        Consider integral $J_t\left(\mathbb{R}^n \setminus B^n(R) \right)$. Function $\phi$ has a compact support and sequence $\psi_t \to 0$, then for all $\varepsilon > 0$ and $y \in \mathbb{R}^n \setminus B^n(R)$ exists big enough constant $T = T(\varepsilon, R) = T(\varepsilon) \in \mathbb{N}$ such that
        for all $t \geq T$ holds that $\phi\left(\frac{y}{\psi_t}\right) \leq \varepsilon$, hence $|J_t\left(\mathbb{R}^n \setminus B^n(R) \right)| \leq \varepsilon$.

        Therefore, $f_t(x) \underset{t \to \infty}{\longrightarrow} \zeta(x)$ in a weak sense.

    \end{proof}

\section{Proof of Lemma~\ref{moments}} \label{pr_moments}
    \begin{proof}
        Let's first prove first term. By the definition of $k$-moment we have
        \begin{equation*}
            \nu_{2k}^t = \int\limits_{-\infty}^{+\infty} x^{2k} f_t(x) dx \leq \int\limits_{-\infty}^{+\infty} x^{2k} \psi_t g(\psi_t x) dx.
        \end{equation*}
        This inequality is true since $x^{2k} \geq 0$. If we make variable substitution $y = \psi_t \cdot x$, when we have
        \begin{equation*}
            \nu_{2k}^t \leq \int\limits_{-\infty}^{+\infty} \dfrac{y^{2k}}{\psi_t^{2k}} g(y) dy = \psi_t^{-2k} \nu_{2k}^0.
        \end{equation*}
        Thus, the first term is proved. Consider the second term. If the system \eqref{system} evolution operator satisfies \eqref{cool_D}, then
        \begin{equation*}
            \nu_{k}^t = \int\limits_{-\infty}^{+\infty} \dfrac{y^k}{\psi_t^k} g(y) dy = \psi_t^{-k} \nu_{k}^0.
        \end{equation*}
        Therefore, the second term has been proven. Consider the third term.
        \begin{equation*}
            \|\{\nu_k^t\}_{k=1}^{+\infty}\|_1 = \|\{\psi_t^{-k} \nu_k^0\}_{k=1}^{+\infty}\|_1 \leq \|\{\psi_t^{-k}\}_{k=1}^{+\infty}\|_p \cdot \|\{\nu_k^0\}_{k=1}^{+\infty}\|_q.
        \end{equation*}
        The second step follows from Helder's inequality.
        Now let's calculate $\|\{\psi_t^{-k}\}_{k=1}^{+\infty}\|_p$ for $p \in [1; +\infty)$:
        \begin{equation*}
            \|\{\psi_t^{-k}\}_{k=1}^{+\infty}\|_p^p = \sum\limits_{k=1}^{+\infty}\psi_t^{-kp} = \dfrac{\psi_t^{-p}}{1 - \psi_t^{-p}} = \dfrac{1}{\psi_t^p - 1}.
        \end{equation*}
        The first equality holds only if $\psi_t > 1$ and the second step follows from the sum of infinitely decreasing geometric progression. Next we obtain
        \begin{equation*}
            \|\{\psi_t^{-k}\}_{k=1}^{+\infty}\|_p = \left( \dfrac{1}{\psi_t^p - 1} \right)^{1/p} \underset{t \to +\infty}{\longrightarrow} 0 ~~~\forall p \in [1; +\infty).
        \end{equation*}
        If $p = +\infty$:
        \begin{equation*}
            \|\{\psi_t^{-k}\}_{k=1}^{+\infty}\|_{\infty} = \left[ \text{ if } \psi_t > 1 \right] = \psi_t^{-1} \underset{t \to +\infty}{\longrightarrow} 0.
        \end{equation*}
        Therefore, we have 
        \begin{equation*}
            \|\{\nu_k^t\}_{k=1}^{+\infty}\|_1 \leq \|\{\psi_t^{-k}\}_{k=1}^{+\infty}\|_p \cdot \|\{\nu_k^0\}_{k=1}^{+\infty}\|_q \underset{t \to +\infty}{\longrightarrow} 0.
        \end{equation*}
        Because $\|\{\nu_k^0\}_{k=1}^{+\infty}\|_q < +\infty$ as a condition of Lemma.
        
    \end{proof}

\section{Proof of Lemma~\ref{ineq_q}} \label{pr_ineq_q}
    \begin{proof}
        First of all let's calculate $\|f_A\|_p$:
        \begin{equation*}
            \|f_A\|_p = \left( \text{ } \int\limits_{\mathbb{R}^n}\left(\dfrac{1}{\lambda(A)}\right)^p \cdot \textbf{1}_{A}(x) dx \text{ } \right)^{1/p} = \dfrac{1}{\lambda(A)} \cdot (\lambda(A))^{1/p} = (\lambda(A))^{-1 + 1/p}.
        \end{equation*}
        That is, $f_A \in L_p(\mathbb{R}^n)$ for all $A \subset \mathbb{R}^n : 0 < \lambda(A) < +\infty$ and $1 \leq p \leq +\infty$. 
        Now write out a Helder's inequality. For $q$ such that $\frac{1}{p} + \frac{1}{q} = 1$ the following inequality holds
        \begin{equation*}
            \|f_A\|_p \cdot \|\text{D}_t(f_A)\|_q \geq \|f_A \cdot \text{D}_t(f_A)\|_1.
        \end{equation*}
        Using common inequality on operators norm $\|\text{D}_t(f)\|_q \leq \|\text{D}_t\|_q \cdot \|f\|_q ~\forall f \in L_q(\mathbb{R}^n)$ we get
        \begin{equation*}
            \|f_A\|_p \|f_A\|_q \cdot \|\text{D}_t\|_q \geq \|f_A \cdot \text{D}_t(f_A)\|_1.
        \end{equation*}
        Since $\|f_A\|_p \|f_A\|_q = (\lambda(A))^{-1 + 1/p} \cdot (\lambda(A))^{-1 + 1/q} = (\lambda(A))^{-2 + 1/p + 1/q} = (\lambda(A))^{-1}$ we get:
        \begin{equation*}
            \|\text{D}_t\|_q \geq \lambda(A) \cdot \|f_A \cdot \text{D}_t(f_A)\|_1.
        \end{equation*}
        Consider $\|f_A \cdot \text{D}_t(f_A)\|_1$:
        \begin{equation*}
            \|f_A \cdot \text{D}_t(f_A)\|_1 = \int\limits_{\mathbb{R}^n} \text{D}_t(f_A)(x) \cdot \dfrac{1}{\lambda(A)} \textbf{1}_{A}(x) dx = \dfrac{1}{\lambda(A)} \int\limits_{A} \text{D}_t(f_A)(x)dx.
        \end{equation*}
        Finally, we get the desired inequality
        \begin{equation*}
            \|\text{D}_t\|_q \geq \int\limits_{A} \text{D}_t(f_A)(x)dx.
        \end{equation*}
        
    \end{proof}
\section{Proof of Theorem~\ref{semigroup}} \label{pr_semigroup}
    \begin{proof}
        \begin{equation*}
            (\text{D}_{\overline{1, \tau}} \circ \text{D}_{\overline{1, \kappa}})(f)(x) = \text{D}_{\overline{1, \tau}}(\psi_{\kappa}^n \cdot f(\psi_{\kappa} \cdot x)) = \psi_{\tau}^n\psi_{\kappa}^n \cdot f(\psi_{\tau}\psi_{\kappa} \cdot x),
        \end{equation*}
        \begin{equation*}
            \text{D}_{\overline{1, \tau + \kappa}}(f)(x) = \psi_{\tau + \kappa}^n \cdot f(\psi_{\tau + \kappa} \cdot x).
        \end{equation*}
        That is, system \eqref{system} is autonomous if and only if $\psi_{\tau + \kappa} = \psi_{\tau} \cdot \psi_{\kappa}$ for all $ \tau, \kappa \in \mathbb{N}$.
    \end{proof}

\section{Supplementary Diagrams for the Experiments}
\label{Supplementary_materials}

    In this Section we report the results of Experiment~\ref{exp_2}, which consisted of analysing the training sample for normality and Experiment~\ref{exp_3} were we test the predictions of Theorem~\ref{delta}. 

    \begin{figure}[ht]
        \centering
        \includegraphics[width=0.49\linewidth]{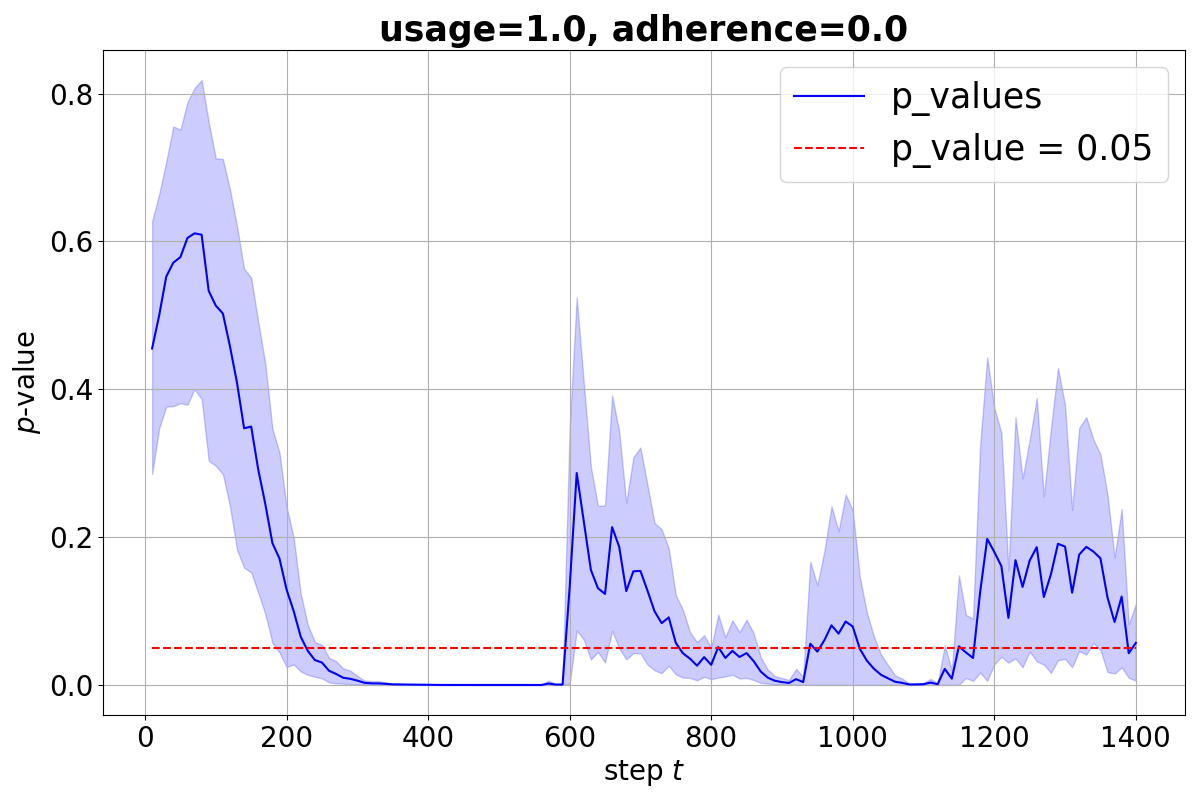}
        \includegraphics[width=0.49\linewidth]{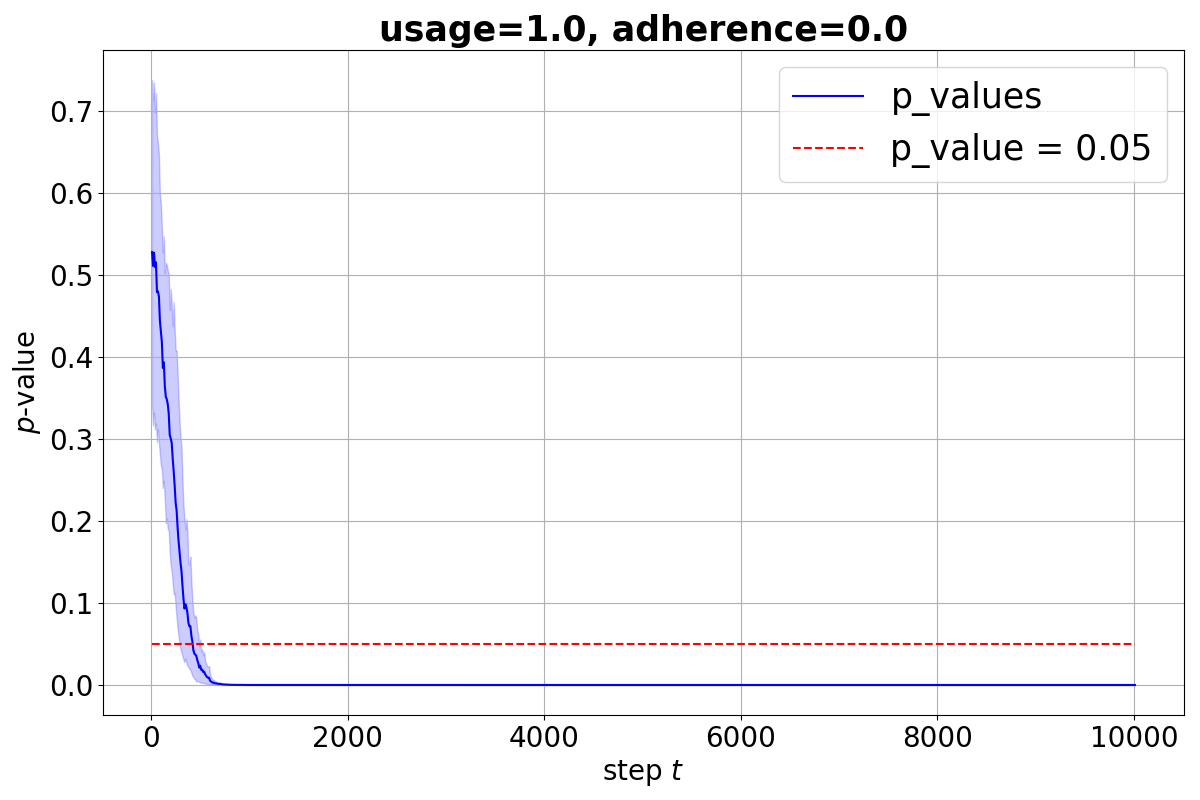}
        
        \caption{Testing the distribution of the model error for normality over time for SGD regression model on synthetic linear data set. Sliding window setup (left), sampling update setup (right).}
        \label{p_value}
    \end{figure}

    As we can see from Fig.~\ref{p_value}, the $p$-value becomes less than $0.05$ with time, that is the normality of the data breaks down.

    \begin{figure}[ht]
        \centering
        \includegraphics[width=0.49\linewidth]{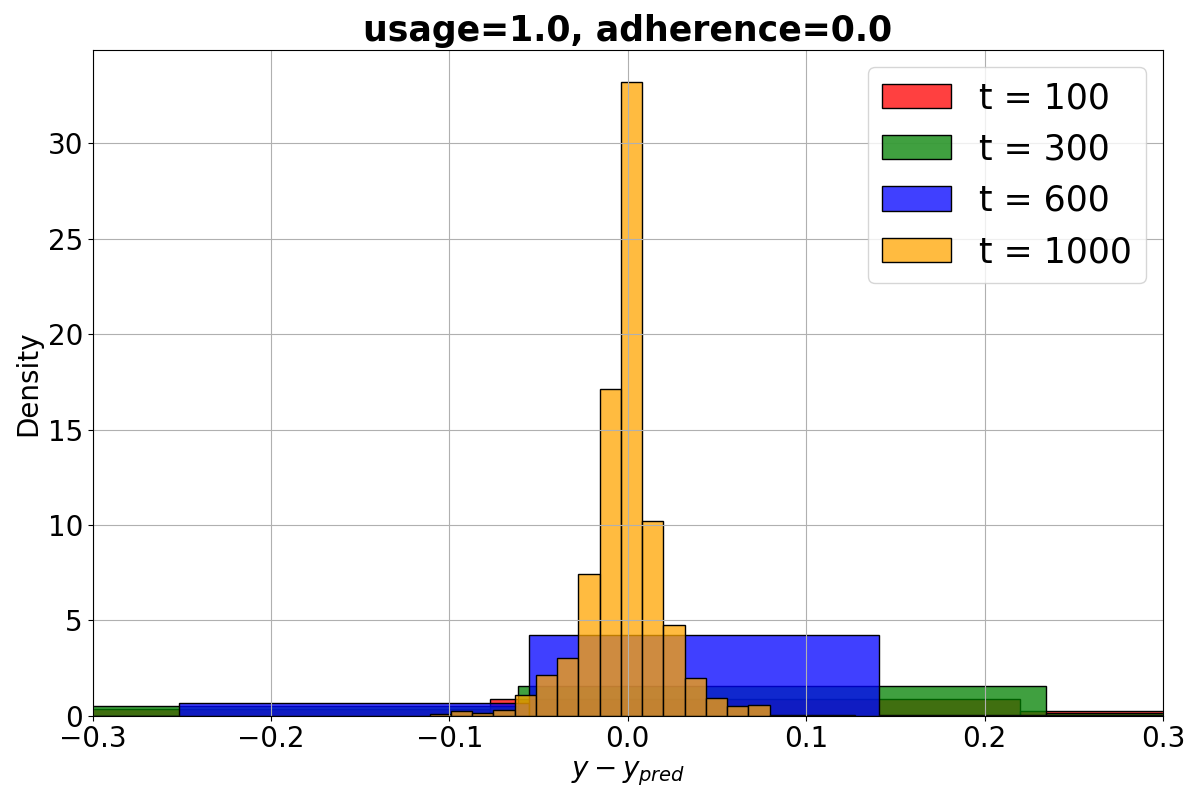}
        \includegraphics[width=0.49\linewidth]{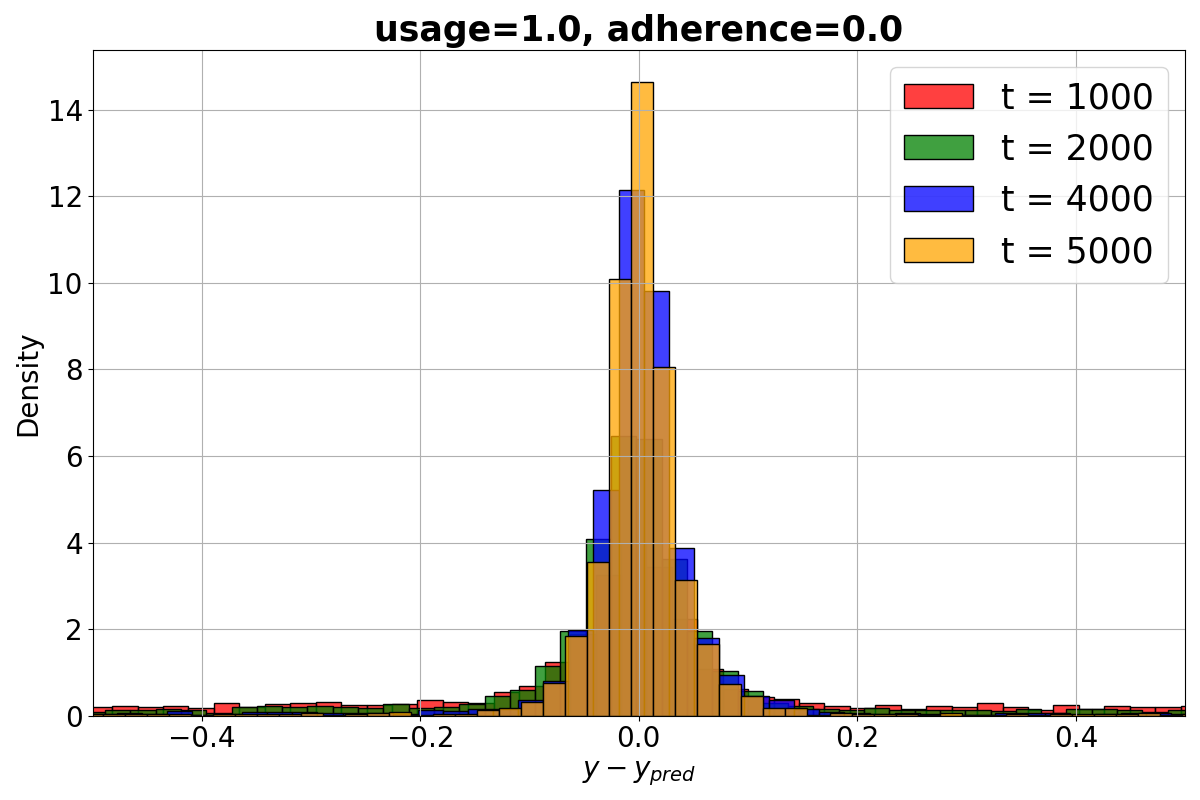}
        
        \caption{Histograms of the model error over time for SGD regression model on synthetic linear data set. Sliding window setup (left), sampling update setup (right).}
        \label{hist}
    \end{figure}

    From Fig.~\ref{hist} we may conclude that $\mathbf{y} - \mathbf{y'}$ is a mixture of the two probability distributions in the current set.

    \begin{figure}
        \centering
        \includegraphics[width=0.32\linewidth]{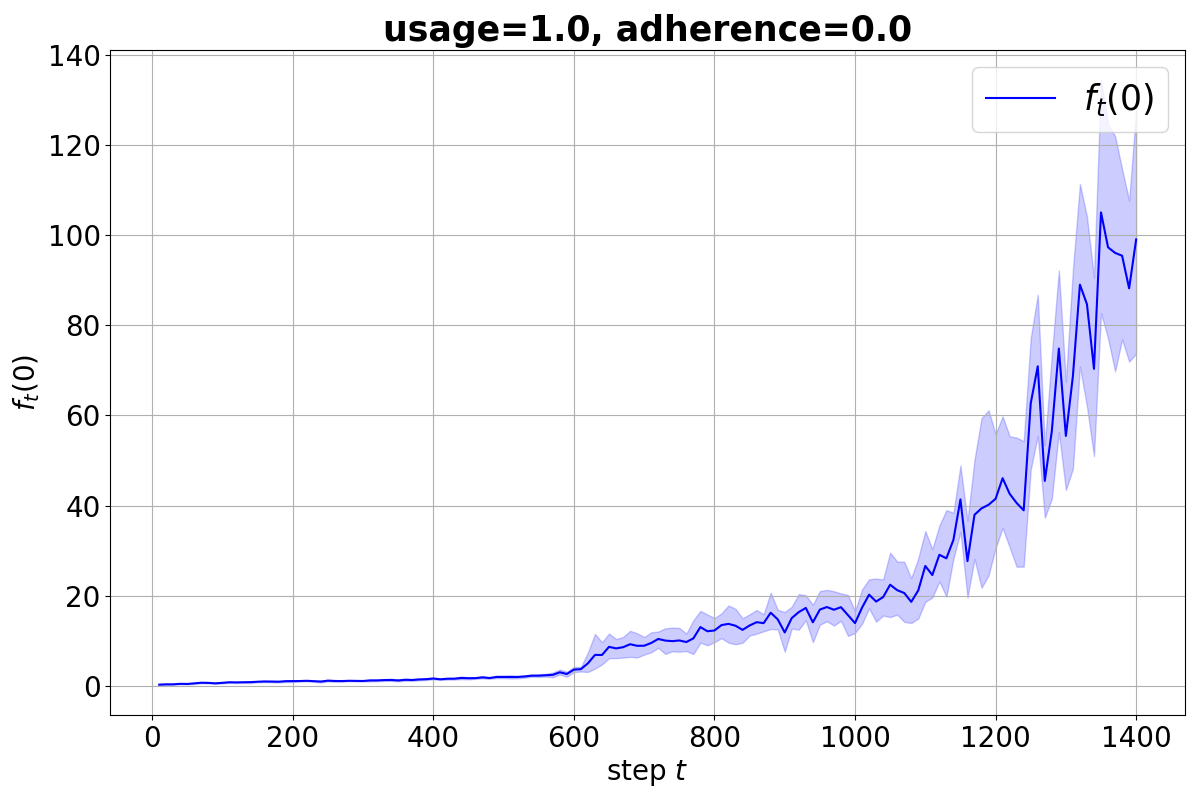}~
        \includegraphics[width=0.32\linewidth]{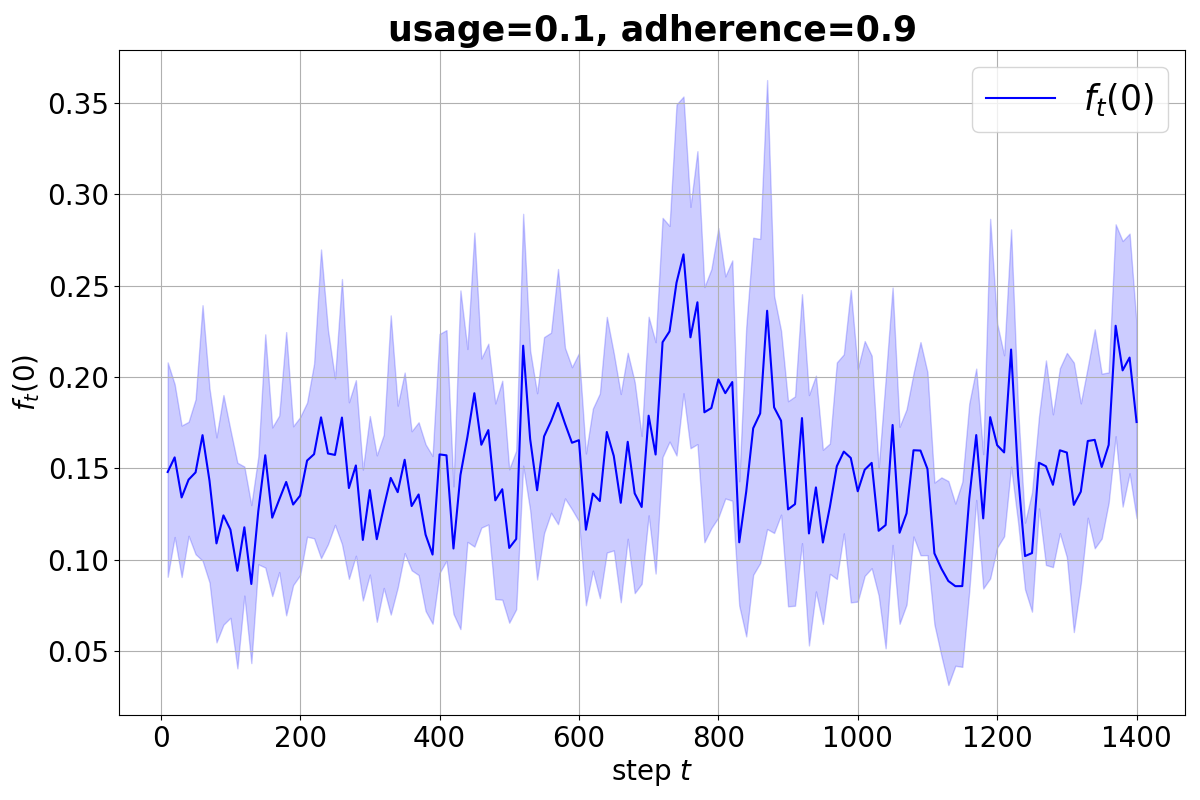}~
        \includegraphics[width=0.32\linewidth]{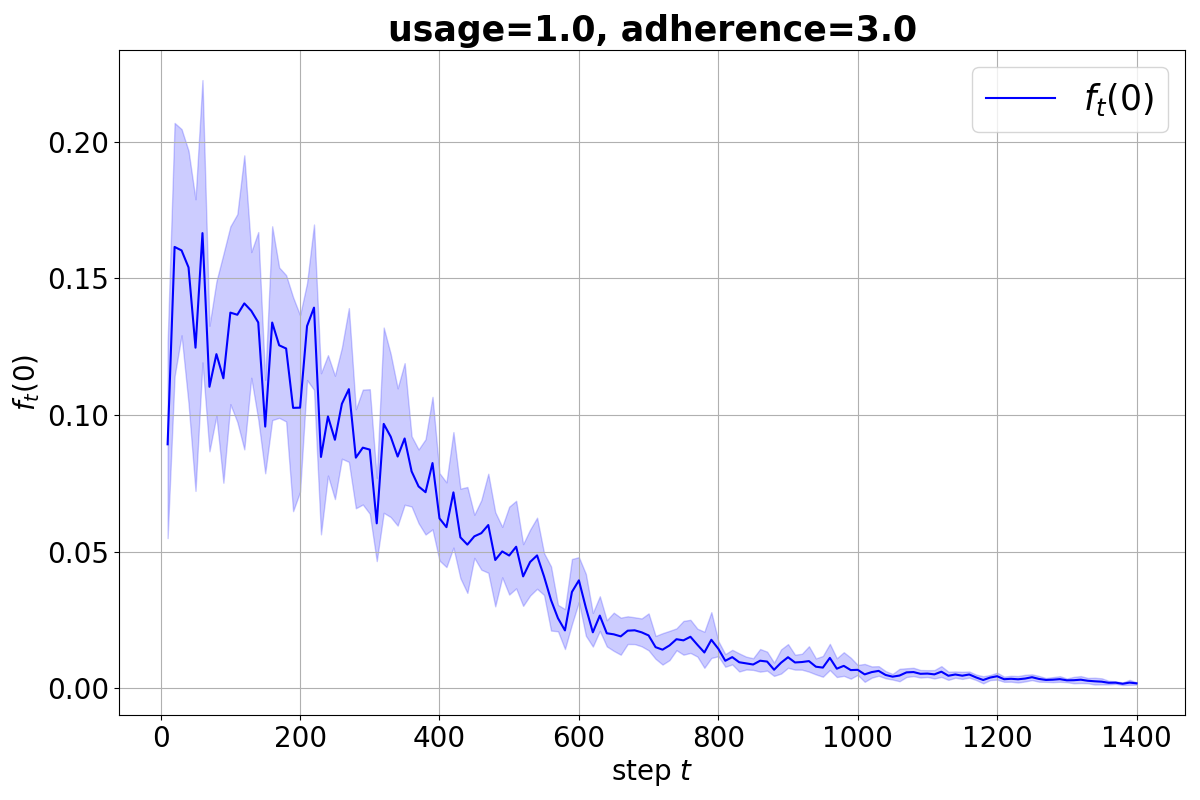}
    
        \includegraphics[width=0.32\linewidth]{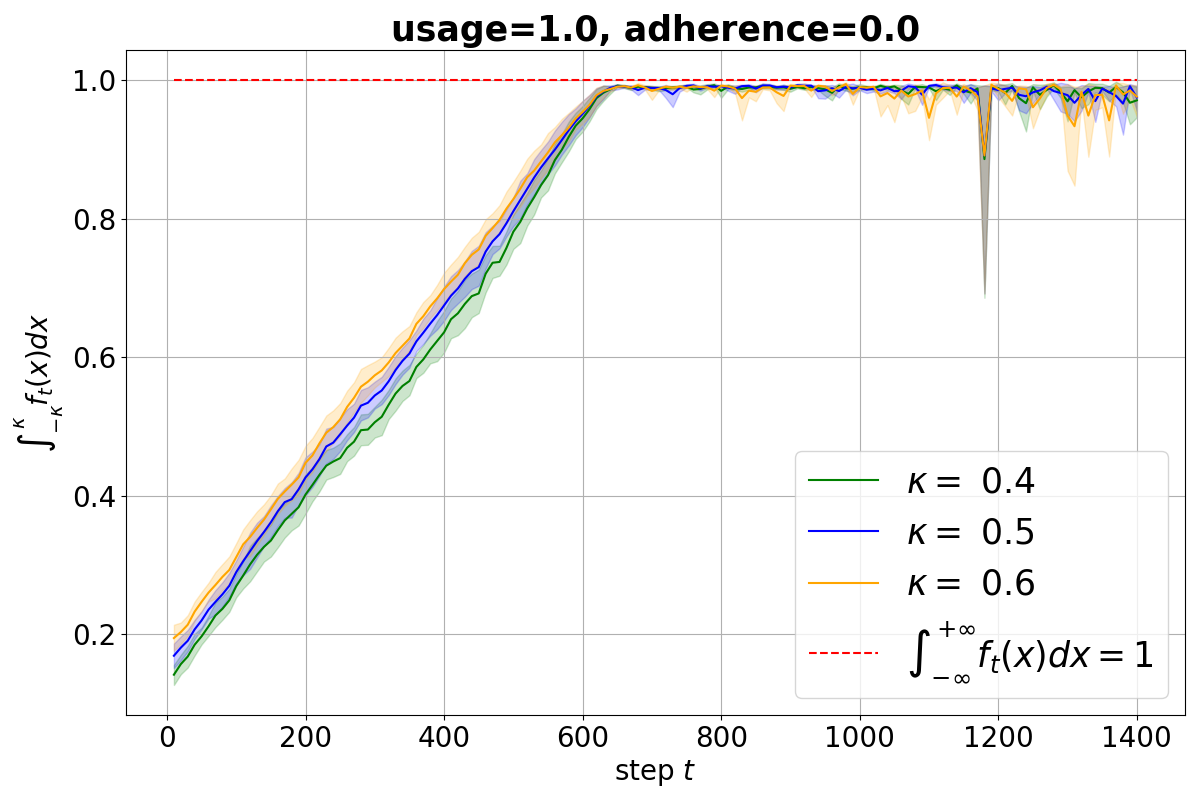}~
        \includegraphics[width=0.32\linewidth]{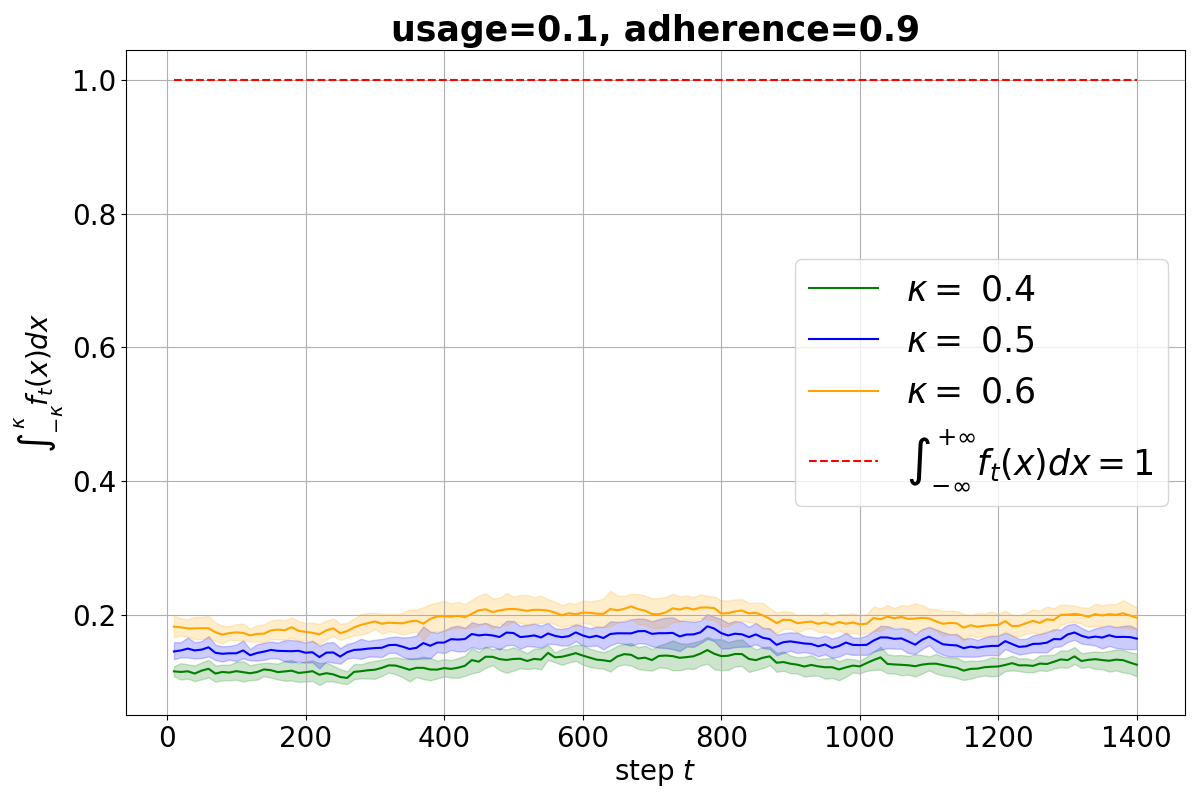}~
        \includegraphics[width=0.32\linewidth]{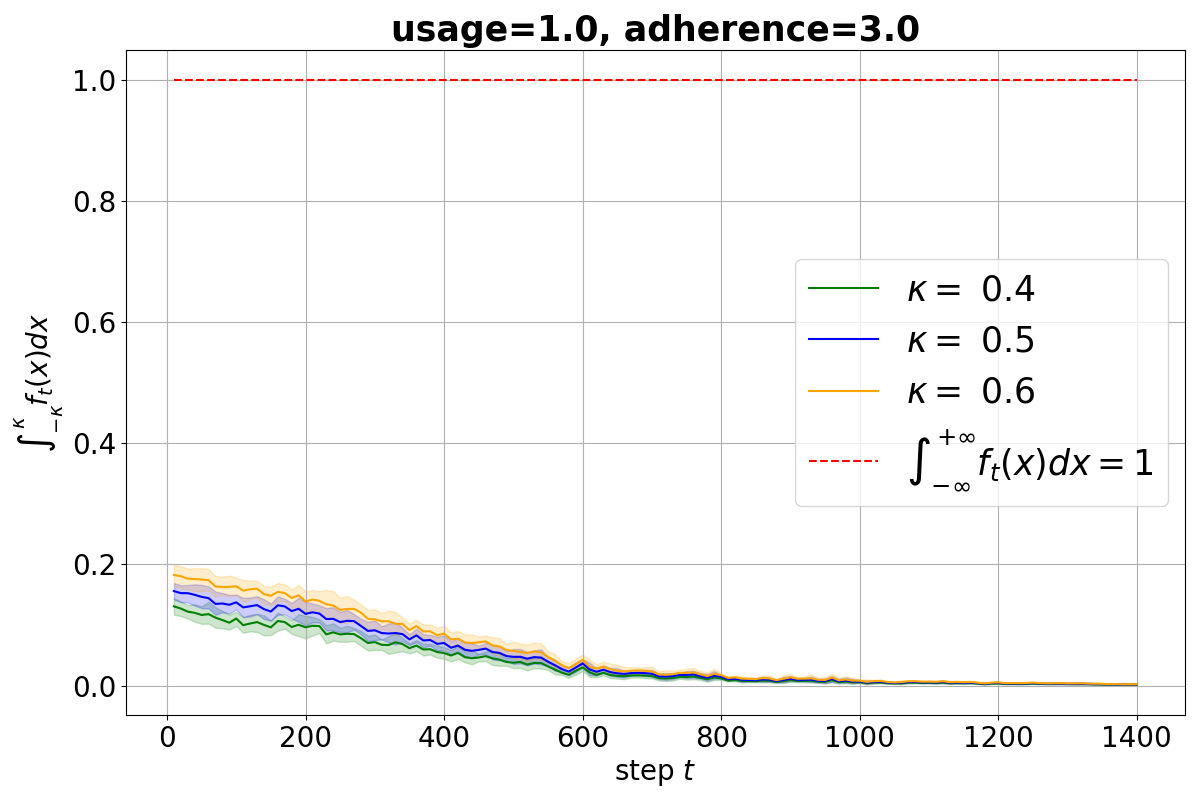}
        
        \caption{Counting $f_t(0)$ and $\int_{-\kappa}^{\kappa}f_t(x)dx$ for sliding window setup for Ridge regression model on Friedman data set. We consider such parameters: usage, adherence = $1$, $0$ (left); $0.1$, $0.9$ (middle); $1$, $3$ (right). In this picture, we can see the entire limit set of the system \eqref{system} from Theorem~\ref{delta}.}
        \label{delta_loop_1}
    \end{figure}

    \begin{figure}
        \centering
        \includegraphics[width=0.32\linewidth]{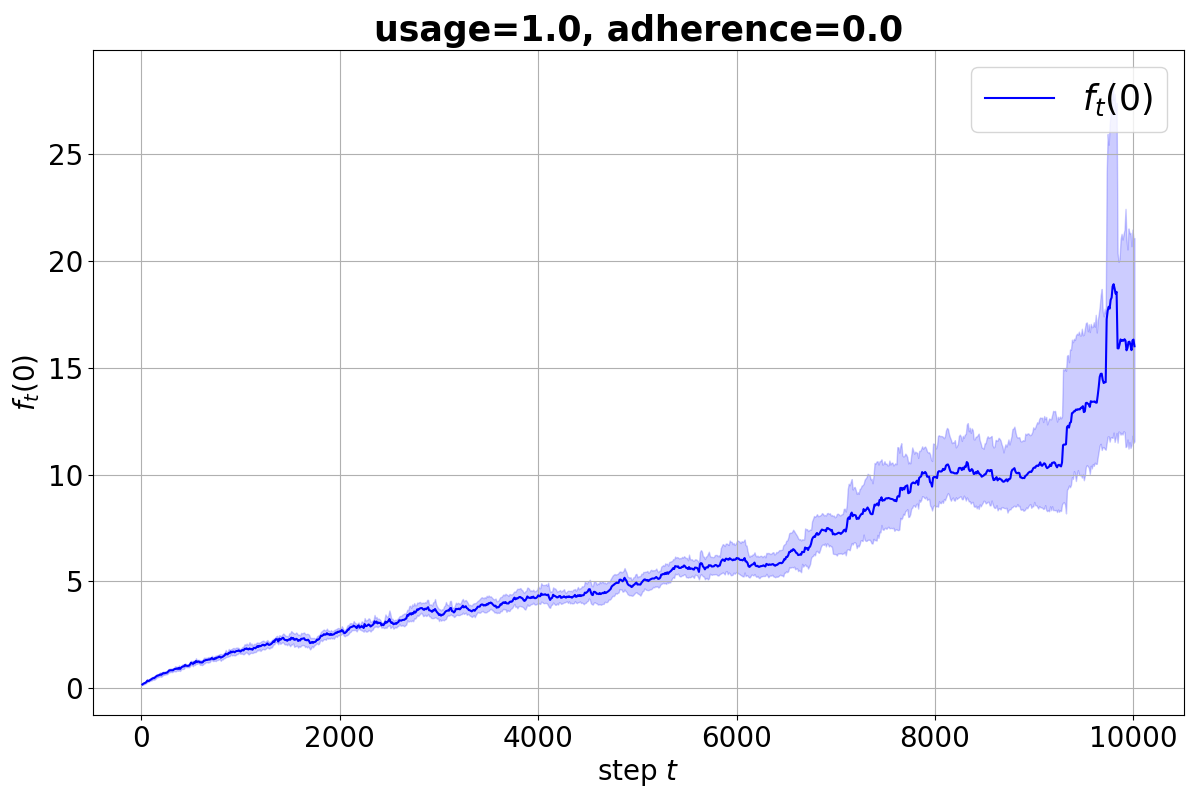}~
        \includegraphics[width=0.32\linewidth]{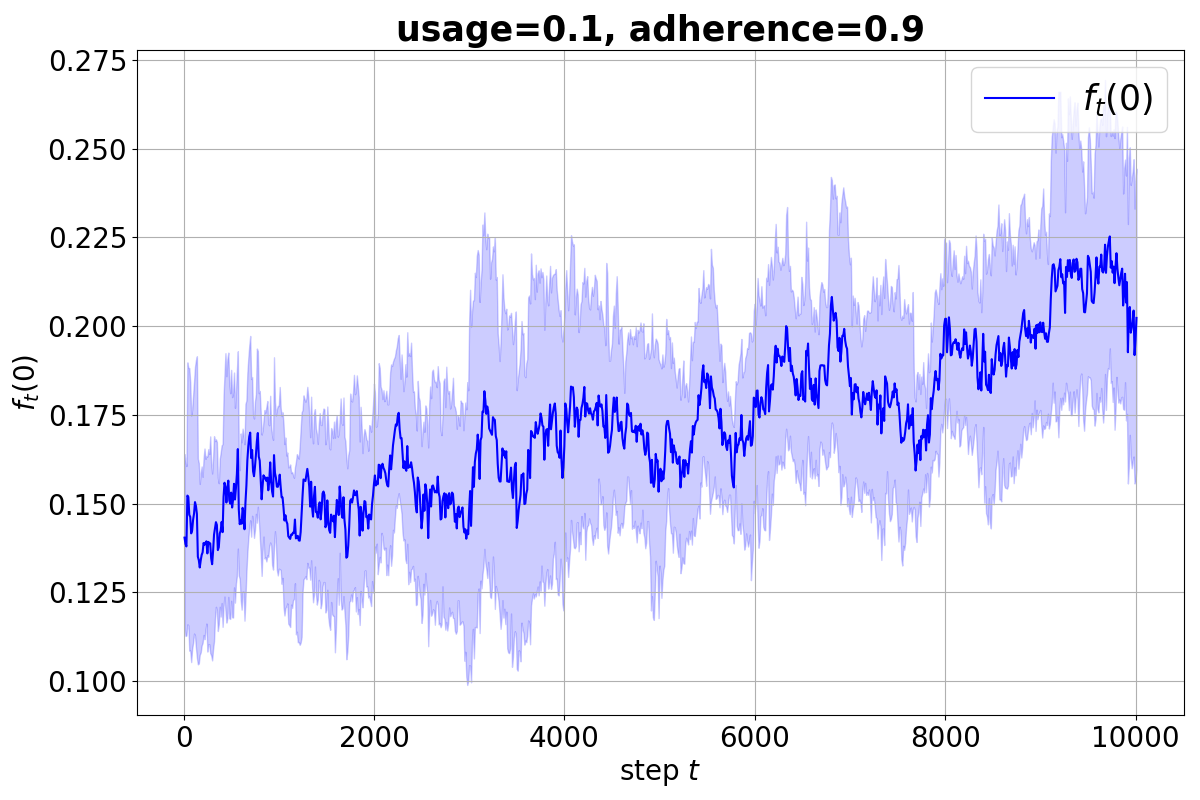}~
        \includegraphics[width=0.32\linewidth]{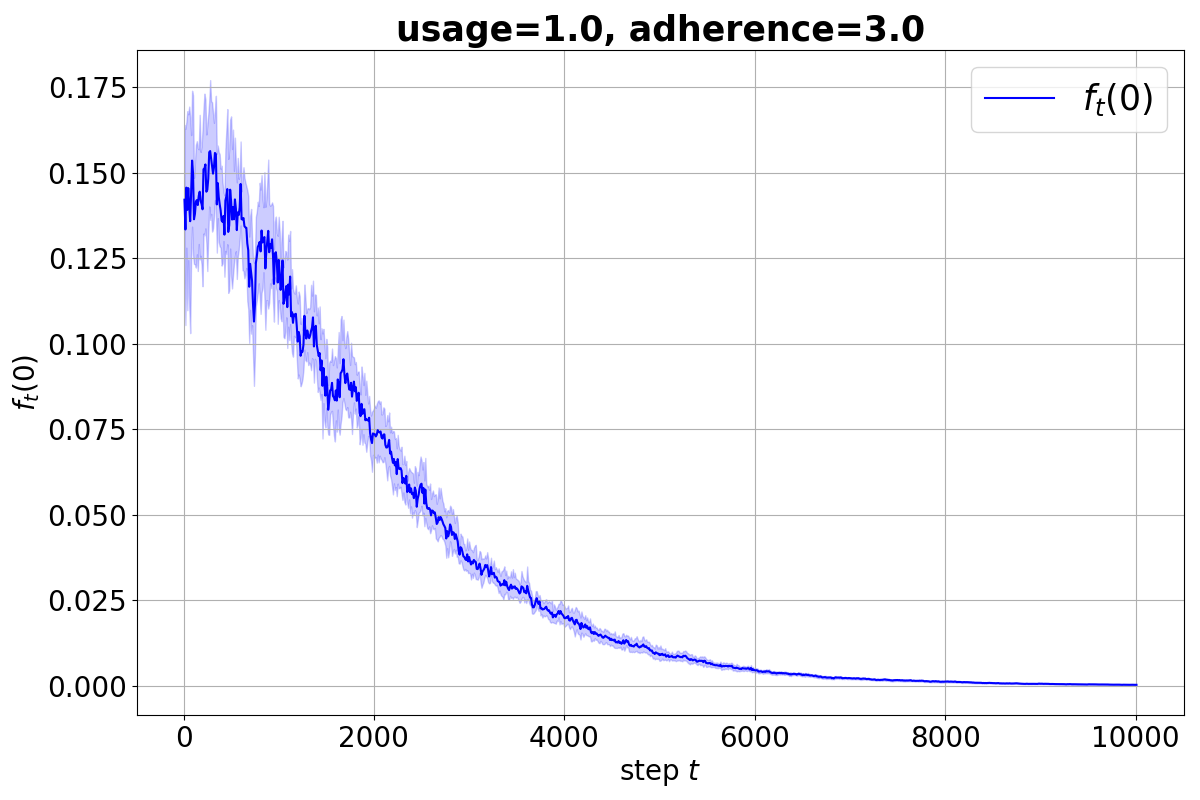}

        \includegraphics[width=0.32\linewidth]{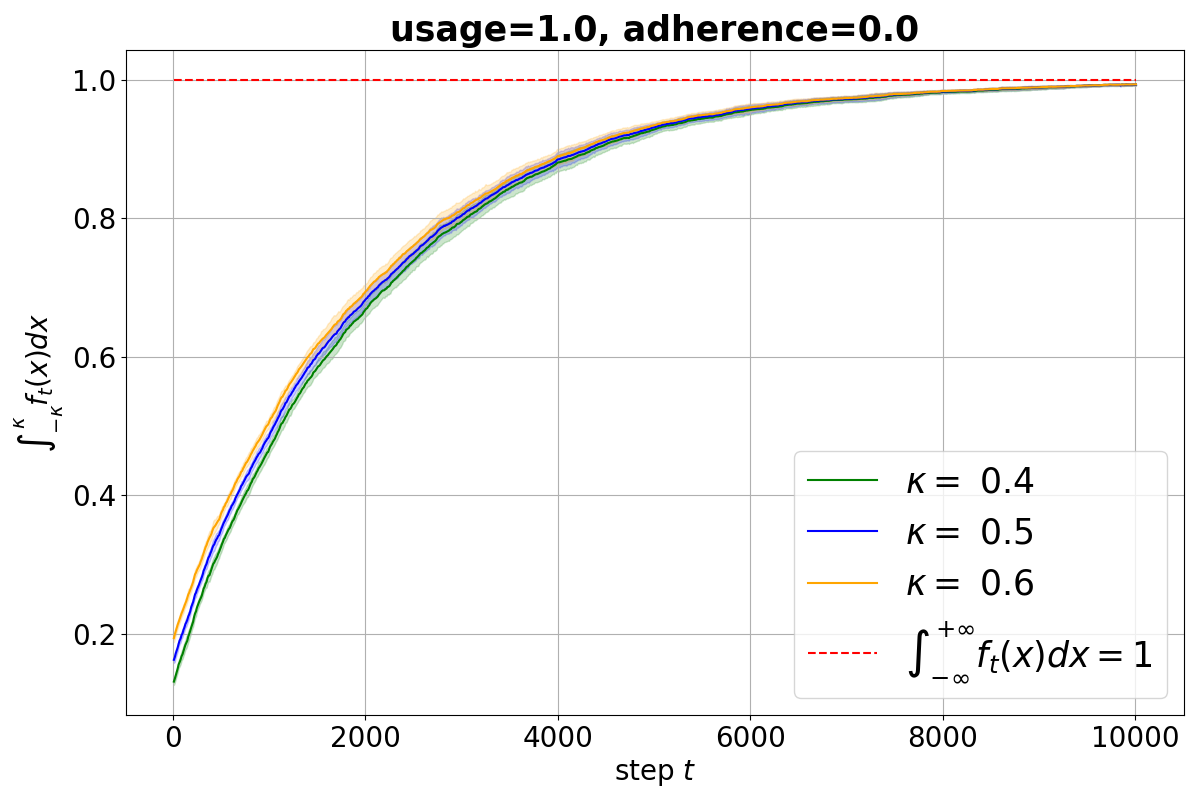}~
        \includegraphics[width=0.32\linewidth]{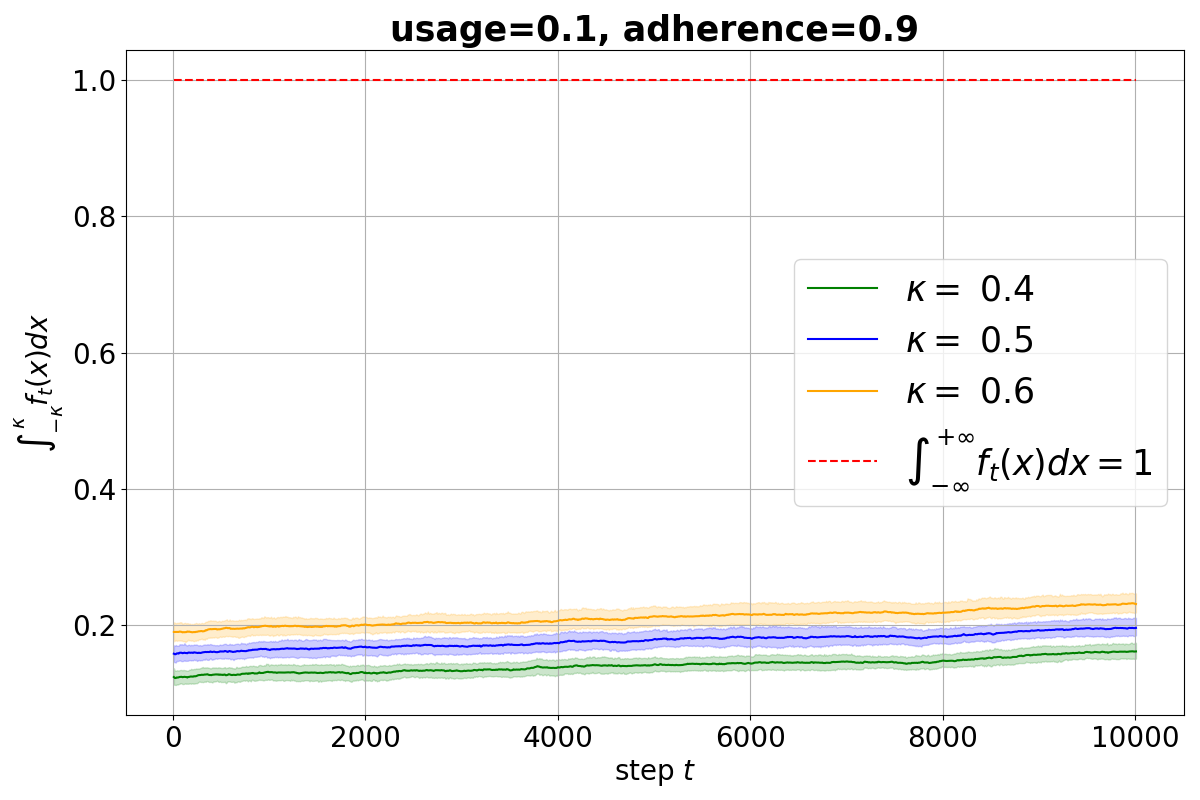}~
        \includegraphics[width=0.32\linewidth]{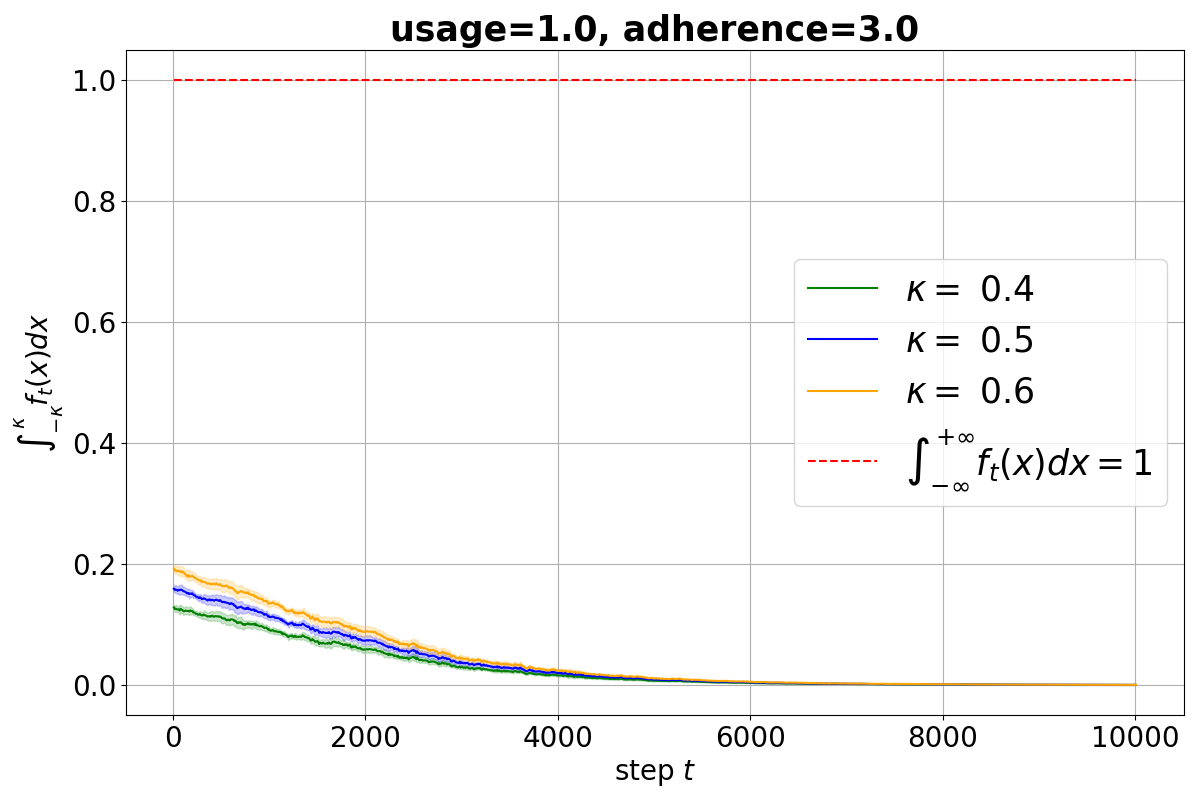}
        
        \caption{Counting $f_t(0)$ and $\int_{-\kappa}^{\kappa}f_t(x)dx$ for sampling update setup for Ridge regression model on Friedman data set. We consider such parameters: usage, adherence = $1$, $0$ (left); $0.1$, $0.9$ (middle); $1$, $3$ (right). In this picture, we can see the entire limit set of the system \eqref{system} from Theorem~\ref{delta}.}
        \label{delta_sample_1}
    \end{figure}

    Results on Fig.~\ref{delta_loop_1}~and~\ref{delta_sample_1} are similar to Fig.~\ref{delta_loop}~and~\ref{delta_sample}, therefore we can conclude that results of Theorem~\ref{delta} generalize to different data sets and machine learning models.

\end{document}